\def\year{2021}\relax
\documentclass[letterpaper]{article} 
\usepackage{aaai21}  
\usepackage{times}  
\usepackage{helvet} 
\usepackage{courier}  
\usepackage[hyphens]{url}  
\usepackage{graphicx} 
\usepackage{natbib}  
\usepackage{caption} 
\usepackage{placeins}

\graphicspath{{images/}}

\usepackage{amsmath}
\usepackage{amssymb}
\usepackage{amsthm}
\usepackage{amsfonts}
\usepackage{hyperref}
\usepackage{booktabs}
\usepackage{subfigure}
\usepackage{multirow}
\usepackage{enumitem}
\usepackage[switch]{lineno}
\usepackage{tabularx}

\let\vec\mathbf

\usepackage{xcolor}

\title{\textsc{Bayes-TrEx}: a Bayesian Sampling Approach to\\Model Transparency by Example}
\author{
    Serena Booth\textsuperscript{\rm *},
    Yilun Zhou\textsuperscript{\rm *},
    Ankit Shah,
    Julie Shah
}
\affiliations{

    \textsuperscript{\rm *}Equal Contribution\\
    CSAIL, Massachusetts Institute of Technology\\
    \{serenabooth, yilun, ajshah, julie\_a\_shah\}@csail.mit.edu

}

\newcommand{\BLOCKCOMMENT}[1]{}

\begin{document}
\maketitle

\begin{abstract}
Post-hoc explanation methods are gaining popularity for interpreting, understanding, and debugging neural networks. Most analyses using such methods explain decisions in response to inputs drawn from the test set. However, the test set may have few 
examples that trigger some model behaviors, such as high-confidence failures or ambiguous classifications. To address these challenges, we introduce a flexible model inspection framework: \textsc{Bayes-TrEx}. Given a data distribution, \textsc{Bayes-TrEx} finds in-distribution examples with a specified prediction confidence. We demonstrate several use cases of \textsc{Bayes-TrEx}, including revealing highly confident (mis)classifications, visualizing class boundaries via ambiguous examples, understanding novel-class extrapolation behavior, and exposing neural network overconfidence. We use \textsc{Bayes-TrEx} to study classifiers trained on CLEVR, MNIST, and Fashion-MNIST, and we show that this framework enables more flexible holistic model analysis than just inspecting the test set. Code is available at \url{https://github.com/serenabooth/Bayes-TrEx}.

\end{abstract}

\begin{figure}[h!]
    \centering
    \includegraphics[width=0.826\columnwidth]{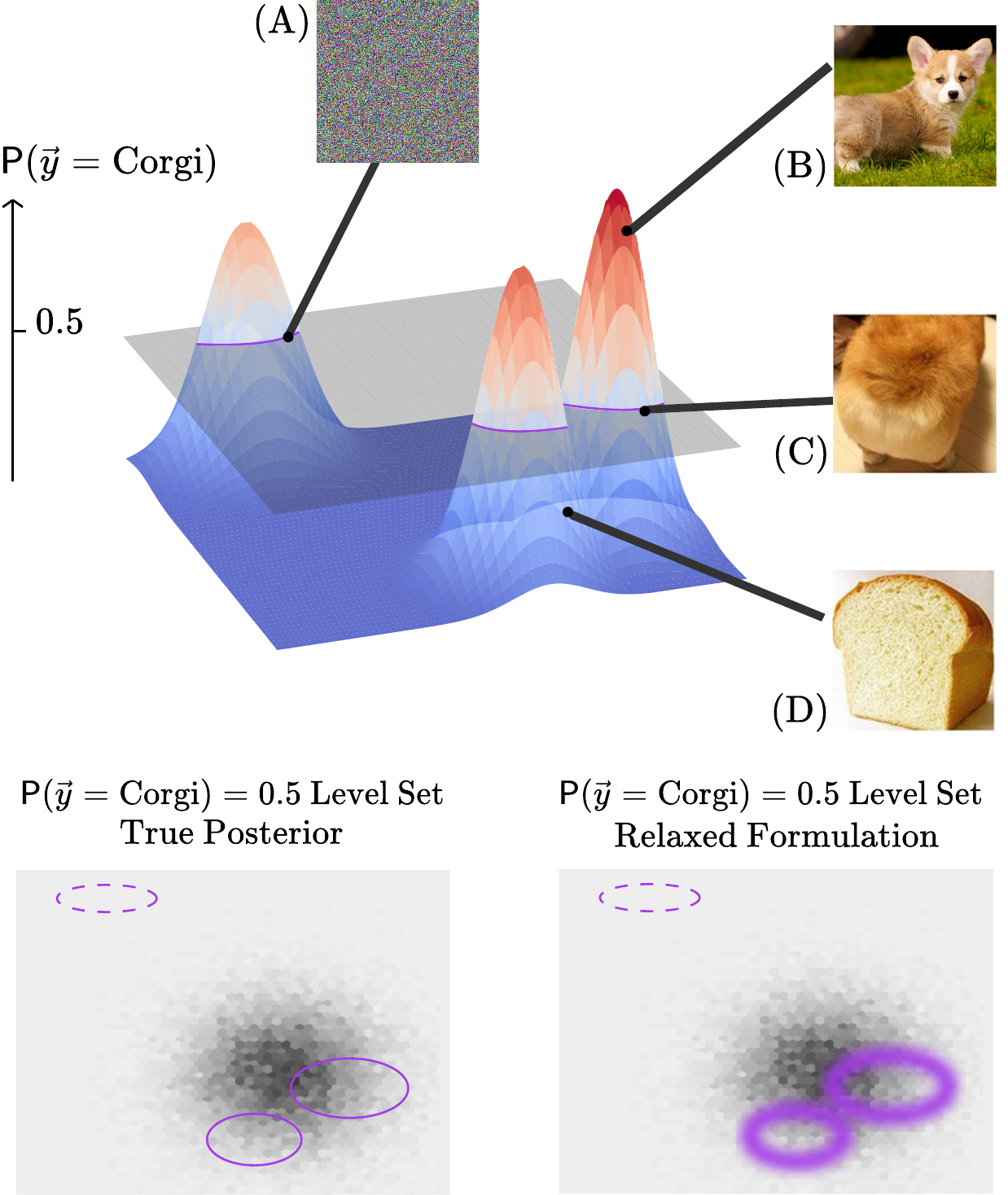}
    \caption{Top: given a Corgi/Bread classifier, we generate \emph{prediction level sets}, or sets of examples of a target prediction confidence. One way of finding such examples is by perturbing an arbitrary image to the target confidence (e.g., $\vec p_\text{Corgi}=\vec p_\text{Bread}=0.5$), as shown in (A). However, such examples give little insights into the typical model behavior because they are extremely unlikely in realistic situations. 
    \textsc{Bayes-TrEx} explicitly considers a data distribution (gray shade on the bottom plots) and finds in-distribution examples in a particular level set (e.g., likely Corgi (B), likely Bread (D), or ambiguous between Corgi and Bread (C)). 
    Bottom left: the classifier level set of $\vec p_\text{Corgi}=\vec p_\text{Bread}=0.5$ overlaid on the data distribution. Example (A) is unlikely to be sampled by \textsc{Bayes-TrEx} due to near-zero density under the distribution, while example (C) is likely to be. Bottom right: Sampling directly from the true posterior is infeasible, so we relax the formulation by ``widening'' the level set. By using different data distributions and confidences, \textsc{Bayes-TrEx} can uncover examples that invoke various model behaviors to improve model transparency. 
    }
    \label{fig:overview}
\end{figure}

\section{Introduction}

\begin{figure}[t]
\centering
    \includegraphics[width=\columnwidth,trim={1cm 0.1cm 0 0},clip]{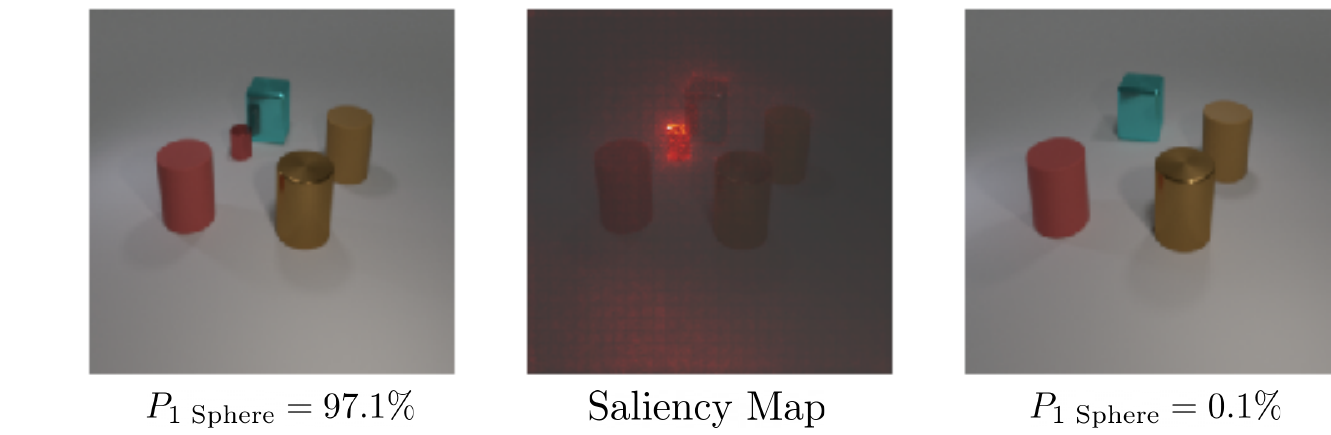}
    \caption{\textsc{Bayes-TrEx} finds a CLEVR scene which is incorrectly classified as containing a sphere. The generated example (left) is composed of only cylinders and cubes, but the classifier is $97.1\%$ confident this scene contains one sphere. The SmoothGrad~\cite{smilkov2017smoothgrad} saliency map highlights the small red cylinder as the object that is confused for a sphere. When we remove it, the classifier's confidence that the scene contains one sphere drops to $0.1\%$.
    }
    \label{fig:saliency-map-example-intro}
\end{figure}

Debugging, interpreting, and understanding neural networks can be challenging~\cite{doshi2017towards, lipton2018mythos, pmlr-v97-odena19a}. Existing interpretability methods include visualizing filters~\cite{zeiler2014visualizing}, saliency maps~\cite{simonyan2013deep}, input perturbations~\cite{ribeiro2016should, lundberg2017unified}, prototype anchoring~\cite{li2018deep, chen2019looks}, tracing with influence functions~\cite{koh2017understanding}, and concept quantification~\cite{ghorbani2019automating}. While some methods analyze intermediary network components such as convolutional layers~\cite{bau2017network, olah2017feature}, most methods instead explain decisions based on specific inputs. These inputs are typically selected from the test set, which may lack examples that lead to highly confident misclassifications or ambiguous predictions. Thus, it may be challenging to extract meaningful insights and attain a holistic understanding of model behaviors by using only test set inputs. Finding and analyzing inputs that invoke the gamut of model behaviours would improve \emph{model transparency by example}. 


To move beyond the test set, \textsc{Bayes-TrEx} takes a data distribution---either manually defined or learned with generative models---and finds in-distribution examples that trigger various model behaviors. \textsc{Bayes-TrEx} finds examples with target prediction confidences $\vec p$ by applying Markov-Chain Monte-Carlo (MCMC) methods on the posterior of a hierarchical Bayesian model. For example, Fig.~\ref{fig:overview} shows a Corgi/Bread classifier. For different $\vec p$-level set targets (e.g. $\vec p_\text{Corgi}=\vec p_\text{Bread}=0.5$), \textsc{Bayes-TrEx} can find examples where the model is highly confident in the Corgi class, in the Bread class, or ambiguous between the two. 
We use \textsc{Bayes-TrEx} to analyze classifiers trained on CLEVR~\cite{johnson2017clevr} with a manually defined data distribution, as well as MNIST~\cite{lecun-mnisthandwrittendigit-2010} and Fashion-MNIST~\cite{xiao2017fashion} with data distributions learned by variational autoencoders (VAEs)~\cite{kingma2013auto} or generative adversarial networks (GANs)~\cite{goodfellow2014generative}.

\textsc{Bayes-TrEx} can aid model transparency by example across several contexts.
Each context requires a different data distribution and a specified prediction confidence target. For example, \textsc{Bayes-TrEx} can generate \emph{ambiguous} examples to visualize class boundaries; \emph{high-confidence misclassification} examples to understand failure modes; \emph{novel class} examples to study model extrapolation behaviors; and \emph{high-confidence} examples to reveal model overconfidence (e.g.,  in domain-adaptation). In all of these use cases, the discovered examples can be further assessed with existing local explanation techniques such as saliency maps (Fig.~\ref{fig:saliency-map-example-intro}). 

The main current alternative to \textsc{Bayes-TrEx} is to inspect a model by using test set examples. As a baseline comparison, we search for highly confident misclassifications and ambiguous examples in the (Fashion-)MNIST and CLEVR test sets. We find few such test set examples meet these constraints, and the majority of these can be attributed to mislabeling in the dataset collection pipeline rather than misclassification by the model. In contrast, \textsc{Bayes-TrEx} consistently finds more highly confident misclassified and ambiguous examples, which enables more flexible and comprehensive model inspection and understanding.

\section{Related Work}

\subsection{Model Transparency}


Broadly, transparency is achieved when a user can develop a correct understanding and expectation of model behavior. One common technique for developing transparency is the test set confusion matrix: this matrix expresses the classifier's tendency of mistaking one class for another. Other transparency methods try to ``open" black-box models---for example, by visualizing convolutional filters through optimization~\cite{erhan2009visualizing, olah2017feature} or image patches~\cite{bau2017network}. Like \textsc{Bayes-TrEx}, other transparency methods communicate model behaviors through examples---for example, with counterfactuals~\cite{antoran2020getting,kenny2020generating} or with student-teacher learning examples~\cite{pruthi2020evaluating}. 

Some transparency methods aim to explain a model's response to an individual input. For example, saliency maps compute a heat map over the input that represents the importance of each pixel~\cite{simonyan2013deep, zeiler2014visualizing}. Importantly, these input-based methods require a two-stage pipeline: finding interesting inputs $\rightarrow$ explaining the model responses (e.g., with saliency maps). Current efforts are focused on the second stage with inputs simply retrieved from the test set. To the best of our knowledge, \textsc{Bayes-TrEx} is the first work dedicated to the first stage of finding interesting inputs. The examples uncovered by \textsc{Bayes-TrEx} can be used with any input-based method for further analysis (Fig.~\ref{fig:saliency-map-example-intro} and App.~\ref{saliency-maps-supp}).

\subsection{Model Testing}

\textsc{TensorFuzz} \cite{pmlr-v97-odena19a} is a fuzzing test framework for neural networks which finds inputs that achieve a wide coverage of user-specified constraints. \textsc{TensorFuzz} is similar to \textsc{Bayes-TrEx} in that both methods aim to find examples that elicit certain model behaviors. While \textsc{TensorFuzz} is designed to find \textit{rare} inputs that trigger edge cases such as numerical errors, \textsc{Bayes-TrEx} finds common, in-distribution examples. As such, \textsc{Bayes-TrEx} is more suitable to help humans develop a correct mental model of the classifier. \textsc{Scenic}~\cite{fremont2019scenic} is a domain-specific language for model testing by generating failure-inducing examples. While \textsc{Bayes-TrEx} is in part inspired by \textsc{Scenic}, its formulation is more flexible.

\subsection{Natural Adversarial Examples}

One \textsc{Bayes-TrEx} use case is uncovering high-confidence classification failures in the data distribution. This idea is related to, but different from, natural adversarial attacks \cite{zhao2018generating}. Most adversarial attacks inject crafted high-frequency information to mislead a trained model~\cite{szegedy2013intriguing, goodfellow2014explaining, nguyen2015deep}, but such artifacts are non-existent in natural images. Zhao et al.~\shortcite{zhao2018generating} instead proposed a method to find \emph{natural} adversarial examples by performing the perturbation in the latent space of a GAN. While this method finds an example which looks like a specific input, \textsc{Bayes-TrEx} finds high-confidence misclassifications in the entire data distribution. 

\subsection{Confidence in Neural Networks}
\textsc{Bayes-TrEx} can also be used to detect overconfidence in neural networks. An overconfident neural network~\cite{guo2017calibration} makes many mistakes with disproportionately high confidence. While many approaches aim to address this network overconfidence problem
\cite{blundell2015weight, gal2016dropout, lee2017training, thulasidasan2019mixup}, \textsc{Bayes-TrEx} is complementary to these efforts. Rather than altering the confidence of a neural network, it instead infers examples of a particular confidence. If the model is overconfident, it may return few, if any, samples with ambiguous predictions. At the same time, it may find many misclassifications with high confidence. In our experiments (Sec. \ref{sec:da}), we discover that the popular adversarial discriminative domain adaptation (ADDA) technique produces a more overconfident model than the baseline.

\section{Methodology}

\label{sec:methodology}

Given a classifier $f: X \rightarrow \Delta_K$ which maps a data point to the probability simplex of $K$ classes, the goal is to find an input $\vec x\in X$ in a given data distribution $p(\vec x)$ such that $f(\vec x)=\vec p$ for some prediction confidence $\vec p \in \Delta_K$. We consider the problem of sampling from the posterior
\begin{align} \label{eqn:infeasible_posterior}
    p(\vec x|f(\vec x)=\vec p)&\propto p(\vec x) \, p(f(\vec x)=\vec p|\vec x).
\end{align}

A common approach to posterior sampling is to use Markov Chain Monte-Carlo (MCMC) methods~\cite{brooks2011handbook}. However, when the measure of the level set $\{\vec x: f(\vec x)=\vec p\}$ is small or even zero, sampling directly from this posterior using MCMC is infeasible: the posterior being zero everywhere outside of the level set makes it unlikely for a random-walk Metropolis sampler to land on $\vec x$ with non-zero posterior~\cite{hastings1970monte}, and the gradient being zero everywhere outside of the level set means that a Hamiltonian Monte Carlo sampler does not have the necessary gradient guidance toward the level set~\cite{neal2011mcmc}. 

To enable efficient sampling, we relax the formulation by ``widening'' the level set and accepting $\vec x$ when $f(\vec x)$ is close to the target $\vec p$ (Fig.~\ref{fig:overview}). Specifically, we introduce a random vector $\vec u = [u_1, \dots, u_K]^T$, distributed as
\begin{align}\label{eqn:general_case}
    u_i | \vec x \sim \mathcal{N}\left(f(\vec x)_i, \sigma^2\right), 
\end{align}
where $\sigma$ is a hyper-parameter.

Instead of directly sampling from Eqn.~\ref{eqn:infeasible_posterior}, we can now sample from the new posterior: 
\begin{align}
    p(\vec x|\vec u= \vec u^*)&\propto p(\vec x)p(\vec u= \vec u^*|\vec x),\\
    \vec u^*&=\vec p.
\end{align}

The hyper-parameter $\sigma$ controls the peakiness of the relaxed posterior. A smaller $\alpha$ makes it closer to the true posterior and makes the distribution peakier and harder to sample, while a larger $\alpha$ makes it closer to the data distribution $p(\vec x)$ and easier to sample. As $\sigma$ goes to 0, it approaches the true posterior. Formally, 
\begin{align}
\lim_{\sigma\rightarrow 0}p(\vec x|\vec u=\vec u^*)=p(\vec x|f(\vec x)=\vec p). 
\end{align}

While the formulation in Eqn.~\ref{eqn:general_case} is applicable to arbitrary confidence $\vec p$, the dimension of $\vec u$ is equal to the number of classes, which poses scalability issues for large numbers of classes. However, for a wide range of interesting use cases of \textsc{Bayes-TrEx}, we can use the following reductions: 
\begin{enumerate}[leftmargin=*]
    \item \label{case:highconf} Highly confident in class $i$: $\vec p_i=1, \vec p_{\lnot i}=0$. We have
\begin{align}\label{eqn:high-conf}
    u | \vec x \sim \mathcal{N}\left(f(\vec x)_i, \sigma^2\right),\,\,\,\,\,\,\,\,\,\,\,u^* = 1. 
\end{align}
    \item \label{case:ambivalent} Ambiguous between class $i$ and $j$: $\vec p_i = \vec p_j=0.5$, $\vec p_{\lnot i,j}=0$. We have
\begin{align}\label{eqn:ambiv-conf-1}
    u_1 | \vec x &\sim \mathcal{N}\left(|f(\vec x)_i-f(\vec x)_j|, \sigma_1^2\right), \\
    u_2|\vec x &\sim \mathcal{N}(\min(f(\vec x)_i, f(\vec x)_j)-\max_{k\neq i, j}f(\vec x)_k, \sigma_2^2), \label{eqn:ambiv-conf-2} \\
     u_1^*&=0, u_2^*=0.5. 
\end{align}
$\sigma_1$ and $\sigma_2$ are hyperparameters.
\end{enumerate}

In addition, most high dimensional data distributions, such as those for images, are implicitly defined by a transformation $g: Z\rightarrow X$ from a latent distribution $p(\vec z)$. Consequently, given
\begin{align}
    \vec x &= g(\vec z), \\
    \vec u | \vec z &\sim \mathcal{N}(f(\vec x), \sigma^2),\\
    p(\vec z|\vec u=\vec u^*)&\propto p(\vec z)p(\vec u=\vec u^*|\vec z),\label{posterior}
\end{align}
\textsc{Bayes-TrEx} samples $\vec z$ according to Eqn.~\ref{posterior} and reconstruct the example $\vec x=g(\vec z)$ for model inspection.

\section{Experiments}

\label{section:experiments}

\subsection{Overview}

\begin{figure}[b]
    \centering
    \includegraphics[width=\columnwidth]{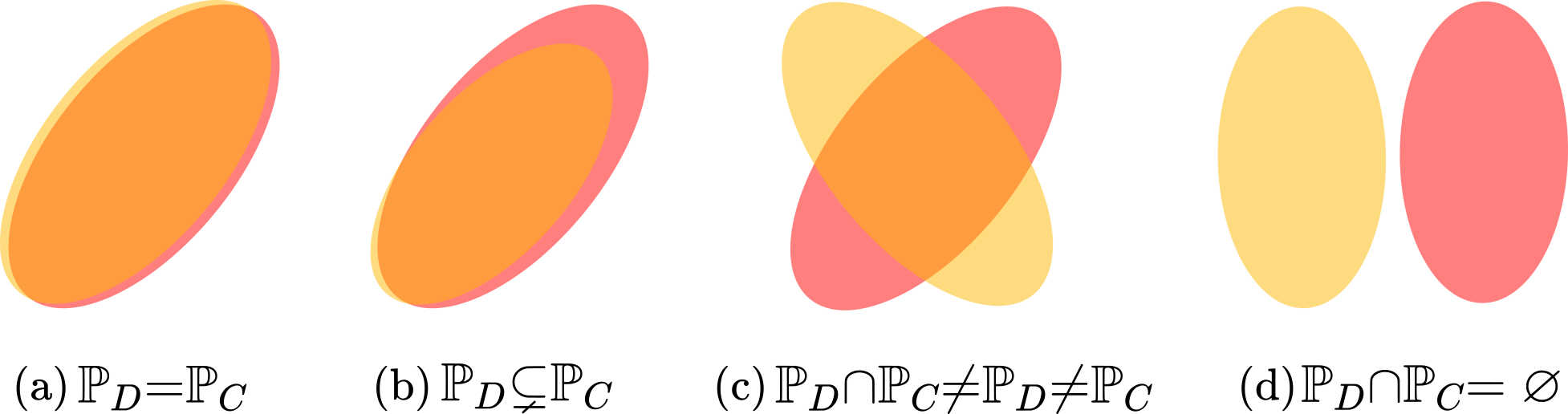}
    \caption{Different relations between the classifier training distribution ($\mathbb P_C$, red) and \textsc{Bayes-TrEx} data distribution ($\mathbb P_D$, yellow). (a) $\mathbb P_C$ and $\mathbb P_D$ are equal. (b) The support of $\mathbb P_D$ is a subset of that of $\mathbb P_C$. (c) $\mathbb P_D$ and $\mathbb P_C$ have overlapping supports. (d) Supports of $\mathbb P_C$ and $\mathbb P_D$ are disjoint. }
    \label{fig:distrs}
\end{figure}

A key strength of \textsc{Bayes-TrEx} is the ability to evaluate a classifier on a data distribution $\mathbb P_D$, independent of its training distribution $\mathbb P_C$. We demonstrate the versatility of \textsc{Bayes-TrEx} on four relationships between $\mathbb P_D$ and $\mathbb P_C$ (Fig.~\ref{fig:distrs}). With $\mathbb P_C=\mathbb P_D$ (Fig.~\ref{fig:distrs}(a)), Sec.~\ref{sec:high-conf-2} and \ref{sec:ambiguous-conf} present examples that trigger high and ambiguous model confidence and Sec.~\ref{sec:graded} presents examples that interpolate between two classes. In Sec.~\ref{sec:adv}, we consider $\mathbb P_D$ with narrower support than $\mathbb P_C$ (Fig.~\ref{fig:distrs}(b)), where the support of $\mathbb P_D$ excludes examples from a particular class. In this case, high-confidence examples---as judged by the classifier---correspond to high-confidence misclassifications. In Sec.~\ref{sec:ood} and \ref{sec:da}, we analyze the classifier $C$ for novel class extrapolation and domain adaptation behaviors with overlapping or disjoint supports of $\mathbb P_C$ and $\mathbb P_D$ (Fig.~\ref{fig:distrs}(c, d)). Representative results are in the main text; further results are in the appendix.

\subsection{Datasets and Inference Details}
\label{sec:dataset-setup}
We evaluate \textsc{Bayes-TrEx} on rendered images (CLEVR) and organic datasets (MNIST and Fashion-MNIST). For all CLEVR experiments, we use the pre-trained classifier distributed by the original authors\footnote{\url{https://github.com/facebookresearch/clevr-iep}}. The transition kernel uses a Gaussian proposal for the continuous variables (e.g., $x$-position) and categorical proposal for the discrete variables (e.g., color), both centered around and peaked at the current value. For (Fashion-)MNIST experiments, architectures and training details are described in Appx.~\ref{arch-supp}. For domain adaptation analysis, we train ADDA and baseline models using the code provided by the authors\footnote{\url{https://github.com/erictzeng/adda}}.

CLEVR images are rendered from scene graphs, on which we define the latent distribution $p(\vec z)$. Since the (Fashion-)MNIST groundtruth data distribution is unknown, we estimate it using a VAE or GAN with unit Gaussian $p(\vec z)$. These learned data distribution representations have known limitations, which may affect sample quality~\cite{arora2017gans}. Table~\ref{tab:fid-main} lists the Fr\'echet Inception Distance (FID)~\cite{heusel2017gans} for two VAE and GAN models, with the full table in Appx.~\ref{fid-supp}. The FID scores show the GANs generate more representative samples than the VAEs. 

\begin{table}[t]
    \centering
    \caption{Fr\'echet Inception Distance (FID) for VAE and GAN models trained on the entire dataset. A lower value indicates higher quality. Appx.~\ref{fid-supp} presents the statistics for all models. }
    \begin{tabular}{lll}\toprule
    \small
        Model & Dataset & FID \\\midrule
        \multirow{2}{*}{VAE} & MNIST & 72.33 \\
            & Fashion-MNIST & 87.89 \\\midrule
        \multirow{2}{*}{GAN} & MNIST & 11.83 \\
            & Fashion-MNIST & 29.44 \\\bottomrule
    \end{tabular}
    \label{tab:fid-main}
\end{table}

We consider two MCMC samplers: random-walk Metropolis (RWM) and Hamiltonian Monte Carlo (HMC). We use the former in CLEVR where the rendering function is non-differentiable, and the latter for (Fashion-)MNIST. For HMC, we use the No-U-Turn sampler \cite{hoffman2014no,neal2011mcmc} implemented in the probabilistic programming language Pyro~\cite{bingham2018pyro}. We choose $\sigma=0.05$ for all experiments. Alternatively, $\sigma$ can be annealed to gradually reduce the relaxation.

Selecting appropriate stopping criteria for MCMC methods is an open problem. State-of-the-art approaches require a gold standard inference algorithm~\cite{cusumano2017aide} or specific posterior distribution properties, such as log-concavity~\cite{gorham2015measuring}. As neither of these requirements are met for our domains, we select stopping criteria based on heuristic performance and cost of compute. CLEVR scenes require GPU-intensive rendering, so we stop after 500 samples. (Fashion-)MNIST samples are cheaper to generate, so we stop after 2,000 samples. Empirically, we find each sampling step takes 3.75 seconds for CLEVR, 1.18s for MNIST, and 1.96s for Fashion-MNIST, all on a single NVIDIA GeForce 1080 GPU.

\subsection{Quantitative Evaluation}
\label{sec:quant}
We first evaluate the quality of \textsc{Bayes-TrEx} samples by assessing whether the classifier's prediction confidence matches the specified target on the generated examples. Table~\ref{tab:quant-conf} presents the mean and standard deviation of the confidence on a selection of representative settings, and Appx.~\ref{full-table-supp} lists the full set of such evaluations. In general, the prediction confidences are tightly concentrated around the target, indicating sampler success. 

\begin{table}[t]
\centering
\caption{Mean and standard deviation of the prediction confidence of the samples. Reported values are for the target class, or two target classes in ambiguous confidence and confidence interpolation cases. Appx.~\ref{full-table-supp} presents the full table of statistics for all experiments.}
\resizebox{\linewidth}{!}{
\begin{tabular}{ lllr }
\toprule
Use Case & Dataset & Target & Prediction Confidence \\  \midrule
\multirow{3}{*}{High Conf.}
 & MNIST & $\vec p_4 = 1$ & 1.00 $\pm$ 0.01 \\
 & Fashion & $\vec p_\text{Coat} = 1$ & 0.98 $\pm$ 0.02 \\
 & CLEVR & $\vec p_\text{2 Blue Spheres} = 1$ & 0.89 $\pm$ 0.25 \\\midrule
\multirow{2}{*}{Ambiguous} 
 & MNIST & $\vec p_1 = \vec p_7 = 0.5$ & $0.49 \pm 0.02, 0.49 \pm 0.03$\\
 & Fashion & $\vec p_\text{T-shirt} = \vec p_\text{Dress} = 0.5$ & $0.48 \pm 0.02, 0.48 \pm 0.02$ \\\midrule
\multirow{2}{*}{Interpolation}
 & MNIST & $\vec p_{8}=0.6, \vec p_{9}=0.4$ & $0.58 \pm 0.04, 0.37 \pm 0.04$\\
 & Fashion & $\vec p_{\text{T-Shirt}}=0.2, \vec p_{\text{Trousers}}=0.8$ & $0.17 \pm 0.04, 0.79 \pm 0.04$ \\\midrule
\multirow{3}{*}{Misclassified}
 & MNIST & $\vec p_8 = 1$ & 0.98 $\pm$ 0.02 \\
 & Fashion & $\vec p_\text{Bag} = 1$ & 0.97 $\pm$ 0.03 \\
 & CLEVR & $\vec p_\text{1 Cube} = 1$ & 0.93 $\pm$ 0.06 \\
 \midrule
\multirow{3}{*}{Extrapolation}
 & MNIST & $\vec p_6 = 1$             & 1.00 $\pm$ 0.01 \\
 & Fashion & $\vec p_\text{Sandal} = 1$ & 1.00 $\pm$ 0.01 \\
 & CLEVR & $\vec p_\text{1 Cylinder}=1$ & 0.96 $\pm$ 0.03 \\
 \midrule
Domain Adapt. & MNIST & $\vec p_5 = 1$ & 1.00 $\pm$ 0.01 \\\bottomrule
\end{tabular}
}
\label{tab:quant-conf}
\end{table}

\subsection{High Confidence}
\label{sec:high-conf-2}

\begin{figure}[t]
    \centering
    \subfigure[$\vec p_\text{5 Spheres} = 95.7\%$]{\includegraphics[width=0.36\columnwidth]{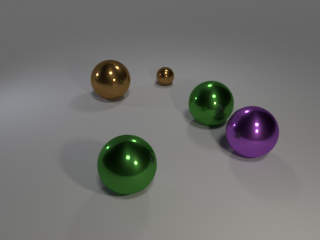}\label{fig:confident_5_spheres}}
    \subfigure[$\vec p_\text{2 Blue Sph.} = 91.1\%$]{\includegraphics[width=0.36\columnwidth]{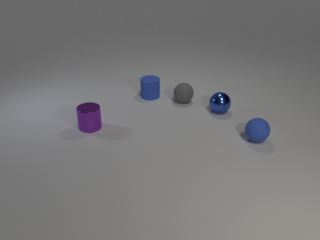}\label{fig:confident_2_blue_spheres}}
    \subfigure[MNIST]{\includegraphics[width=0.36\columnwidth]{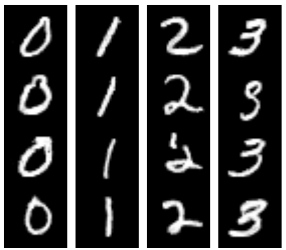}}
    \subfigure[Fashion-MNIST]{\includegraphics[width=0.36\columnwidth]{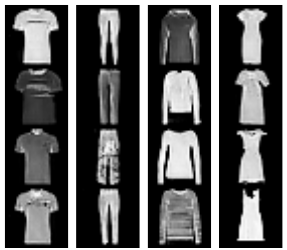}}
    \caption{High-confidence samples. (a, b) CLEVR. (c) MNIST, digits 0-3. (d) Fashion-MNIST, left to right: T-shirt, trousers, pullover and dress. More examples in Appx.~\ref{high-conf-supp}. }
    \label{fig:high-confidence}
\end{figure}

As an initial smoke test, we evaluate \textsc{Bayes-TrEx} by finding highly confident examples. (Fashion-)MNIST data distributions are learned by GAN. Fig.~\ref{fig:high-confidence} depicts samples on the three datasets. Additional examples are in Appx.~\ref{high-conf-supp}.

\subsection{Ambiguous Confidence}
\label{sec:ambiguous-conf}
\begin{figure}[t]
    \centering
     \includegraphics[width=0.9\columnwidth]{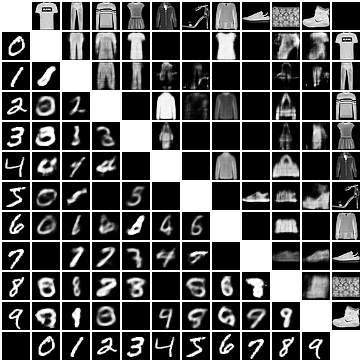}    \caption{Each entry of the matrix is an ambiguous MNIST or Fashion-MNIST example for the classes on its row and column. Blacked-out cells indicate sampling failures. Examples on the outermost edges of the matrix are class representations (e.g., 0-9 for MNIST).}
\label{fig:ambivalent}
\end{figure}

Next, we find ambiguous (Fashion-)MNIST examples for which the classifier has similar prediction confidence between two classes, using data distributions learned by a VAE. Fig.~\ref{fig:ambivalent} shows ambiguous examples from each pair of classes (e.g. 0v1, 0v2, ..., 8v9). Note the examples presented are ambiguous from the classifier's perspective, though some may be readily classified by a human. Not all pairs result in successful sampling: for example, we were unable to find an ambiguous example with equal prediction confidence between the visually dissimilar classes $0$ and $7$. These ambiguous examples are useful for visualizing and understanding class boundaries; Appx.~\ref{ambivalent-supp} presents a supporting class boundary latent space visualization. \emph{Blended} ambiguous examples have previously been shown to be useful for data augmentation~\cite{tokozume2018between}. While these generated ambiguous examples may be similarly useful, we leave this exploration to future work.

\begin{figure}[!tbh]
    \centering
    \includegraphics[width=0.8\columnwidth, trim={0 0 0 310px}, clip]{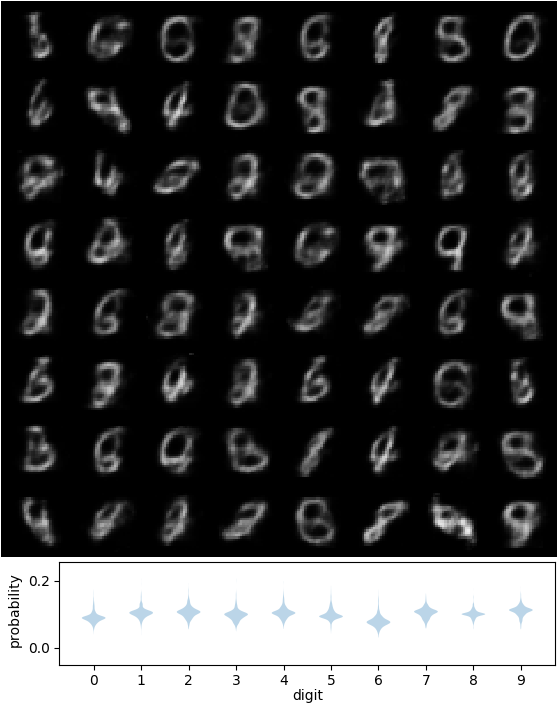}
    \caption{Samples of uniformly ambiguous predictions.}
    \label{fig:uniform-ambigu-main}
\end{figure}

\textsc{Bayes-TrEx} can also find examples which are ambiguous across more than two classes; Fig.~\ref{fig:uniform-ambigu-main} presents samples that are equally ambiguous across all 10 MNIST classes. All these images appear to be very blurry and not very realistic. This is intuitive: even for a human, it would be hard to write a digit in such a way that it is equally unrecognizable across all 10 classes. Details about the sampling formulation and visualizations are presented in Appx.~\ref{ambivalent-supp}. 

\begin{figure}[!htb]
    \centering
    \includegraphics[height=0.4\columnwidth]{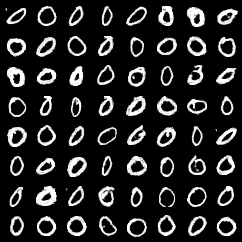}
    \includegraphics[height=0.4\columnwidth, trim={0 0 0 30px}, clip]{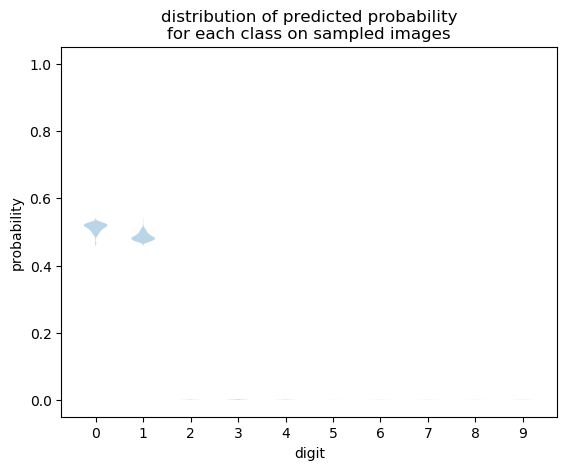}
    \caption{0v1 ambiguous samples and confidence plot with the GAN distribution and always ambiguous classifier. }
    \label{fig:gan-study}
\end{figure}

For ambiguous examples, we observed only rare successes with data distributions learned by a GAN, which generates sharper and more visually realistic images than a VAE. There are two candidate explanations: \begin{enumerate}
    \item GAN-distributions prevent efficient MCMC sampling.
    \item The classifier rarely makes ambiguous predictions on sharp and realistic images.
\end{enumerate}
To experimentally check the second explanation, we train a classifier to be consistently ambiguous between class $i$ and $i+1$ for an image of digit $i$ (wrapping around at $10=0$) using the following KL-divergence loss: 
\begin{align}
    l(y, f(\vec x)) &= \mathbb{KL}(\vec p_y, f(\vec x)), \label{eq:kl}\\
    \vec p_{y, i} &= \begin{cases}
    0.5 & \text{$i=y$ or $i=(y+1)$ mod 10}, \\
    0 & \text{otherwise}. 
    \end{cases}
\end{align}
Using this classifier, we sample ambiguous examples for 0v1, 1v2, ..., 9v0. Sampling succeeds for all ten pairs, even when using the same GAN model that rarely succeeded in the prior experiment. Fig.~\ref{fig:gan-study} presents the 0v1 samples and predicted confidence by this modified classifier, and the remaining pairs are visualized in Appx.~\ref{gan-failure-supp}. Given this sampling success, we conclude that the second explanation is correct. 

In addition, \textsc{Bayes-TrEx} is also unable to generate ambiguous examples for CLEVR with the manually defined data distribution. Given that the pre-trained classifier only achieves $\approx$60\% accuracy, the result suggests that the model is likely overconfident. Indeed, this has previously been observed in similar settings~\cite{kim2018not}. 

\subsection{Confidence Interpolation}
\label{sec:graded}
\begin{figure}[!h]
    \centering
    \includegraphics[height=0.97\columnwidth]{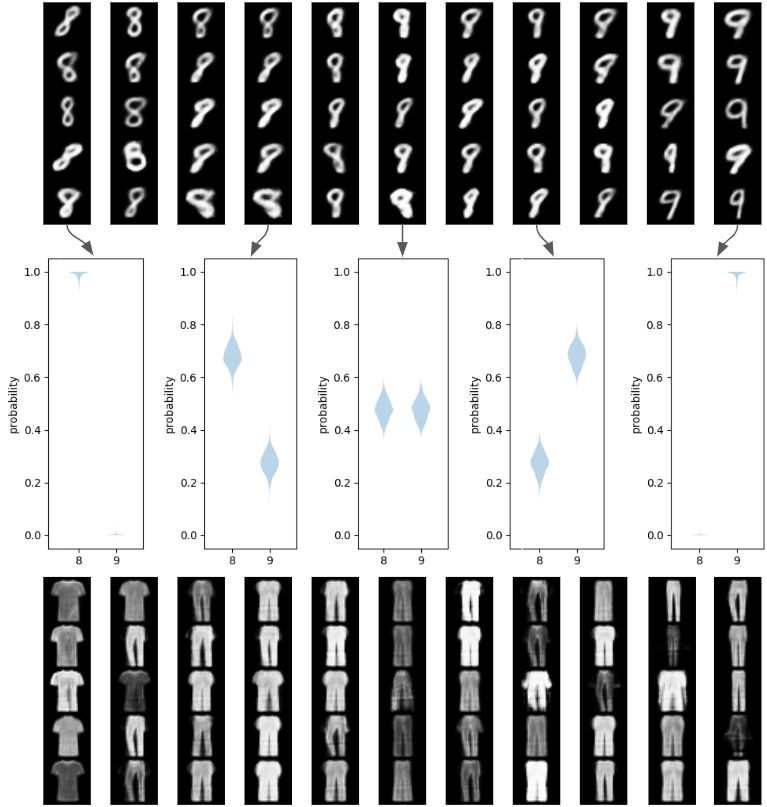}
    \caption{Confidence interpolation between digit 8 and 9 for MNIST and between T-shirt and trousers for Fashion-MNIST. Each of the 11 columns show samples of confidence ranging from [$\vec p_\text{class a} = 1.0$, $\vec p_\text{class b} = 0.0$] (left) to [$\vec p_\text{class a} = 0.0$, $\vec p_\text{class b} = 1.0$] (right), with an interval of 0.1. Some confidence plots for MNIST are shown in the middle. }
    \label{fig:graded-supp}
\end{figure}

\textsc{Bayes-TrEx} can find examples that interpolate between classes. In Fig.~\ref{fig:graded-supp}, we show MNIST samples which interpolate from $(P_8=1.0, P_9=0.0)$ to $(P_8=0.0, P_9=1.0)$ and Fashion-MNIST samples from $(P_{\text{T-shirt}}=1.0, P_{\text{Trousers}}=0.0)$ to $(P_{\text{T-shirt}}=0.0, P_{\text{Trousers}}=1.0)$ over intervals of $0.1$, with a VAE-learned data distribution. 

The interpolation between two very different classes reveal insights into the model behavior. For example, the interpolation from 8 to 9 generally shrinks the bottom circle toward a stroke, which is the key difference between digits 8 and 9. For Fashion-MNIST, the presence of two legs is important for trousers classification, even appearing in samples with $(\vec p_{\text{T-shirt}}=0.9, \vec p_{\text{Trousers}}=0.1)$ (second column). By contrast, a wider top and the appearance of sleeves are important properties for T-shirt classification. These two trends result in most of the interpolated samples having a short sleeve on the top and two distinct legs on the bottom. 

\subsection{High-Confidence Failures}
\label{sec:adv}

\begin{figure}[t]
    \centering
    \subfigure[$\vec p_\text{1 Cube} = 93.5\%$]{\includegraphics[width=0.36\columnwidth]{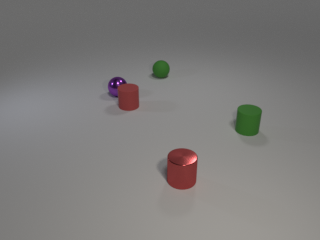}\label{fig:adversarial_1_cube}}
    \hspace{0.05in}
    \subfigure[$\vec p_\text{2 Cylinders} = 90.2\%$]{\includegraphics[width=0.36\columnwidth]{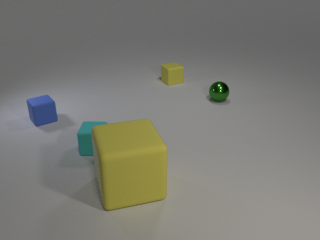}\label{fig:adversarial_2_cylinders}}
    \hspace{0.05in}
    \subfigure[MNIST]{\includegraphics[width=0.36\columnwidth]{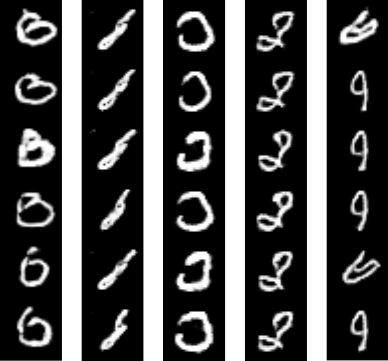}\label{fig:adversarial_mnist}}
    \hspace{0.05in}
    \subfigure[Fashion-MNIST]{\includegraphics[width=0.36\columnwidth]{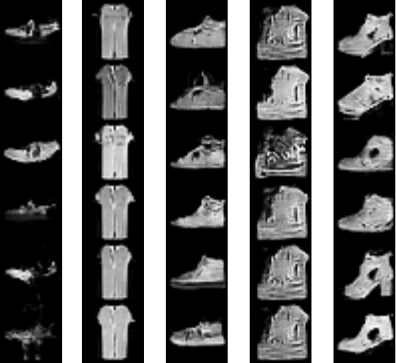}\label{fig:adversarial_fmnist}}
    \vspace{-2mm}
    \caption{High confidence classification failures. (a): CLEVR, 1 Cube. Note that no cube is present in the sample. (b): CLEVR, 2 Cylinders---again, containing no cylinders. (c) MNIST failures for digits 0-4. 0s are composed of 6s; 1s of 8s; 2s of 0s, and so on. (d) Fashion-MNIST failures for sandal, shirt, sneaker, bag, and ankle boot. Additional examples are presented in Appx.~\ref{adv-supp}. }
    \label{fig:adv}
    \vspace{-4mm}
\end{figure}

With neural networks being increasingly used for high-stakes decision making, high-confidence failures are one area of concern, as these failures may go unnoticed. \textsc{Bayes-TrEx} can find such failures. Specifically, if the data distribution (Fig.~\ref{fig:distrs}(b)) does \emph{not} include a particular class, then the resulting high-confidence examples correspond to high-confidence \textit{misclassifications} for that class. For example, in Fig.~\ref{fig:adversarial_1_cube}, the CLEVR classifier is highly confident that there is one cube though there is no cube in the image. In App.~\ref{saliency-maps-supp}, the saliency map for Fig.~\ref{fig:adversarial_1_cube} reveals that classifier mistakes the front shiny red cylinder for a cube. Removing this cylinder causes the confidence to drop to 29.0\%. In addition, such high-confidence failures can also be used for data augmentation to increase network reliability~\cite{fremont2019scenic}. 

For (Fashion-)MNIST, a GAN is trained on all data sans a single class, resulting in the learned data distribution excluding the given class. Figs.~\ref{fig:adversarial_mnist} and \ref{fig:adversarial_fmnist} depict high-confidence misclassifications for digits 0-4 in MNIST and sandal, shirt, sneaker, bag, and ankle boot in Fashion-MNIST, respectively. By evaluating these examples, we can assess how well human-aligned a classifier is. For example, for MNIST, some thin 8s are classified as 1s and particular styles of 6s and 9s are classified as 4s. These results seem intuitive, as a human might make these same mistakes. Likewise, for Fashion-MNIST, most failures come from semantically similar classes, e.g.~sneaker $\longleftrightarrow$ ankle boot. Less intuitively, however, chunky shoes are likely to be classified as bags. Additional visualizations are presented in Appx.~\ref{adv-supp}. 

\begin{figure}[t]
\centering
\hspace{-0.1in}\subfigure[$\vec p_\text{1 Cube} = 98.5\%$]{\includegraphics[width=0.36\columnwidth]{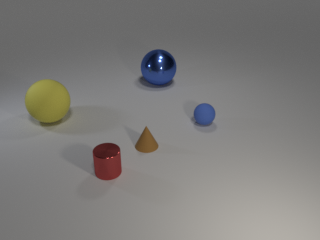}\label{fig:extrapolation_1_cube}}
    \hspace{0.05in}
\subfigure[$\vec p_\text{5 Cubes} = 92.5\%$]{\includegraphics[width=0.36\columnwidth]{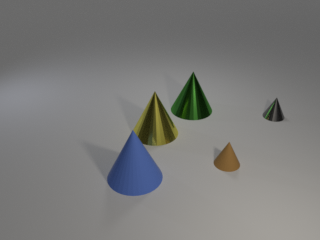}\label{fig:extrapolation_5_cubes}}
\subfigure[MNIST]{\includegraphics[height=0.32\columnwidth]{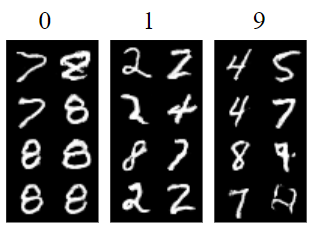}\label{fig:ood-mnist}}
\subfigure[Fashion-MNIST]{\includegraphics[height=0.32\columnwidth]{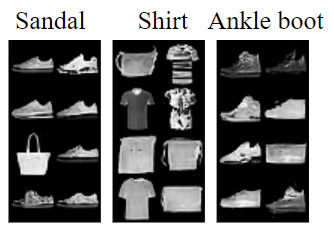}\label{fig:ood-fashion-mnist}}
\caption[short]{Novel class extrapolation examples. (a, b): For CLEVR, the novel cone objects are mistaken for cubes. (c, d): For (Fashion-)MNIST, we train classifiers on subsets of the data (digits 0, 1, 3, 6, 9 and pullover, dress, sandal, shirt, and ankle boot), and train GANs with the excluded data. Samples for which the classifier is highly confident ($\approx 99\%)$ in several target classes are shown (e.g., targets 0, 1, and 9 for MNIST). Additional examples are presented in Appx.~\ref{ood-supp}. }
\label{fig:ood-extrapolation}
\end{figure}

\subsection{Novel Class Extrapolation}
\label{sec:ood}

It is important to understand the novel class extrapolation behavior of a model before deployment. For example, during training an autonomous vehicle might learn to safely operate around pedestrians, cyclists, and cars. But can we predict how the vehicle will behave when it encounters a novel class, like a tandem bicycle? \textsc{Bayes-TrEx} can be used to understand such behaviors by sampling high-confidence examples with a data distribution that contains novel classes, while excluding the true target classes (Fig.~\ref{fig:distrs}(c, d)). 

For CLEVR, we add a novel cone object to the data distribution and remove the existing cube from it. We sample images that the classifier is confident to include cubes, shown in Fig.~\ref{fig:ood-extrapolation} (a, b). A saliency map analysis in Appx.~\ref{saliency-maps-supp} confirms that the classifier indeed mistakes these cones for cubes. In Appx.~\ref{ood-supp}, we assess CLEVR's novel class extrapolation for cylinders and spheres, and similarly show the model readily confuses cones for these classes as well.

For MNIST and Fashion-MNIST, we train the respective classifiers on digits 0, 1, 3, 6, 9 and pullover, dress, sandal, shirt and ankle boot classes. We train GANs using only the excluded classes (e.g., digits 2, 4, 5, 7, 8 for MNIST). Using these GANs, we find examples where the classifier has high prediction confidence, as shown in Fig.~\ref{fig:ood-extrapolation} (c, d). For MNIST, there are few reasonable extrapolation behaviors, most likely due to the visual distinctiveness between digits. By comparison, some Fashion-MNIST extrapolations are expected, such as confusing the unseen sneaker class for sandals and ankle boots. However, the classifier also confidently mistakes various styles of bags as sandals, shirts, and ankle boots. App.~\ref{ood-supp} contains additional visualizations.

\subsection{Domain Adaptation}
\label{sec:da}
Finally, we use \textsc{Bayes-TrEx} to analyze domain adaptation behaviors. We reproduce the SVHN~\cite{netzer2011reading} $\rightarrow$ MNIST experiment studied by Tzeng, et al.~\shortcite{tzeng2017adversarial}. We train two classifiers, a baseline classifier on labeled SVHN data only, and the ADDA classifier on labeled SVHN data and unlabeled MNIST data. Indeed, domain adaptation improves classification accuracy: 61\% for the baseline classifier on MNIST vs.~71\% for the ADDA classifier. 

\begin{figure}[t]
    \centering
    \subfigure[Baseline examples]{\includegraphics[width=0.8\columnwidth]{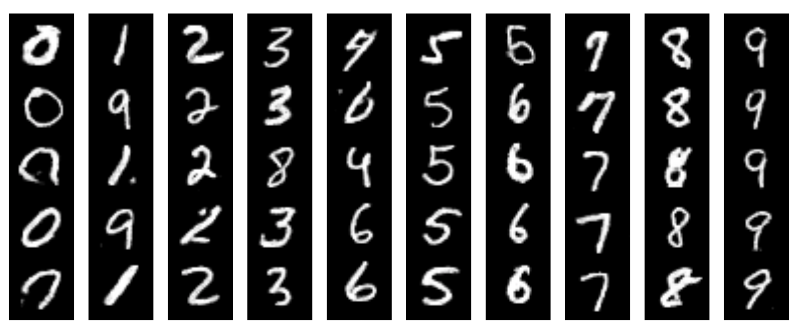} \label{fig:baseline-adda}}
    \subfigure[ADDA examples]{\includegraphics[width=0.8\columnwidth]{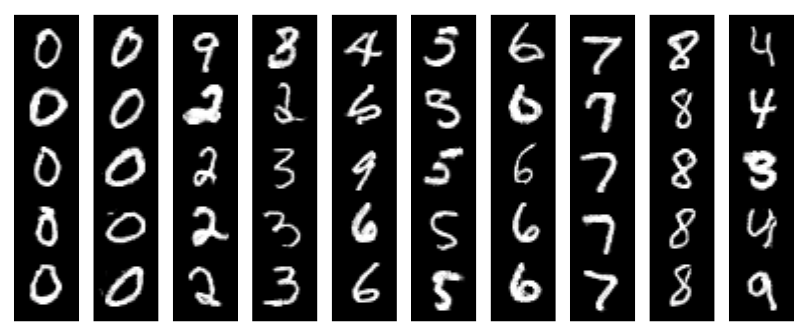} \label{fig:adda}}
    \caption{Highly confident examples for each class (0 to 9) of the baseline model and ADDA model. Additional examples are presented in App.~\ref{da-supp}. }
    \label{fig:mnist-da}
\end{figure}

But is this the whole story? To study model performance in the high-confidence range, we use \textsc{Bayes-TrEx} to generate high-confidence examples for both classifiers with the MNIST data distribution learned by GAN, as shown Fig.~\ref{fig:mnist-da}. It appears the ADDA model makes \textit{more} mistakes in these images---for example, in the 2nd column in Fig.~\ref{fig:adda}, all images where the classifier is highly confident to be 1 are actually 0s. To further study this, we hand-label 10 images per class and compute the classifier accuracy on them. Table~\ref{tab:mnist-da} shows the accuracy per digit class, as well as the overall accuracy. This analysis confirms the baseline model is more accurate than the ADDA model on these samples, suggesting that ADDA is more overconfident than the baseline. While this result does not contradict the higher overall accuracy of ADDA, it does caution against deploying such domain adaptation models without further inspection and confidence calibration assessment.

\begin{table}[t]
    \centering
    \caption{Per-digit and overall accuracy among high-confidence MNIST samples for the baseline and ADDA models. While ADDA has higher overall accuracy (0.71 vs.~0.61), it performs worse on high-confidence samples (0.72 vs.~0.80). This suggests overconfidence.}
    \resizebox{\columnwidth}{!}{
    \begin{tabular}{lrrrrrrrrrrr}
        \toprule
         & 0 & 1 & 2 & 3 & 4 & 5 & 6 & 7 & 8 & 9 & All\\
        \midrule
        Baseline & 1.0 & 0.6 & 1.0 & 0.7 & 0.5 & 0.9 & 0.9 & 0.7 & 1.0 & 0.7 & 0.80\\
        ADDA & 0.9 & 0.0 & 0.8 & 0.9 & 0.2 & 1.0 & 0.8 & 1.0 & 1.0 & 0.6 & 0.72\\
        \bottomrule
    \end{tabular}
    }
    \label{tab:mnist-da}
\end{table}

\subsection{Test-Set Comparison}
\label{sec:test-set}
Standard model evaluations are typically performed on the test set. While inspecting test set examples is not an apples-to-apples comparison for all \textsc{Bayes-TrEx} use cases (e.g. domain adaptation), we study the comparable ones. 

\subsubsection{Ambiguous Confidence}
\begin{figure}[t]
    \centering
    \includegraphics[width=0.9\columnwidth]{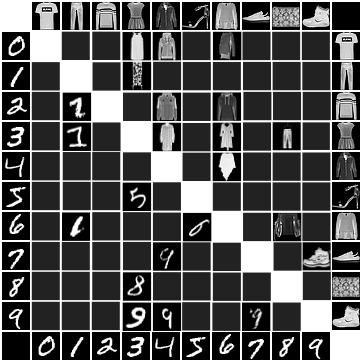}
    \caption{Test set ambiguous examples for (Fashion-) MNIST. Compared to those found by \textsc{Bayes-TrEx} in Fig.~\ref{fig:ambivalent}, test set examples have much poorer coverage. }
    \label{fig:test-set-amb}
\end{figure}

We find ambiguous examples in the (Fashion-)MNIST datasets where the classifier has confidence in $[40\%, 60\%]$ for two classes. Out of 10,000 test examples on each dataset, we find only 12 MNIST examples across 10 class pairings, and 162 Fashion-MNIST examples across 12 pairings, as shown in Fig.~\ref{fig:test-set-amb}. By comparison, \textsc{Bayes-TrEx} found ambiguous examples for 38 MNIST pairings and 28 Fashion-MNIST pairings (cf. Fig.~\ref{fig:ambivalent}). 

\subsubsection{High-Confidence Failures}
We collect and inspect highly confident test set misclassifications (confidence $\geq85\%$). For CLEVR, out of $15,000$ test images, the baseline discovers between 0 and 15 examples for each target. Notably, there are no 2-cylinder misclassifications in the test set, but \textsc{Bayes-TrEx} successful generated some (Fig.~\ref{fig:adversarial_2_cylinders}). 

From the 10,000 test examples in (Fashion-)MNIST, 84 MNIST images and 802 Fashion-MNIST images were confidently misclassified. Upon closer inspection, however, we find that the a large fraction of the failures are actually due to \textit{mislabeling}, rather than misclassification. We manually relabel all 84 MNIST misclassifications and ten Fashion-MNIST misclassifications per class, except for the trousers class which only has 3 misclassifiations. We find that the 60 out of 84 MNIST images 42 out of 93 Fashion-MNIST images are mislabeled, rather than misclassified. 

Table~\ref{tab:misclassification-num} gives detailed statistics of the number of genuinely misclassified examples. Given the scene graph data representation, all CLEVR misclassifications are genuine. Table~\ref{tab:baseline-summary} visualizes some misclassified vs.~mislabeled images, with additional classes in Appx.~\ref{test-set-supp}. Identifying mislabeled examples may be useful for correcting the dataset, but is not for our task of model understanding.

\begin{table}[!tbh]
\centering
\caption{Number of \textit{genuine} high-confidence misclassifications from the test set. Counts for CLEVR and MNIST are for the entire test set; counts for Fashion-MNIST are computed from ten random high-confidence misclassifications per class, except for trousers which only has 3 misclassifications. Fashion-MNIST classes 0-9 corresponds to T-shirt, trousers, pullover, dress, coat, sandal, shirt, sneaker, bag and ankle boot, in that order. }
\label{tab:misclassification-num}
\resizebox{\columnwidth}{!}{
\begin{tabular}{llcr}\toprule
\multirow{2}{*}{CLEVR} & class & \multirow{2}{*}{\begin{tabular}{cccc}
\phantom{x}1 Sph.\phantom{x} & \phantom{x}1 Cube\phantom{x} & \phantom{x}1 Cyl.\phantom{x} & \phantom{x}2 Cyl.\phantom{x}\vspace{0.016in}\\
5 & 8 & 15 & 0\end{tabular}} & Total\\
& count & & 28/28\\\midrule
\multirow{2}{*}{MNIST} & class & 
\multirow{4.15}{*}{\begin{tabular}{cccccccccc}
0 & 1 & 2 & 3 & 4 & 5 & 6 & 7 & 8 & 9\\
3 & 3 & 0 & 5 & 3 & 1 & 3 & 4 & 0 & 2\vspace{0.07in}\\
0 & 1 & 2 & 3 & 4 & 5 & 6 & 7 & 8 & 9\\
2 & 0 & 9 & 4 & 9 & 1 & 3 & 2 & 1 & 10\end{tabular}} & Total\\
& count & & 24/84\\\midrule
\multirow{2}{*}{Fashion} & class & & Total\\
& count & & 51/93\\\bottomrule
\end{tabular}
}
\end{table}

\begin{table}[!tbh]
\centering
\caption{High confidence misclassifications from the test set. The majority are due to incorrect ground truth labels, not classifier failures. Full table of all classes in Appx.~\ref{test-set-supp}.}
\label{tab:baseline-summary}
\resizebox{\linewidth}{!}{
\begin{tabular}{ m{1cm} m{2cm} m{5.25cm}}
\toprule
Class & Cause & Images \\  \midrule
\rule{0pt}{3ex}\multirow{2}{*}{0} 
 &  Misclassified  & 
 \includegraphics[width=0.5cm]{test_set_evaluations/MNIST/0/label_0_445.png}
 \includegraphics[width=0.5cm]{test_set_evaluations/MNIST/0/label_0_3422.png}
 \includegraphics[width=0.5cm]{test_set_evaluations/MNIST/0/label_0_2462.png} 
 \\
 & Mislabeled      & 
 \includegraphics[width=0.5cm]{test_set_evaluations/MNIST/0/label_0_2118.png}
 \includegraphics[width=0.5cm]{test_set_evaluations/MNIST/0/label_0_2896.png}
 \includegraphics[width=0.5cm]{test_set_evaluations/MNIST/0/label_0_3030.png}
 \includegraphics[width=0.5cm]{test_set_evaluations/MNIST/0/label_0_4075.png}
 \\\midrule
\rule{0pt}{3ex}\multirow{2}{*}{1} 
 & Misclassified   & 
 \includegraphics[width=0.5cm]{test_set_evaluations/MNIST/1/label_1_1260.png}
 \includegraphics[width=0.5cm]{test_set_evaluations/MNIST/1/label_1_2654.png}
 \includegraphics[width=0.5cm]{test_set_evaluations/MNIST/1/label_1_4699.png}
 \\
 & Mislabeled      & 
 \includegraphics[width=0.5cm]{test_set_evaluations/MNIST/1/label_1_2135.png}
 \includegraphics[width=0.5cm]{test_set_evaluations/MNIST/1/label_1_2387.png}
 \includegraphics[width=0.5cm]{test_set_evaluations/MNIST/1/label_1_3503.png}
 \includegraphics[width=0.5cm]{test_set_evaluations/MNIST/1/label_1_4205.png}
 \\\midrule
  \rule{0pt}{3ex}\multirow{2}{*}{2} 
 & Misclassified   & $\varnothing$
 \\
 & Mislabeled      & 
 \includegraphics[width=0.5cm]{test_set_evaluations/MNIST/2/label_2_582.png}
 \includegraphics[width=0.5cm]{test_set_evaluations/MNIST/2/label_2_1039.png}
 \includegraphics[width=0.5cm]{test_set_evaluations/MNIST/2/label_2_1226.png}
 \includegraphics[width=0.5cm]{test_set_evaluations/MNIST/2/label_2_2921.png}
 \includegraphics[width=0.5cm]{test_set_evaluations/MNIST/2/label_2_3073.png}
 \includegraphics[width=0.5cm]{test_set_evaluations/MNIST/2/label_2_3767.png}
 \includegraphics[width=0.5cm]{test_set_evaluations/MNIST/2/label_2_9009.png}
 \includegraphics[width=0.5cm]{test_set_evaluations/MNIST/2/label_2_9015.png}
 \\\midrule
 \rule{0pt}{3ex}\multirow{2}{*}{Trouser} 
 & Misclassified   & $\varnothing$
 \\
 & Mislabeled     & 
 \includegraphics[width=0.5cm]{test_set_evaluations/Fashion/1/label_1_635.png}
 \includegraphics[width=0.5cm]{test_set_evaluations/Fashion/1/label_1_3829.png}
 \includegraphics[width=0.5cm]{test_set_evaluations/Fashion/1/label_1_8621.png}  \\\midrule
  \rule{0pt}{3ex}\multirow{2}{*}{Bag} 
 & Misclassified   & 
 \includegraphics[width=0.5cm]{test_set_evaluations/Fashion/8/label_8_1645.png}
 \\
 & Mislabeled     & 
 \includegraphics[width=0.5cm]{test_set_evaluations/Fashion/8/label_8_434.png}
 \includegraphics[width=0.5cm]{test_set_evaluations/Fashion/8/label_8_661.png}
 \includegraphics[width=0.5cm]{test_set_evaluations/Fashion/8/label_8_713.png}
 \includegraphics[width=0.5cm]{test_set_evaluations/Fashion/8/label_8_930.png}
 \includegraphics[width=0.5cm]{test_set_evaluations/Fashion/8/label_8_1617.png}
 \includegraphics[width=0.5cm]{test_set_evaluations/Fashion/8/label_8_1642.png}
 \includegraphics[width=0.5cm]{test_set_evaluations/Fashion/8/label_8_2037.png}
 \includegraphics[width=0.5cm]{test_set_evaluations/Fashion/8/label_8_2279.png}
 \includegraphics[width=0.5cm]{test_set_evaluations/Fashion/8/label_8_2373.png}
                            \\\bottomrule     
\end{tabular}
}
\vspace{0.12in}
\end{table}

\subsubsection{Novel Class Extrapolation}
In Sec.~\ref{sec:ood} analysis, we find that the model mistakes some bags for ankle boots. Interestingly, this propensity is not evident from test set evaluations: the test set confusion matrix in Appx.~\ref{test-set-supp} shows that no bags are misclassified as ankle boots. This provides further evidence of the value of holistic evaluations with \textsc{Bayes-TrEx}, beyond standard test set evaluations.

\section{Discussion}

We presented \textsc{Bayes-TrEx}, a Bayesian inference approach for generating new examples that trigger various model behaviors. These examples can be further analyzed with downstream interpretability methods (Fig.~\ref{fig:saliency-map-example-intro} and Appx.~\ref{saliency-maps-supp}). To make \textsc{Bayes-TrEx} easier for model designers to use, future work should develop methods to cluster and visualize trends in the generated examples, as well as to estimate the overall coverage of the level set.

For organic data, the underlying data distributions can be learned with VAEs or GANs. These have known limitations in sample diversity~\cite{arora2017gans} and are computationally expensive to train, especially for high resolution images. In principle, \textsc{Bayes-TrEx} is agnostic to the distribution learner form and can benefit from future research in this area. Applying MCMC sampling to high dimensional latent spaces is an open problem, so \textsc{Bayes-TrEx} is currently limited to low dimensional latent spaces. 

Finally, while we analyzed only classification models with \textsc{Bayes-TrEx}, it also has the potential for analyzing structured prediction models such as machine translation or robotic control. For these domains, dependency among outputs and may need to be explicitly taken into account. We plan to extend \textsc{Bayes-TrEx} to these areas in the future.

\clearpage
\noindent \textbf{Ethics Statement.} \textsc{Bayes-TrEx} has potential to allow humans to build more accurate mental models of how neural networks make decisions. Further, \textsc{Bayes-TrEx} can be useful for debugging, interpreting, and understanding networks---all of which can help us build \emph{better}, less biased, increasingly human-aligned models. However, \textsc{Bayes-TrEx} is subject to the same caveats as typical software testing approaches: the absence of exposed bad samples does not mean the system is free from defects. One concern is how system designers and users will interact with \textsc{Bayes-TrEx} in practice. If \textsc{Bayes-TrEx} does not reveal degenerate examples, these stakeholders might develop inordinate trust~\cite{lee2004trust} in their models.

Additionally, one \textsc{Bayes-TrEx} use case is to generate examples for use with downstream local explanation methods. As a community, we know many of these methods can be challenging to understand~\cite{olah2017feature,nguyen2019understanding}, misleading~\cite{adebayo2018sanity,kindermans2019reliability,rudin2019stop}, or susceptible to adversarial attacks~\cite{slack2019can}. In human-human interaction, even nonsensical explanations can increase compliance~\cite{langer1978mindlessness}. As we build post-hoc explanation techniques, we must evaluate whether the produced explanations help humans moderate trust and act appropriately---for example, by overriding the model's decisions. 
\\\newline
\noindent \textbf{Acknowledgements.} The authors would like to thank: \href{http://alexlew.net/}{Alex Lew}, \href{https://www.mct.dev/}{Marco Cusumano-Towner}, and \href{https://www.researchgate.net/profile/Tan_Zhi-Xuan}{Tan Zhi-Xuan} for their insights on how to formulate this inference problem and use probabilistic programming effectively; Christian Muise and Hendrik Strobelt for helpful early discussions; and James Tompkin and the anonymous reviewers for comments on the draft. SB is supported by an NSF GRFP.

\bibliography{bibliography}

\clearpage
\onecolumn
\appendix

\begin{minipage}[t][0.75cm][b]{0,5\textwidth}
\end{minipage}
\section*{\textsc{Bayes-TrEx} -- Appendix}
\begin{minipage}[t][0.5cm][b]{0,5\textwidth}
\end{minipage}

\contentsline {section}{\numberline {A}Network Architecture for MNIST \& Fashion-MNIST}{}{}%
\contentsline {section}{\numberline {B}Fr\'echet Inception Distance (FID) for VAE and GAN}{}{}%
\contentsline {section}{\numberline {C}Quantitative Prediction Confidence Summary}{}{}%
\contentsline {section}{\numberline {D}High-Confidence Examples}{}{}%
\contentsline {section}{\numberline {E}Ambiguous Confidence Examples}{}{}%
\contentsline {section}{\numberline {F}Ambiguous Confidence with GAN and Modified Classifier}{}{}%
\contentsline {section}{\numberline {G}High-Confidence Failure Analysis}{}{}%
\contentsline {section}{\numberline {H}Novel Class Extrapolation Analysis}{}{}%
\contentsline {section}{\numberline {I}Domain Adaptation Analysis}{}{}%
\contentsline {section}{\numberline {J}Test Set Evaluation}{}{}%
\contentsline {section}{\numberline {K}\textsc{Bayes-TrEx} with Saliency Maps}{}{}%
\begin{minipage}[t][0.75cm][b]{0,5\textwidth}
\end{minipage}


\clearpage
\section{Network Architecture for MNIST \& Fashion-MNIST}
\label{arch-supp}

For all experiments on MNIST and Fashion-MNIST, the VAE architecture is shown in Table \ref{tab:generator} (left), and the GAN architecture is shown in Table \ref{tab:generator} (right). For all experiments on MNIST and Fashion-MNIST except for the domain adaptation analysis, the classifier architecture is shown in Table \ref{tab:classifier} (left). The classifier used in the domain adaptation analysis is the LeNet architecture, following the provided source code, shown in Table \ref{tab:classifier} (right). VAEs and GANs are trained with binary cross entropy loss. Classifiers are trained with negative log likelihood loss. 

\begin{table}[htbp]
    \caption{Left: VAE architecture; right: GAN architecture. }
    \centering
    \begin{tabular}{c}\toprule
         Encoder input: $28\times 28\times 1$ \\\midrule
         Flatten\\\midrule
         Fully-connected $784\times 400$\\\midrule
         ReLU\\\midrule
         Mean: Fully-connected $400\times 5$\vspace{0.07in}\\
         Log-variance: Fully-connected $400\times 5$\\\midrule
         Decoder input: 5 (latent dimension)\\\midrule
         Fully-connected $5\times 400$\\\midrule
         ReLU\\\midrule
         Fully-connected $400\times 784$\\\midrule
         Reshape $28\times 28\times 1$\\\midrule
         Sigmoid\\\bottomrule
    \end{tabular}\hspace{0.1in}
    \begin{tabular}{c}\toprule
         Input: 5 (latent dimension) \\\midrule
         Reshape $1\times 1\times 5$\\\midrule
         Conv-transpose: 512 filters, size=$4\times 4$, stride = 1\\\midrule
         Batch-norm, ReLU\\\midrule
         Conv-transpose: 256 filters, size=$4\times 4$, stride = 2\\\midrule
         Batch-norm, ReLU\\\midrule
         Conv-transpose: 128 filters, size=$4\times 4$, stride = 2\\\midrule
         Batch-norm, ReLU\\\midrule
         Conv-transpose: 64 filters, size=$4\times 4$, stride = 2\\\midrule
         Batch-norm, ReLU\\\midrule
         Conv-transpose: 1 filters, size=$1\times 1$, stride = 1\\\midrule
         Sigmoid\\\bottomrule
    \end{tabular}
    \label{tab:generator}
\end{table}

\begin{table}[htbp]
    \caption{Left: classifier architecture in all experiments except domain adaptation analysis; right: LeNet classifier architecture in domain adaptation analysis (used in code released by ADDA authors). }
    \centering
    \begin{tabular}{c}\toprule
         Input: $28\times 28\times 1$ \\\midrule
         Conv: 32 filters, size = $3\times 3$, stride = 1\\\midrule
         ReLU\\\midrule
         Conv: 64 filters, size = $3\times 3$, stride = 1\\\midrule
         Drop-out, prob = 0.25\\\midrule
         Max-pool, size = $2\times 2$\\\midrule
         Flatten\\\midrule
         Fully-connected $9216\times 128$\\\midrule
         ReLU\\\midrule
         Drop-out, prob = 0.5\\\midrule
         Fully-connected $128\times 10$\\\midrule
         Soft-max\\\bottomrule
    \end{tabular}\hspace{0.1in}
    \begin{tabular}{c}\toprule
         Input: $28\times 28\times 1$ \\\midrule
         Conv: 20 filters, size = $5\times 5$, stride = 1\\\midrule
         ReLU\\\midrule
         Max-pool, size = $2\times 2$\\\midrule
         Conv: 50 filters, size = $5\times 5$, stride = 1\\\midrule
         ReLU\\\midrule
         Max-pool, size = $2\times 2$\\\midrule
         Flatten\\\midrule
         Fully-connected $800\times 500$\\\midrule
         ReLU\\\midrule
         Fully-connected $500\times 10$\\\midrule
         Soft-max\\\bottomrule
    \end{tabular}
    \label{tab:classifier}
\end{table}

\clearpage
\section{Fr\'echet Inception Distance (FID) for VAE and GAN}
\label{fid-supp}

Table~\ref{tab:fid-scores} extends Table~\ref{tab:fid-main} in Sec.~\ref{sec:dataset-setup} and lists the FID scores for all VAE and GAN models that we use. These FID scores reveal the GANs are better approximations of the underlying data distributions. Models trained on ``all'' data are used for high confidence, ambiguous confidence, confidence interpolation and domain adaptation settings. Models trained on data ``without [class]'' are used for high-confidence failure settings. Models trained on select classes (\{2, 4, 5, 7, 8\} and \{0, 1, 4, 7, 8\}) are used for the novel class extrapolation settings. 

\begin{table}[htbp]
\centering
\caption{Fr\'echet Inception Distance (FID) scores for all learned data distributions; a lower score indicates a better distribution fit. Results are computed across 1000 samples. Classes 0 to 9 for Fashion-MNIST correspond to 0: T-shirt, 1: Trouser, 2: Pullover, 3: Dress, 4: Coat,  5: Sandal, 6: Shirt, 7: Sneaker, 8: Bag, and 9: Ankle boot.}
\begin{tabular}{ lllr }
\toprule
Model & Dataset & Data Source & FID \\  \midrule

\multirow{24}{*}{GAN} & \multirow{12}{*}{MNIST}
 & All & 11.83\\
& & Without 0 & 12.10\\
& & Without 1 & 12.08\\
& & Without 2 & 13.57\\
& & Without 3 & 12.71\\
& & Without 4 & 12.25\\
& & Without 5 & 12.21\\
& & Without 6 & 11.86\\
& & Without 7 & 11.64\\
& & Without 8 & 12.31\\
& & Without 9 & 12.34\\
& & \{2, 4, 5, 7, 8\} & 13.45\\\cmidrule{2-4}

& \multirow{12}{*}{Fashion}     
& All & 29.44\\
& & Without 0 & 28.91\\
& & Without 1 & 31.18\\
& & Without 2 & 30.11\\
& & Without 3 & 28.95\\
& & Without 4 & 30.43\\
& & Without 5 & 27.67\\
& & Without 6 & 29.68\\
& & Without 7 & 28.56\\
& & Without 8 & 30.87\\
& & Without 9 & 29.22\\
& & \{0, 1, 4, 7, 8\} & 33.11\\\bottomrule
\end{tabular} \hspace{4mm}
\begin{tabular}{ lllr }
\toprule
Model & Dataset & Data Source & FID \\  \midrule
\multirow{24}{*}{VAE} & \multirow{12}{*}{MNIST}
 & All & 72.33\\
& & Without 0 & 71.28\\
& & Without 1 & 75.36\\
& & Without 2 & 64.77\\
& & Without 3 & 63.66\\
& & Without 4 & 66.96\\
& & Without 5 & 63.31\\
& & Without 6 & 67.64\\
& & Without 7 & 62.45\\
& & Without 8 & 64.14\\
& & Without 9 & 66.57\\
& & ----- & ----- \\\cmidrule{2-4}
& \multirow{12}{*}{Fashion}     
 & All & 87.89\\
& & Without 0 & 89.21\\
& & Without 1 & 92.02\\
& & Without 2 & 91.20\\
& & Without 3 & 85.51\\
& & Without 4 & 88.38\\
& & Without 5 & 84.17\\
& & Without 6 & 85.58\\
& & Without 7 & 84.93\\
& & Without 8 & 83.66\\
& & Without 9 & 81.48\\
& & --- & --- \\
\bottomrule
\end{tabular}
\label{tab:fid-scores}
\end{table}

\clearpage
\section{Quantitative Prediction Confidence Summary}
\label{full-table-supp}

Tables \ref{high-conf-adv-full-table-supp}, \ref{graded-conf-full-table-supp}, and \ref{ood-full-table-supp} present the extension of Table \ref{tab:quant-conf} in Sec.~\ref{sec:quant}. These results show that the inferred samples have predicted confidence closely matching the specified confidence targets. This indicates the MCMC methods used by \textsc{Bayes-TrEx} are successful for the tested domains and scenarios. Queries for 5 Cubes in the novel class extrapolation CLEVR experiments use a stopping criterion of 1500 samples instead of the standard 500. Averages reported across 10 inference runs.

\begin{table}[!htb]
\centering
\caption{Prediction confidence for samples on high-confidence examples (left) and high confidence misclassifications (right). }
\begin{tabular}[t]{lr}\toprule
Target & Prediction Confidence \\  \midrule
$\vec p_0 = 1$ & 0.999 $\pm$ 0.006 \\
$\vec p_1 = 1$ & 0.999 $\pm$ 0.003 \\
$\vec p_2 = 1$ & 0.999 $\pm$ 0.006 \\
$\vec p_3 = 1$ & 0.999 $\pm$ 0.005 \\
$\vec p_4 = 1$ & 0.998 $\pm$ 0.008 \\
$\vec p_5 = 1$ & 0.999 $\pm$ 0.006 \\
$\vec p_6 = 1$ & 0.998 $\pm$ 0.007 \\
$\vec p_7 = 1$ & 0.998 $\pm$ 0.007 \\
$\vec p_8 = 1$ & 0.999 $\pm$ 0.004 \\
$\vec p_9 = 1$ & 0.998 $\pm$ 0.007 \\\midrule
$\vec p_\text{T-Shirt} = 1$ & 0.991 $\pm$ 0.016 \\
$\vec p_\text{Trouser} = 1$ & 0.999 $\pm$ 0.006 \\
$\vec p_\text{Pullover} = 1$ & 0.984 $\pm$ 0.019 \\
$\vec p_\text{Dress} = 1$ & 0.993 $\pm$ 0.008 \\
$\vec p_\text{Coat} = 1$ & 0.983 $\pm$ 0.021 \\
$\vec p_\text{Sandal} = 1$ & 0.998 $\pm$ 0.008 \\
$\vec p_\text{Shirt} = 1$ & 0.987 $\pm$ 0.020 \\
$\vec p_\text{Sneaker} = 1$ & 0.994 $\pm$ 0.016 \\
$\vec p_\text{Bag} = 1$ & 0.999 $\pm$ 0.006 \\
$\vec p_\text{Ankle Boot} = 1$ & 0.996 $\pm$ 0.012 \\\midrule
$\vec p_\text{5 Spheres} = 1$ & 0.943 $\pm$ 0.020 \\
$\vec p_\text{2 Blue Spheres} = 1$ & 0.892 $\pm$ 0.245 \\\bottomrule
\end{tabular}
\begin{tabular}[t]{lr}\toprule
Target & Prediction Confidence \\  \midrule
$\vec p_0 = 1$ &  0.981 $\pm$ 0.027 \\
$\vec p_1 = 1$ &  0.953 $\pm$ 0.028 \\
$\vec p_2 = 1$ &  0.968 $\pm$ 0.028 \\
$\vec p_3 = 1$ &  0.969 $\pm$ 0.027 \\
$\vec p_4 = 1$ &  0.955 $\pm$ 0.030 \\
$\vec p_5 = 1$ &  0.990 $\pm$ 0.018 \\
$\vec p_6 = 1$ &  0.970 $\pm$ 0.026 \\
$\vec p_7 = 1$ &  0.968 $\pm$ 0.029 \\
$\vec p_8 = 1$ &  0.982 $\pm$ 0.024 \\
$\vec p_9 = 1$ &  0.983 $\pm$ 0.022 \\\midrule
$\vec p_\text{T-Shirt} = 1$ & 0.964 $\pm$ 0.029 \\
$\vec p_\text{Trouser} = 1$ & (sample failure) \\
$\vec p_\text{Pullover} = 1$ & 0.886 $\pm$ 0.027 \\
$\vec p_\text{Dress} = 1$ & 0.970 $\pm$ 0.026 \\
$\vec p_\text{Coat} = 1$ & 0.938 $\pm$ 0.030 \\
$\vec p_\text{Sandal} = 1$ & 0.968 $\pm$ 0.030 \\
$\vec p_\text{Shirt} = 1$ & 0.938 $\pm$ 0.032 \\
$\vec p_\text{Sneaker} = 1$ & 0.969 $\pm$ 0.028 \\
$\vec p_\text{Bag} = 1$ & 0.967 $\pm$ 0.026 \\
$\vec p_\text{Ankle Boot} = 1$ & 0.971 $\pm$ 0.027 \\\midrule
$\vec p_\text{1 Cube} = 1$ & 0.929 $\pm$ 0.062 \\
$\vec p_\text{1 Cylinder} = 1$ & 0.972 $\pm$ 0.021 \\
$\vec p_\text{1 Sphere} = 1$ & 0.843 $\pm$ 0.266 \\
$\vec p_\text{2 Cylinders} = 1$ & 0.545 $\pm$ 0.230 \\\bottomrule
\end{tabular}
\label{high-conf-adv-full-table-supp}
\end{table}

\begin{table}[!htb]
\centering
\caption{(Fashion-)MNIST confidence interpolation.}
\resizebox{0.46\textwidth}{!}{
\begin{tabular}{lr}\toprule
Target & Prediction Confidence \\  \midrule
$\vec p_{8}=0.0, \vec p_{9}=1.0$ & $(0.002 \pm 0.006, 0.990 \pm 0.016)$ \\
$\vec p_{8}=0.1, \vec p_{9}=0.9$ & $(0.030 \pm 0.039, 0.936 \pm 0.051)$ \\
$\vec p_{8}=0.2, \vec p_{9}=0.8$ & $(0.170 \pm 0.039, 0.788 \pm 0.040)$ \\
$\vec p_{8}=0.3, \vec p_{9}=0.7$ & $(0.275 \pm 0.041, 0.682 \pm 0.040)$ \\
$\vec p_{8}=0.4, \vec p_{9}=0.6$ & $(0.378 \pm 0.040, 0.578 \pm 0.040)$ \\
$\vec p_{8}=0.5, \vec p_{9}=0.5$ & $(0.477 \pm 0.039, 0.477 \pm 0.039)$ \\
$\vec p_{8}=0.6, \vec p_{9}=0.4$ & $(0.581 \pm 0.038, 0.374 \pm 0.039)$ \\
$\vec p_{8}=0.7, \vec p_{9}=0.3$ & $(0.680 \pm 0.041, 0.275 \pm 0.039)$ \\
$\vec p_{8}=0.8, \vec p_{9}=0.2$ & $(0.788 \pm 0.040, 0.167 \pm 0.041)$ \\
$\vec p_{8}=0.9, \vec p_{9}=0.1$ & $(0.926 \pm 0.050, 0.039 \pm 0.040)$ \\
$\vec p_{8}=1.0, \vec p_{9}=0.0$ & $(0.989 \pm 0.016, 0.002 \pm 0.007)$ \\\bottomrule
\end{tabular}
}
\resizebox{0.53\textwidth}{!}{
\begin{tabular}{lr}\toprule
Target & Prediction Confidence \\  \midrule
$\vec p_{\text{T-Shirt}}=0.0, \vec p_{\text{Trousers}}=1.0$ & $(0.001 \pm 0.004, 0.995 \pm 0.012)$ \\
$\vec p_{\text{T-Shirt}}=0.1, \vec p_{\text{Trousers}}=0.9$ & $(0.026 \pm 0.035, 0.950 \pm 0.050)$ \\
$\vec p_{\text{T-Shirt}}=0.2, \vec p_{\text{Trousers}}=0.8$ & $(0.166 \pm 0.040, 0.791 \pm 0.041)$ \\
$\vec p_{\text{T-Shirt}}=0.3, \vec p_{\text{Trousers}}=0.7$ & $(0.275 \pm 0.037, 0.686 \pm 0.038)$ \\
$\vec p_{\text{T-Shirt}}=0.4, \vec p_{\text{Trousers}}=0.6$ & $(0.379 \pm 0.038, 0.586 \pm 0.038)$ \\
$\vec p_{\text{T-Shirt}}=0.5, \vec p_{\text{Trousers}}=0.5$ & $(0.436 \pm 0.040, 0.459 \pm 0.040)$ \\
$\vec p_{\text{T-Shirt}}=0.6, \vec p_{\text{Trousers}}=0.4$ & $(0.583 \pm 0.038, 0.382 \pm 0.037)$ \\
$\vec p_{\text{T-Shirt}}=0.7, \vec p_{\text{Trousers}}=0.3$ & $(0.685 \pm 0.039, 0.281 \pm 0.040)$ \\
$\vec p_{\text{T-Shirt}}=0.8, \vec p_{\text{Trousers}}=0.2$ & $(0.790 \pm 0.037, 0.177 \pm 0.037)$ \\
$\vec p_{\text{T-Shirt}}=0.9, \vec p_{\text{Trousers}}=0.1$ & $(0.936 \pm 0.045, 0.029 \pm 0.041)$ \\
$\vec p_{\text{T-Shirt}}=1.0, \vec p_{\text{Trousers}}=0.0$ & $(0.985 \pm 0.019, 0.000 \pm 0.003)$ \\\bottomrule
\end{tabular}
}\label{graded-conf-full-table-supp}

\end{table}

\begin{table}[!htb]
\centering
\caption{Prediction confidence for novel class extrapolation (left) and domain adaptation (right).}
\begin{tabular}[t]{lr} \toprule
Target & Prediction Confidence \\  \midrule
$\vec p_0 = 1$ & 0.976 $\pm$ 0.025 \\
$\vec p_1 = 1$ & 0.988 $\pm$ 0.186 \\
$\vec p_3 = 1$ & 0.987 $\pm$ 0.020 \\
$\vec p_6 = 1$ & 0.989 $\pm$ 0.018 \\
$\vec p_9 = 1$ & 0.995 $\pm$ 0.013 \\\midrule
$\vec p_\text{Pullover} = 1$ & 0.991 $\pm$ 0.016 \\
$\vec p_\text{Dress} = 1$ & 0.994 $\pm$ 0.013 \\
$\vec p_\text{Sandal} = 1$ & 0.995 $\pm$ 0.013 \\
$\vec p_\text{Shirt} = 1$ & 0.994 $\pm$ 0.012 \\
$\vec p_\text{Ankle Boot} = 1$ & 0.993 $\pm$ 0.015 \\\midrule
$\vec p_\text{1 Cube} = 1$ & 0.983 $\pm$ 0.014 \\
$\vec p_\text{1 Cylinder} = 1$ & 0.959 $\pm$ 0.031 \\
$\vec p_\text{1 Sphere} = 1$ & 0.969 $\pm$ 0.022 \\
$\vec p_\text{5 Cubes} = 1$ & 0.921 $\pm$ 0.029 \\\bottomrule
\end{tabular}
\begin{tabular}[t]{lr} \toprule
Target & Prediction Confidence \\ \midrule
$\vec p_0 = 1$ &  0.996 $\pm$ 0.011 \\
$\vec p_1 = 1$ &  0.994 $\pm$ 0.014 \\
$\vec p_2 = 1$ &  0.998 $\pm$ 0.008 \\
$\vec p_3 = 1$ &  0.994 $\pm$ 0.015 \\
$\vec p_4 = 1$ &  0.997 $\pm$ 0.010 \\
$\vec p_5 = 1$ &  0.998 $\pm$ 0.007 \\
$\vec p_6 = 1$ &  0.996 $\pm$ 0.011 \\
$\vec p_7 = 1$ &  0.996 $\pm$ 0.011 \\
$\vec p_8 = 1$ &  0.995 $\pm$ 0.013 \\
$\vec p_9 = 1$ &  0.996 $\pm$ 0.012 \\\bottomrule
\end{tabular}
\label{ood-full-table-supp}
\end{table}

\begin{figure}[!htb]
    \centering
    \includegraphics[width=0.95\columnwidth]{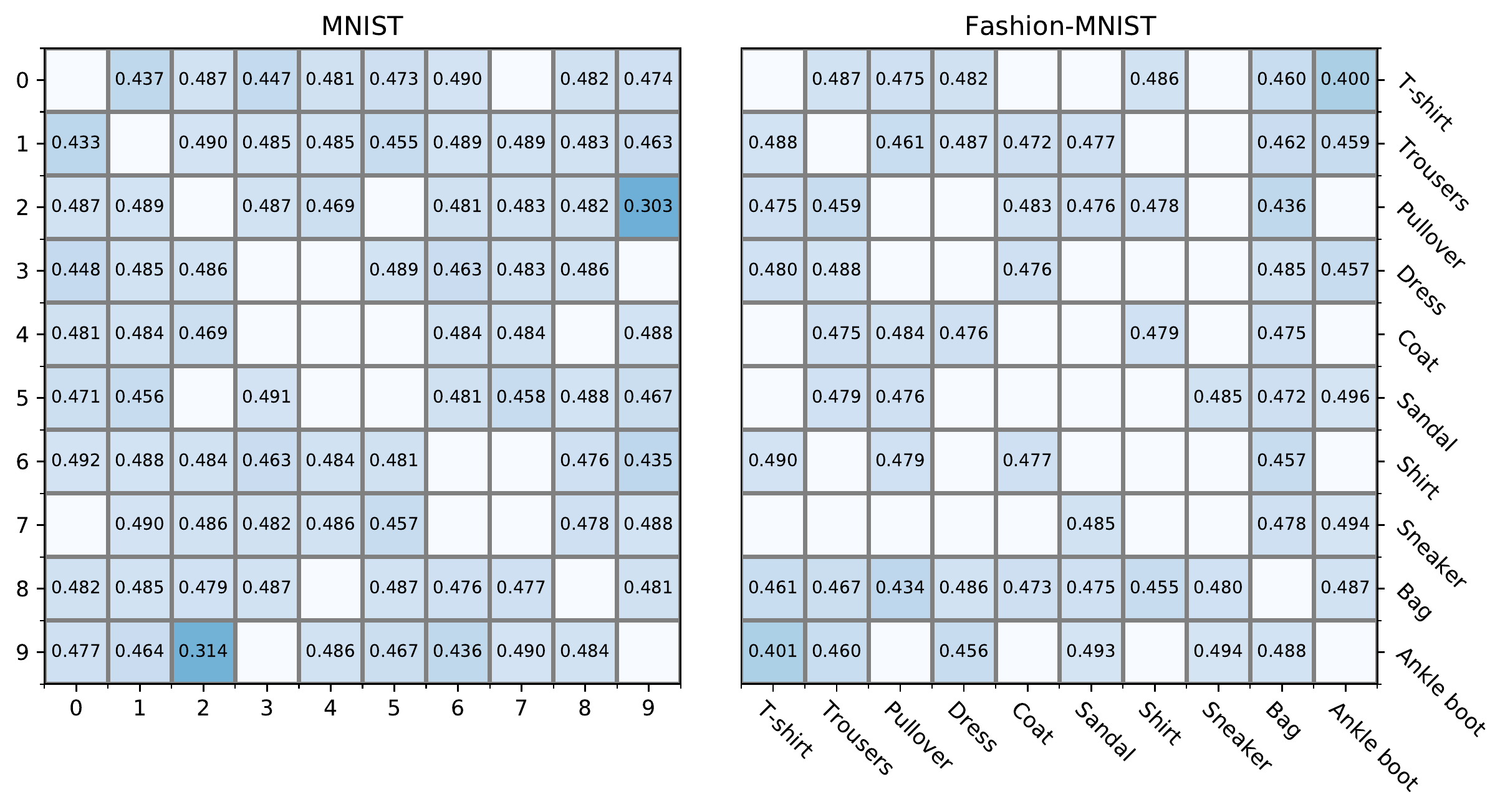}
    \caption{Prediction confidence for (Fashion-)MNIST ambiguous examples. For each class combination, the lower triangle shows the the confidence for the digit denoted on the horizontal axis, and the upper triangle shows the confidence for the digit on the vertical axis. For example, for the MNIST class combination $9$v$0$, the classifier confidence in class 0 is 0.477 (the bottom left entry) while the classifier confidence in class 9 is 0.474 (the top right entry). Diagonal entries are blank since they have the same class on row and column. Off-diagonal blank entries indicate that \textsc{Bayes-TrEx} does not find ambiguous samples for that particular class pair. }
    \label{fig:ambiguous_image_target_accuracy}
\end{figure}

\FloatBarrier
\phantom{x}

\clearpage
\section{High-Confidence Examples}
\label{high-conf-supp}
\begin{figure}[!htb]
    \centering
    \subfigure[$P_\text{5 Sph.} = 94.8\%$]{
    \includegraphics[width=0.18\columnwidth]{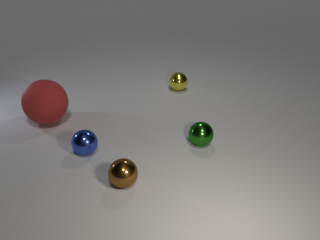}
    \label{first_high_conf_5_spheres}}
    \subfigure[$P_\text{5 Sph.} = 94.5\%$]{
    \includegraphics[width=0.18\columnwidth]{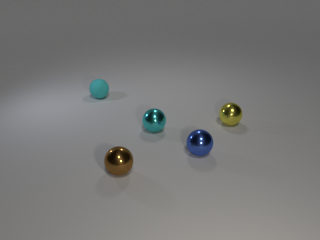}
    }
    \subfigure[$P_\text{5 Sph.} = 94.6\%$]{
    \includegraphics[width=0.18\columnwidth]{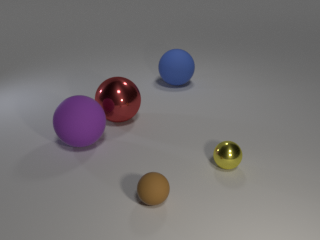}
    }
    \subfigure[$P_\text{5 Sph.} = 95.2\%$]{
    \includegraphics[width=0.18\columnwidth]{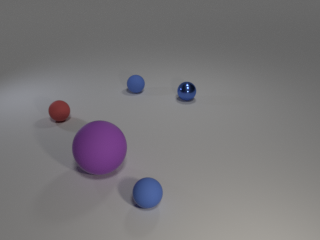}
    }
    \subfigure[$P_\text{5 Sph.} = 92.0\%$]{
    \includegraphics[width=0.18\columnwidth]{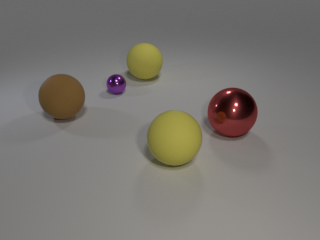}
    \label{last_high_conf_5_spheres}
    }

    \subfigure[$P_\text{2 Blue} = 96.3\%$]{
    \includegraphics[width=0.18\columnwidth]{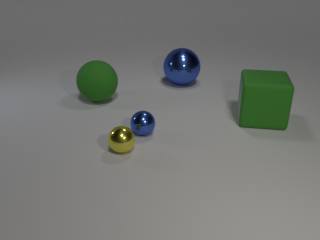}
    \label{first_high_conf_2_blue}
    }
    \subfigure[$P_\text{2 Blue} = 96.1\%$]{
    \includegraphics[width=0.18\columnwidth]{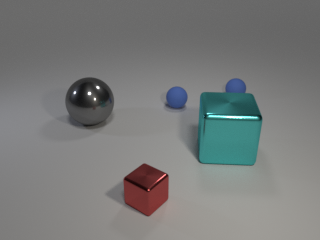}
    }
    \subfigure[$P_\text{2 Blue} = 94.9\%$]{
    \includegraphics[width=0.18\columnwidth]{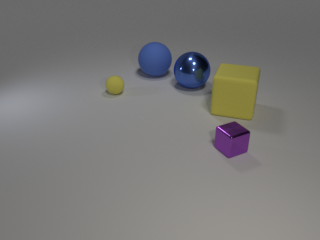}
    }
    \subfigure[$P_\text{2 Blue} = 96.8\%$]{
    \includegraphics[width=0.18\columnwidth]{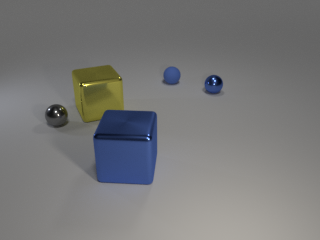}
    }
    \subfigure[$P_\text{2 Blue} = 97.8\%$]{
    \includegraphics[width=0.18\columnwidth]{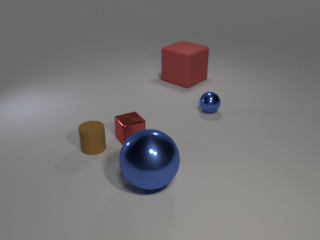}
    \label{last_high_conf_2_blue}
    }
    \caption{Above, \ref{first_high_conf_5_spheres}--\ref{last_high_conf_5_spheres}: selected examples classified as containing 5 spheres with high confidence. Below, \ref{first_high_conf_2_blue}--\ref{last_high_conf_2_blue}: selected examples classified as containing 2 blue spheres with high confidence.}
    \label{fig:high-confidence-clevr}
\end{figure}

Figure \ref{fig:high-confidence-clevr} presents additional high-confidence CLEVR examples and the classifier's predictions. 

\begin{figure}[!htb]
    \centering
    \subfigure[MNIST]{
    \includegraphics[width=0.48\columnwidth]{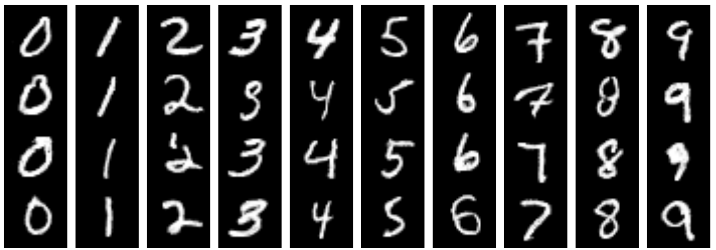}
    }
    \subfigure[Fashion-MNIST]{
    \includegraphics[width=0.48\columnwidth]{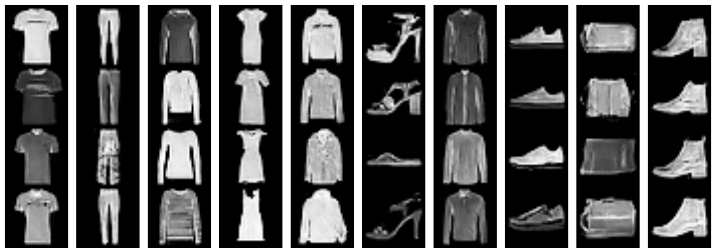}
    }

    \caption{High-confidence examples from MNIST and Fashion-MNIST. There are no misclassifications. MNIST columns represent digit 0 to 9, respectively. Fashion-MNIST columns represent T-shirt, trousers, pullover, dress, coat, sandal, shirt, sneaker, bag, and ankle boot, respectively.}
    \label{fig:high-confidence-mnist}
\end{figure}

Figure \ref{fig:high-confidence-mnist} presents additional high-confidence examples for MNIST and Fashion-MNIST. 

\clearpage
\section{Ambiguous Confidence Examples}
\label{ambivalent-supp}

\begin{figure}[!htb]
    \centering
    \includegraphics[width=\columnwidth]{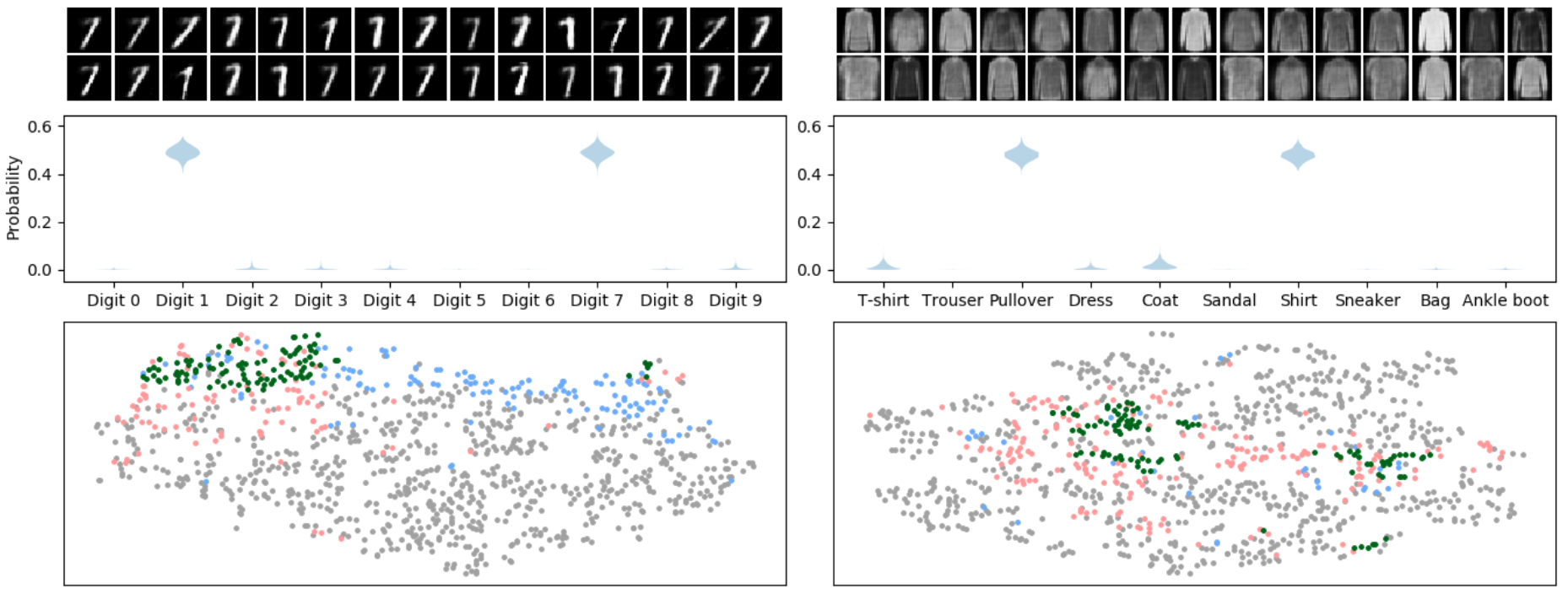}
    \caption{Left: ambiguous samples for digit 1 vs.~7 in MNIST. Right: ambiguous samples for pullover vs.~shirt in Fashion-MNIST. Top: 30 sampled images. Middle: classifier confidence plots on the samples. Bottom: $t$-SNE latent space visualization: green dots represent ambiguous samples from the posterior, red and blue dots represents samples from the prior that are predicted by the classifier to be either class of interest, and gray dots represents other samples from the prior. The ambiguous samples are on the class boundaries.}
    \label{fig:ambivalent-image-detail-supp}
\end{figure}

Figure \ref{fig:ambivalent-image-detail-supp} presents additional visualizations for two pairs, Digit 1 vs.~Digit 7 from MNIST and T-shirt vs.~Pullover from Fashion-MNIST. The confidence plots in the middle confirm that the neural network is indeed making the ambiguous predictions. The $t$-SNE \cite{maaten2008visualizing} latent space visualizations at the bottom indicate that the samples lie around the class boundaries and are also in-distribution (i.e., having close proximity to those sampled from the prior). 

In addition, we also sampled for uniformly ambiguous examples (i.e. images that receive around 10\% confidence for each class) using the following formulation:
\begin{align}
    u|\vec x &\sim \mathrm{No}(\max_i f(\vec x)_i - \min_j f(\vec x)_j, \sigma^2), \\
    u^*&=0. 
\end{align}

\begin{figure}[!htb]
    \centering
    \includegraphics[width=0.8\columnwidth]{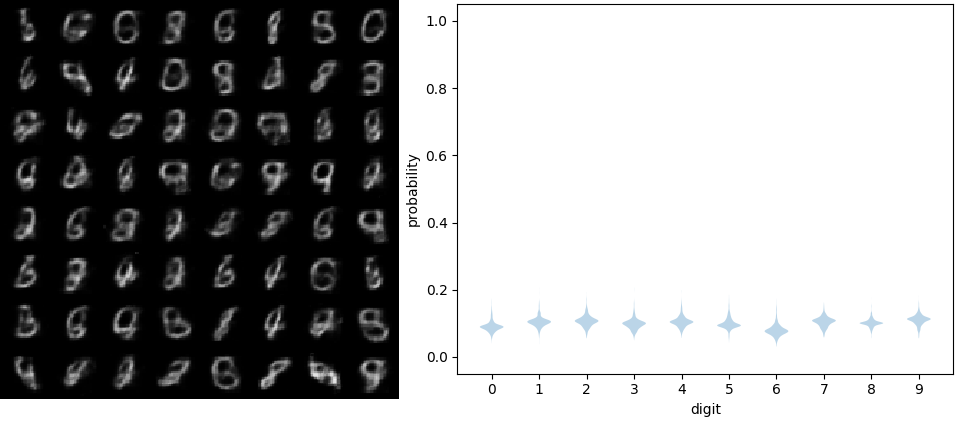}
    \caption{Uniformly ambiguous images and the confidence plot. }
    \label{fig:uniform-ambigu}
\end{figure}

Fig.~\ref{fig:uniform-ambigu} shows these samples and their confidence plot. 

\clearpage
\section{Ambiguous Confidence with GAN and Modified Classifier}
\label{gan-failure-supp}

\begin{figure}[!htb]
    \centering
    \includegraphics[width=0.19\columnwidth]{gan_failure/gan_amb_0v1_img.png}
    \includegraphics[width=0.19\columnwidth]{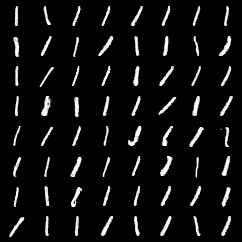}
    \includegraphics[width=0.19\columnwidth]{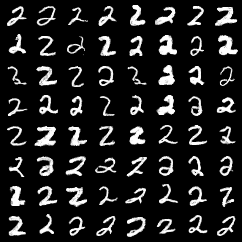}
    \includegraphics[width=0.19\columnwidth]{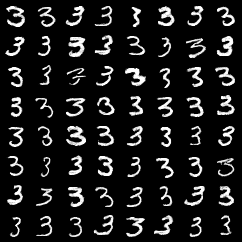}
    \includegraphics[width=0.19\columnwidth]{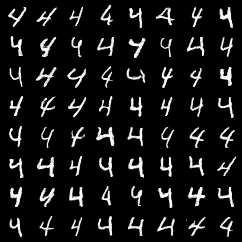}
    
    \includegraphics[width=0.19\columnwidth]{gan_failure/gan_amb_0v1_violin.png}
    \includegraphics[width=0.19\columnwidth]{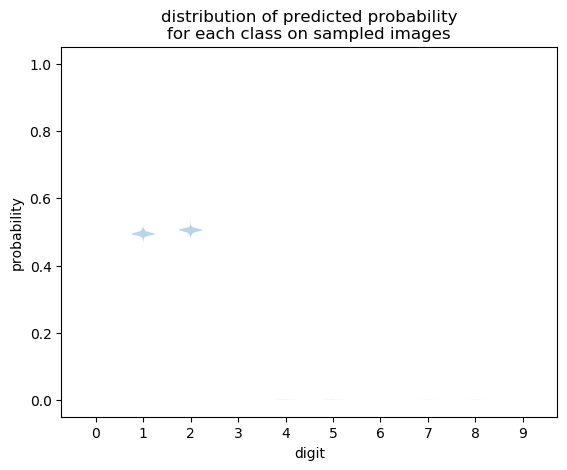}
    \includegraphics[width=0.19\columnwidth]{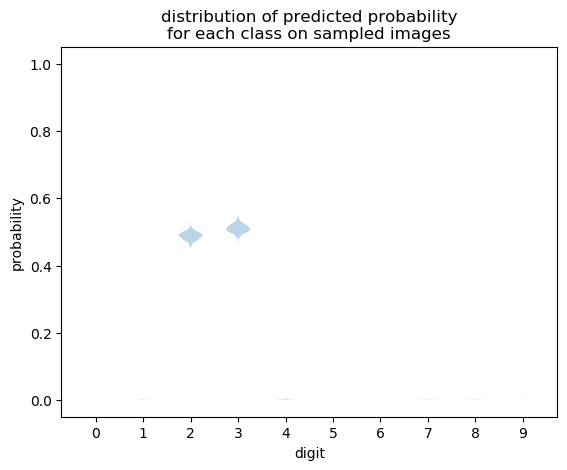}
    \includegraphics[width=0.19\columnwidth]{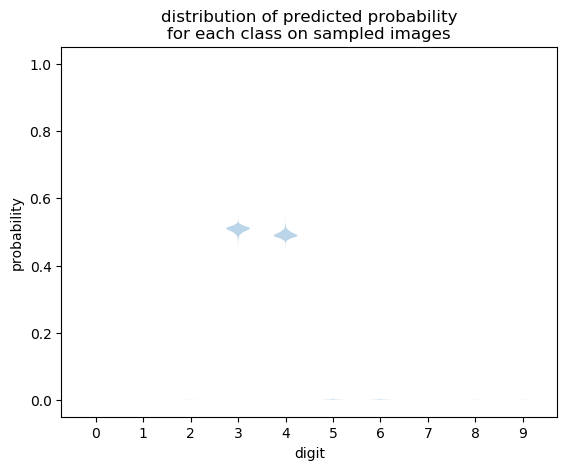}
    \includegraphics[width=0.19\columnwidth]{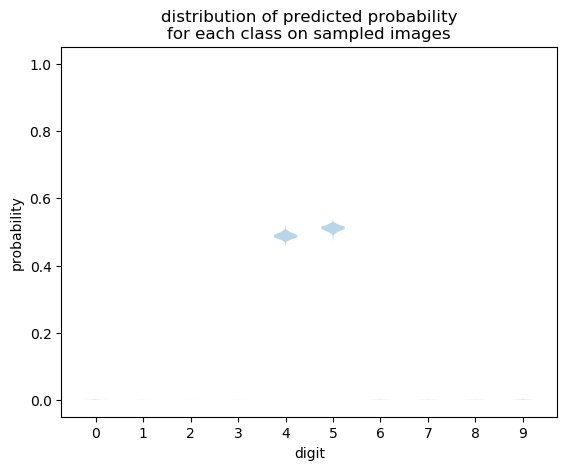}
    
    \includegraphics[width=0.19\columnwidth]{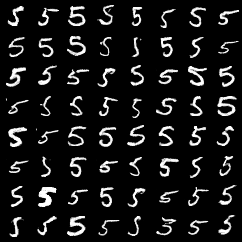}
    \includegraphics[width=0.19\columnwidth]{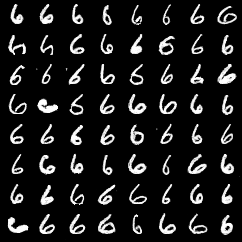}
    \includegraphics[width=0.19\columnwidth]{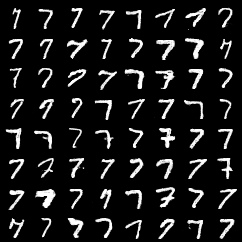}
    \includegraphics[width=0.19\columnwidth]{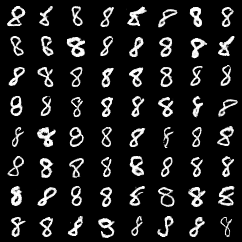}
    \includegraphics[width=0.19\columnwidth]{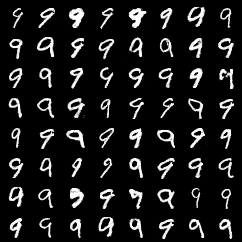}
    
    \includegraphics[width=0.19\columnwidth]{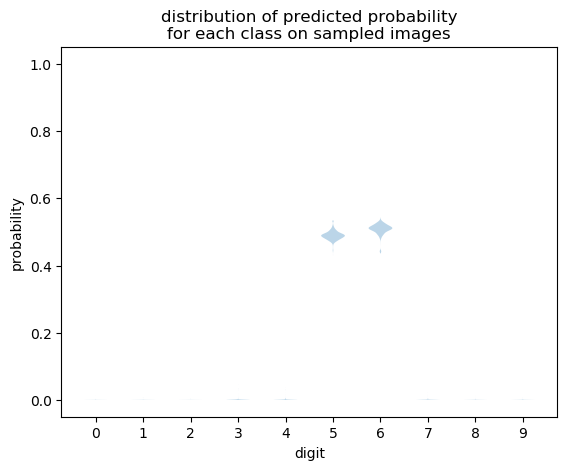}
    \includegraphics[width=0.19\columnwidth]{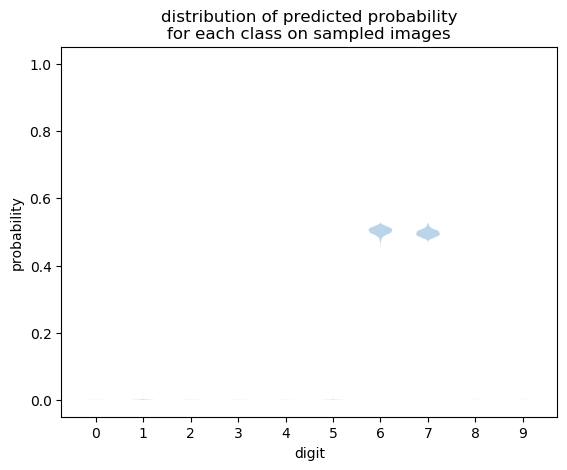}
    \includegraphics[width=0.19\columnwidth]{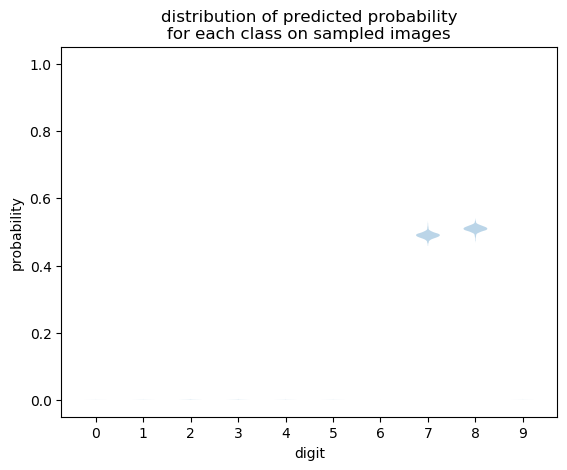}
    \includegraphics[width=0.19\columnwidth]{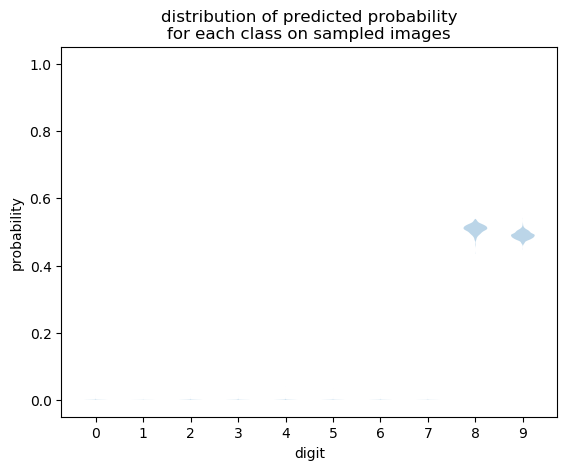}
    \includegraphics[width=0.19\columnwidth]{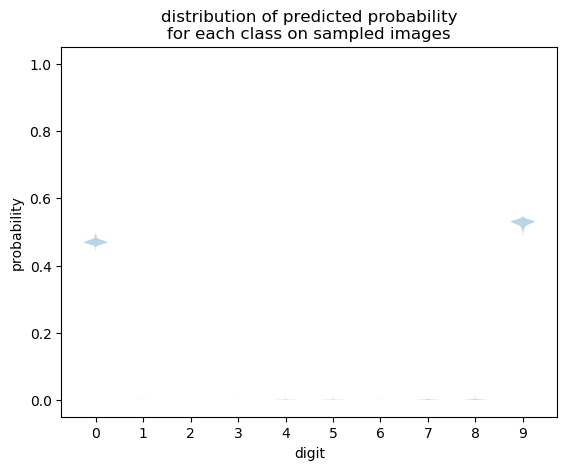}
    \caption{Sampling results with an explicitly ambivalent classifier and a GAN-learned distribution. Top 2 rows: digit $i$ vs.~$i+1$ for $i\in\{0, 1, 2, 3, 4\}$. Bottom 2 rows: digit $i$ vs.~$i+1$ (mod 10) for $i\in\{5, 6, 7, 8, 9\}$. }
    \label{fig:gan-failure-supp}
\end{figure}

Fig.~\ref{fig:gan-failure-supp} shows the ambiguous confidence samples for 0v1, 1v2, ..., 9v0 using the GAN-learned distribution when the classifier is trained with the custom KL loss described in Eq.~\ref{eq:kl}. 

\clearpage

\clearpage
\section{High-Confidence Failure Analysis}
\label{adv-supp}
\begin{figure}[!htb]
    \centering
    \subfigure[$\vec p_\text{1 Cube} =96.0\%$]{
    \includegraphics[width=0.18\columnwidth]{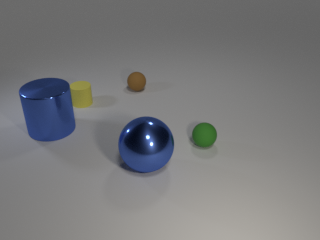}}
    \subfigure[$\vec p_\text{1 Cube} =97.2\%$]{
    \includegraphics[width=0.18\columnwidth]{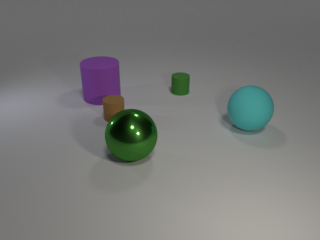}
    }
    \subfigure[$\vec p_\text{1 Cube} =93.5\%$]{
    \includegraphics[width=0.18\columnwidth]{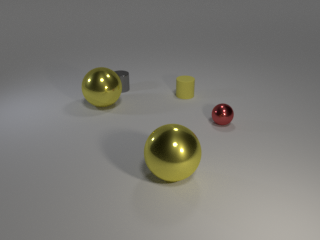}
    }
    \subfigure[$\vec p_\text{1 Cube} =67.3\%$]{
    \includegraphics[width=0.18\columnwidth]{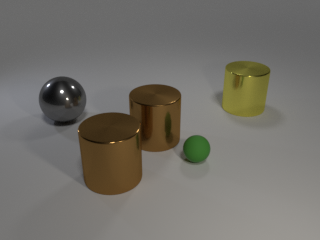}
    }
    \subfigure[$\vec p_\text{1 Cube} =94.5\%$]{
    \includegraphics[width=0.18\columnwidth]{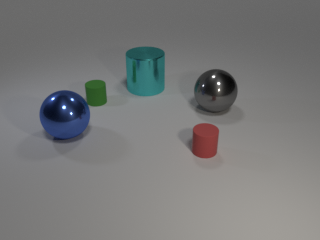}
    }

    \subfigure[$\vec p_\text{1 Sphere} =95.6\%$]{
    \includegraphics[width=0.18\columnwidth]{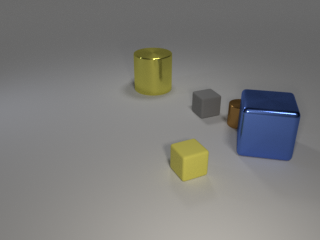}
    }
    \subfigure[$\vec p_\text{1 Sphere} =96.6\%$]{
    \includegraphics[width=0.18\columnwidth]{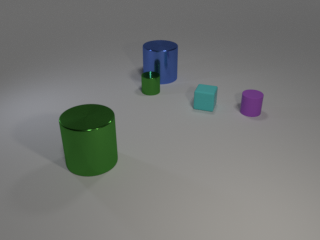}
    }
    \subfigure[$\vec p_\text{1 Sphere} =89.8\%$]{
    \includegraphics[width=0.18\columnwidth]{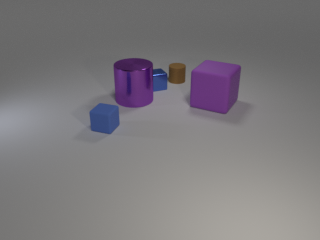}
    }
    \subfigure[$\vec p_\text{1 Sphere} =99.1\%$]{
    \includegraphics[width=0.18\columnwidth]{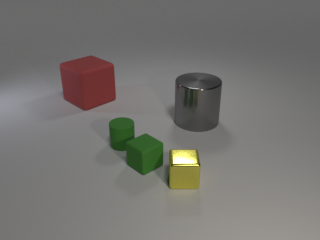}
    }
    \subfigure[$\vec p_\text{1 Sphere} =96.5\%$]{
    \includegraphics[width=0.18\columnwidth]{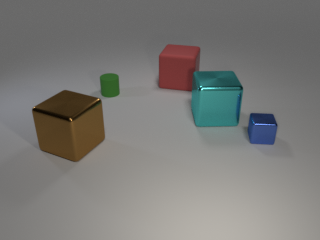}
    }
    
    \subfigure[$\vec p_\text{1 Cyl.} =90.4\%$]{
    \includegraphics[width=0.18\columnwidth]{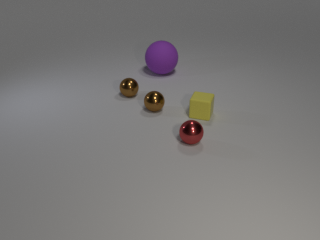}
    }
    \subfigure[$\vec p_\text{1 Cyl.} =98.6\%$]{
    \includegraphics[width=0.18\columnwidth]{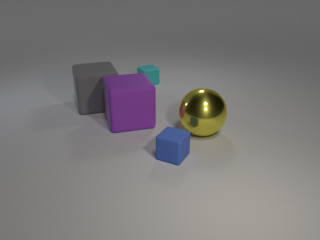}
    }
    \subfigure[$\vec p_\text{1 Cyl.} =94.5\%$]{
    \includegraphics[width=0.18\columnwidth]{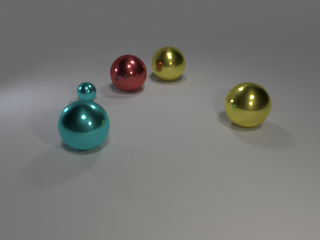}
    }
    \subfigure[$\vec p_\text{1 Cyl.} =96.5\%$]{
    \includegraphics[width=0.18\columnwidth]{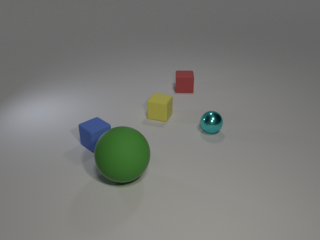}
    }
    \subfigure[$\vec p_\text{1 Cyl.} =98.5\%$]{
    \includegraphics[width=0.18\columnwidth]{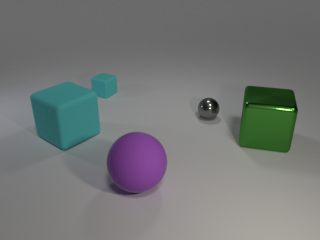}
    }

    \subfigure[$\vec p_\text{2 Cyl.} =85.9\%$]{
    \includegraphics[width=0.18\columnwidth]{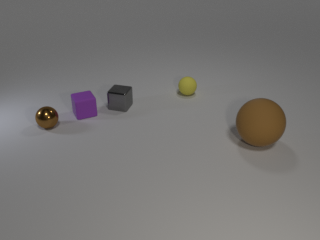}
    }
    \subfigure[$\vec p_\text{2 Cyl.} =60.2\%$]{
    \includegraphics[width=0.18\columnwidth]{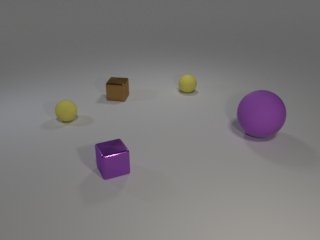}
    }
    \subfigure[$\vec p_\text{2 Cyl.} =79.4\%$]{
    \includegraphics[width=0.18\columnwidth]{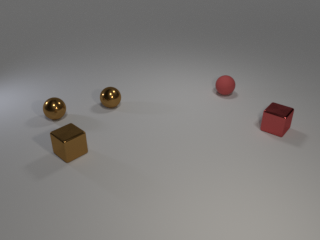}
    }
    \subfigure[$\vec p_\text{2 Cyl.} =48.4\%$]{
    \includegraphics[width=0.18\columnwidth]{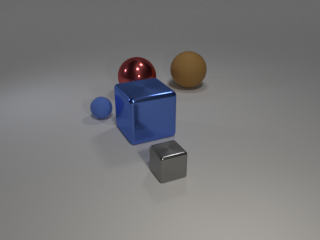}
    }
    \subfigure[$\vec p_\text{2 Cyl.} =60.5\%$]{
    \includegraphics[width=0.18\columnwidth]{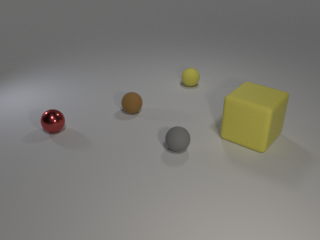}
    }
    \caption{Sampled high confidence misclassified examples and their associated prediction confidences. For each target constraint (e.g., ``1 Cube''), objects from the target class (e.g., cubes) are excluded from the data distribution. The resultant images are composed entirely of non-target-class objects, (e.g., cylinders and spheres).}
    \label{fig:adversarial-clevr-supp}
\end{figure}

Fig.~\ref{fig:adversarial-clevr-supp} shows such examples for CLEVR. For each target inference (e.g. ``1 Cube''), we exclude objects belonging to the target class from the data distribution. 

\begin{figure}[!htb]
    \centering
    \includegraphics[width=0.19\columnwidth]{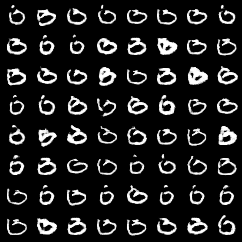}
    \includegraphics[width=0.19\columnwidth]{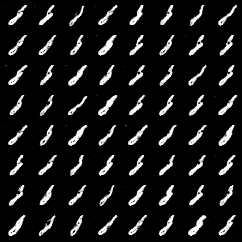}
    \includegraphics[width=0.19\columnwidth]{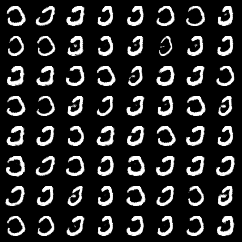}
    \includegraphics[width=0.19\columnwidth]{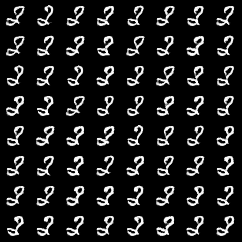}
    \includegraphics[width=0.19\columnwidth]{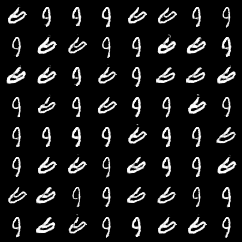}
    
    \includegraphics[width=0.19\columnwidth]{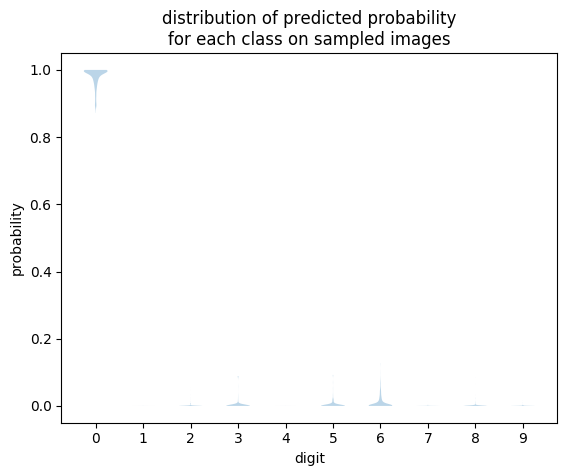}
    \includegraphics[width=0.19\columnwidth]{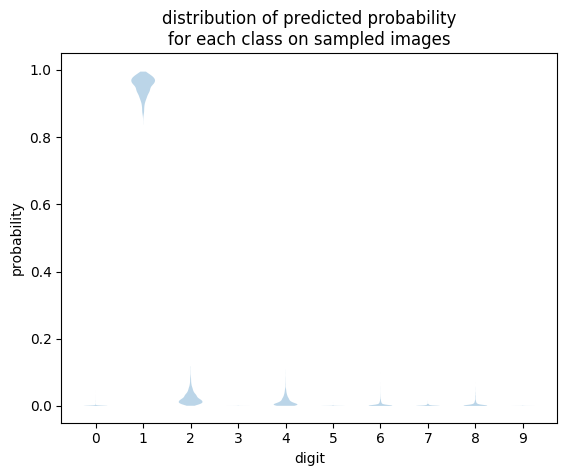}
    \includegraphics[width=0.19\columnwidth]{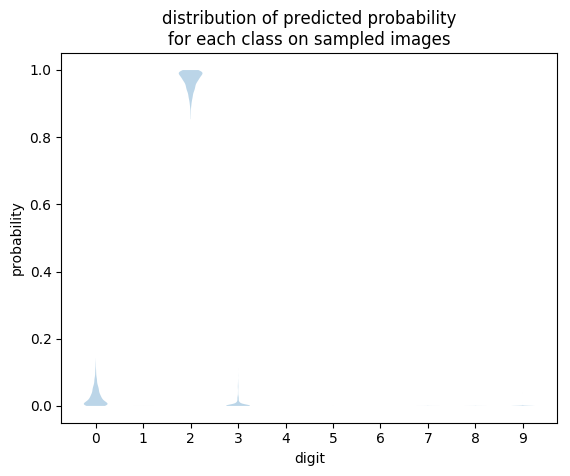}
    \includegraphics[width=0.19\columnwidth]{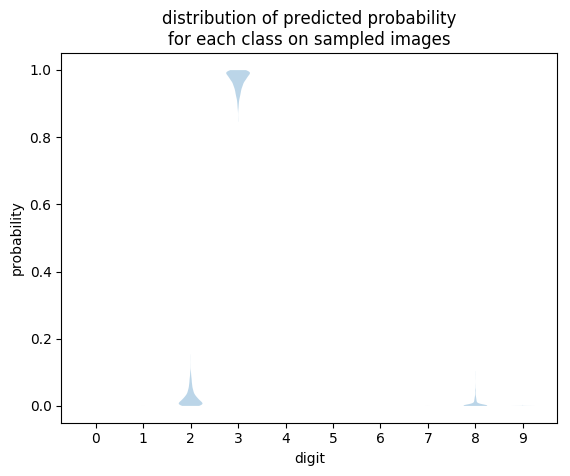}
    \includegraphics[width=0.19\columnwidth]{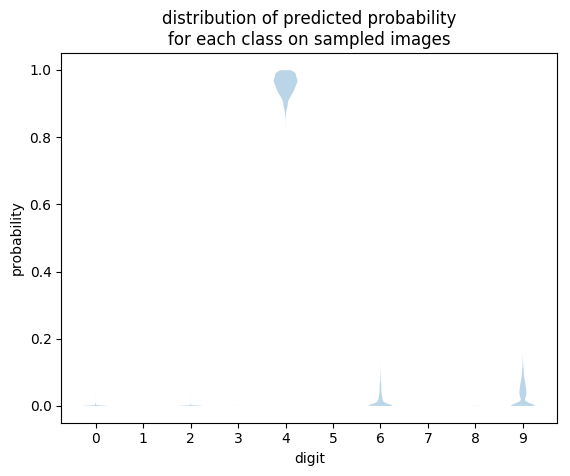}
    
    \includegraphics[width=0.19\columnwidth]{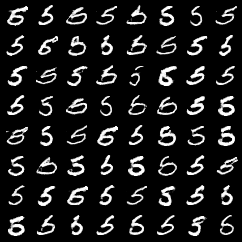}
    \includegraphics[width=0.19\columnwidth]{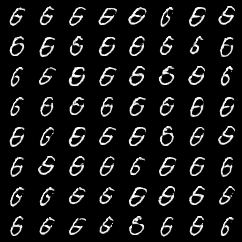}
    \includegraphics[width=0.19\columnwidth]{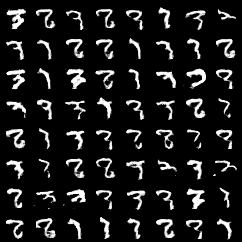}
    \includegraphics[width=0.19\columnwidth]{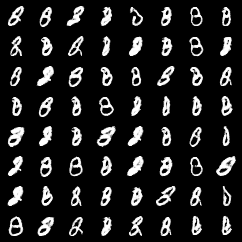}
    \includegraphics[width=0.19\columnwidth]{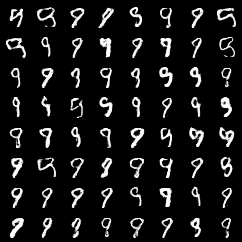}
    
    \includegraphics[width=0.19\columnwidth]{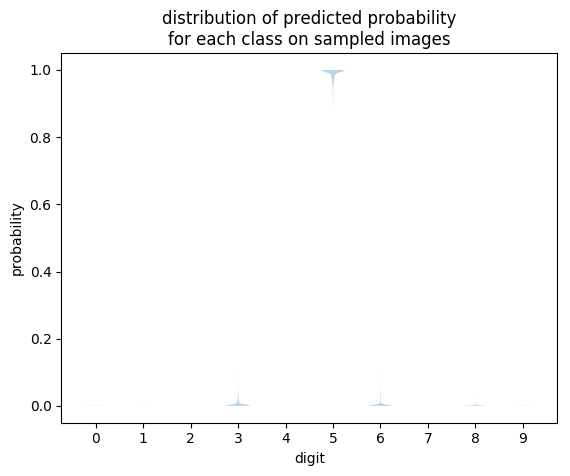}
    \includegraphics[width=0.19\columnwidth]{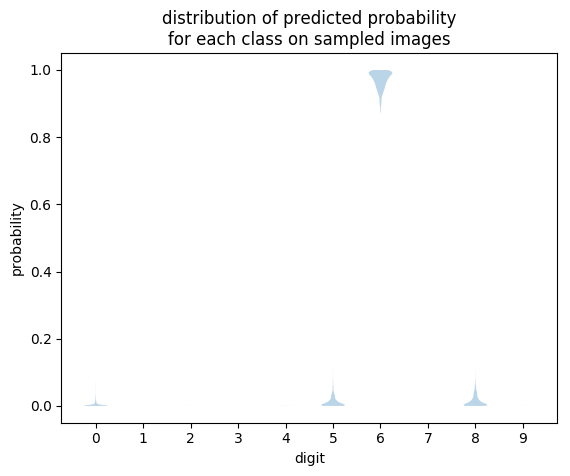}
    \includegraphics[width=0.19\columnwidth]{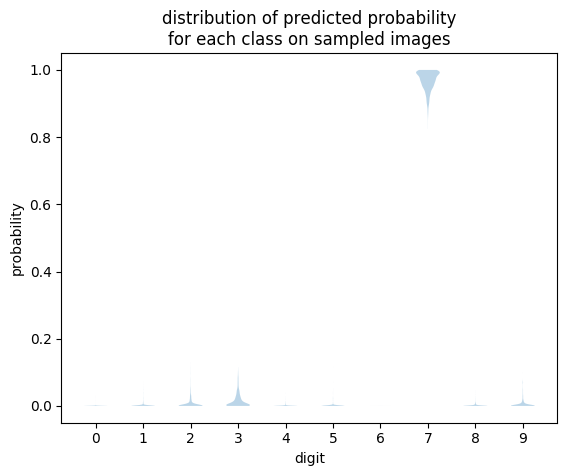}
    \includegraphics[width=0.19\columnwidth]{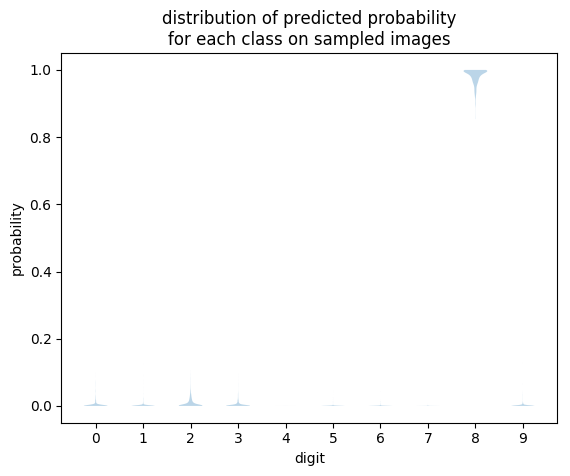}
    \includegraphics[width=0.19\columnwidth]{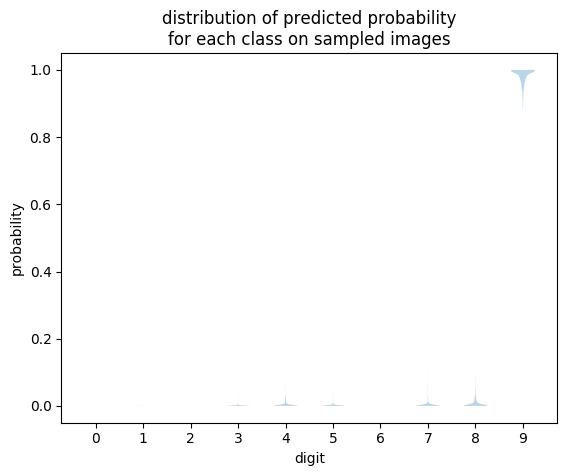}
    \caption{Examples and violin plots for high confidence misclassified examples. Top two rows: 0-4; bottom two rows: 5-9. }
    \label{fig:adv-mnist}
\end{figure}

\clearpage
Fig.~\ref{fig:adv-mnist} presents high-confidence misclassifications for each classes of MNIST, with digit 0-4 on the top two rows and digit 5-9 on the bottom two rows. 

\begin{figure}[!htb]
    \centering
    \includegraphics[width=0.19\columnwidth]{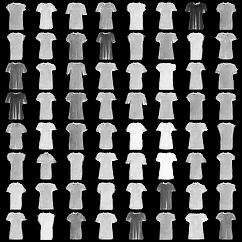}
    \includegraphics[width=0.19\columnwidth]{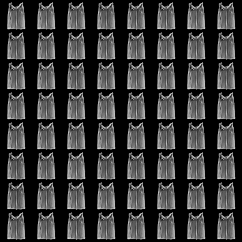}
    \includegraphics[width=0.19\columnwidth]{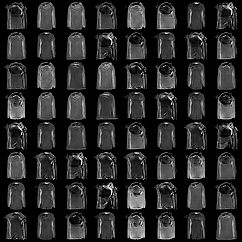}
    \includegraphics[width=0.19\columnwidth]{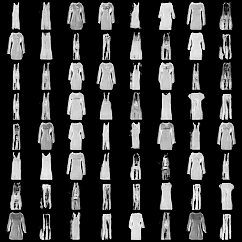}
    \includegraphics[width=0.19\columnwidth]{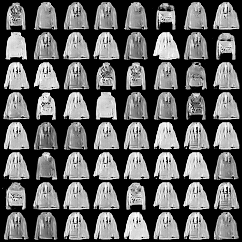}
    
    \includegraphics[width=0.19\columnwidth]{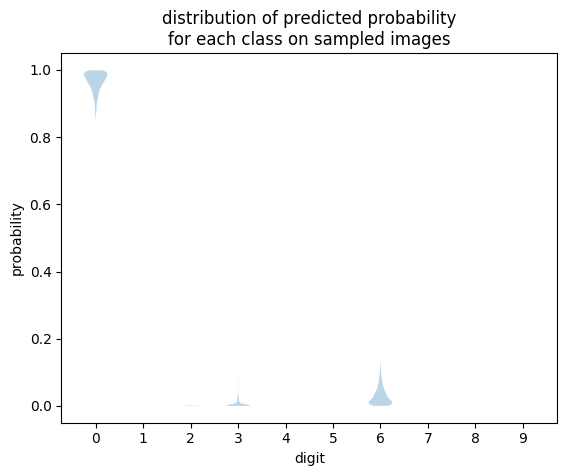}
    \includegraphics[width=0.19\columnwidth]{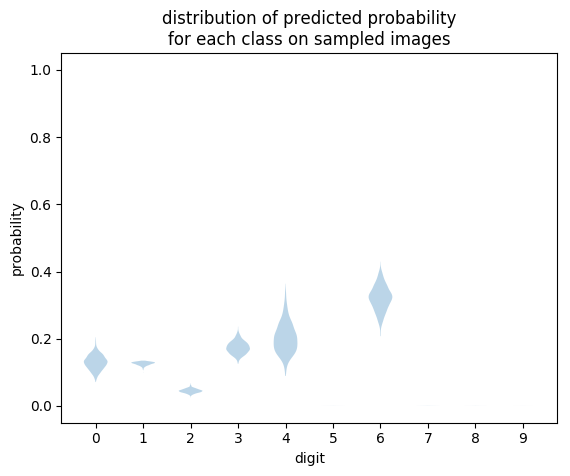}
    \includegraphics[width=0.19\columnwidth]{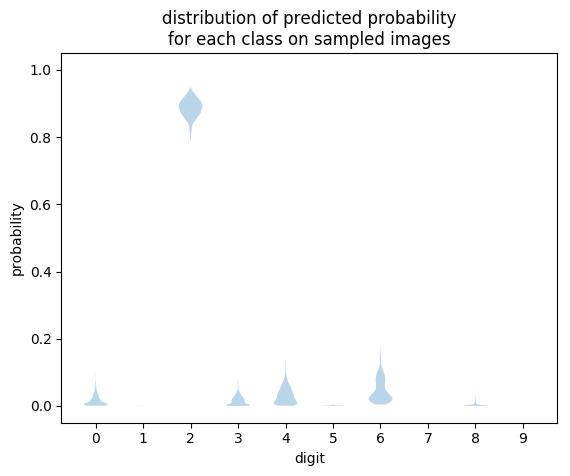}
    \includegraphics[width=0.19\columnwidth]{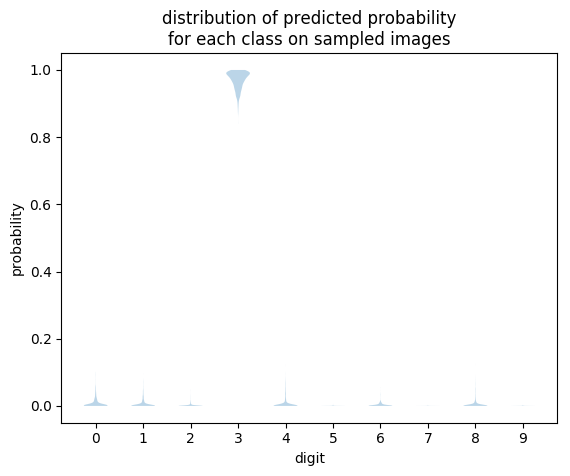}
    \includegraphics[width=0.19\columnwidth]{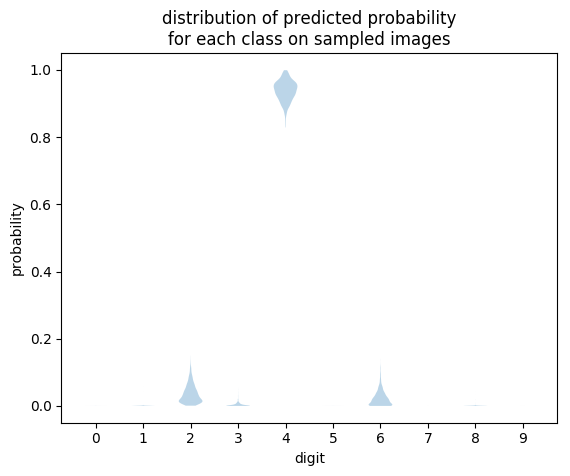}
    
    \includegraphics[width=0.19\columnwidth]{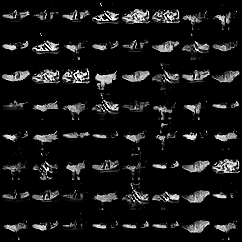}
    \includegraphics[width=0.19\columnwidth]{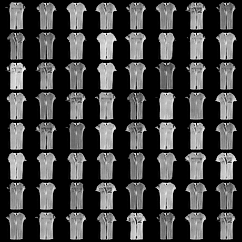}
    \includegraphics[width=0.19\columnwidth]{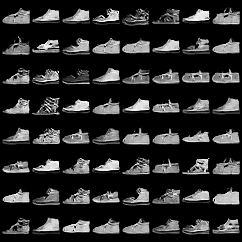}
    \includegraphics[width=0.19\columnwidth]{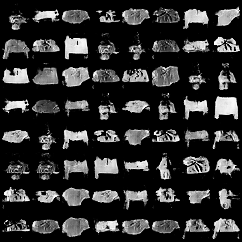}
    \includegraphics[width=0.19\columnwidth]{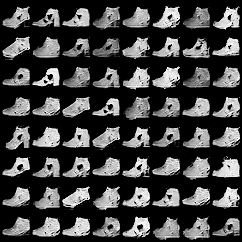}
    
    \includegraphics[width=0.19\columnwidth]{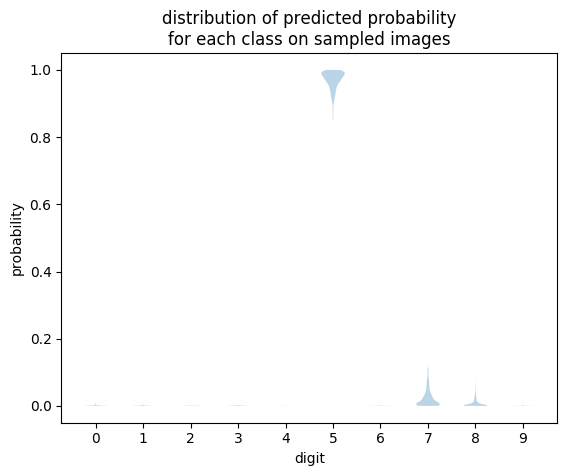}
    \includegraphics[width=0.19\columnwidth]{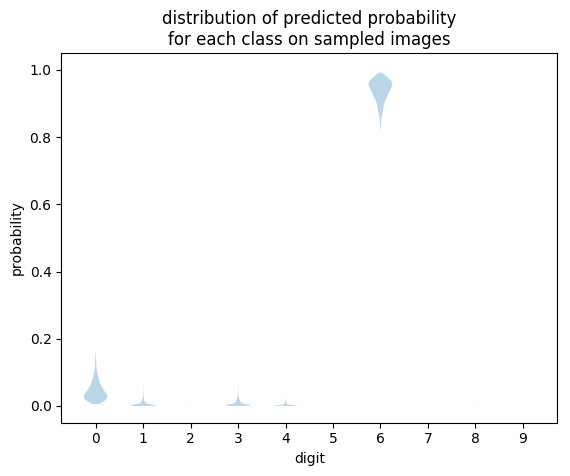}
    \includegraphics[width=0.19\columnwidth]{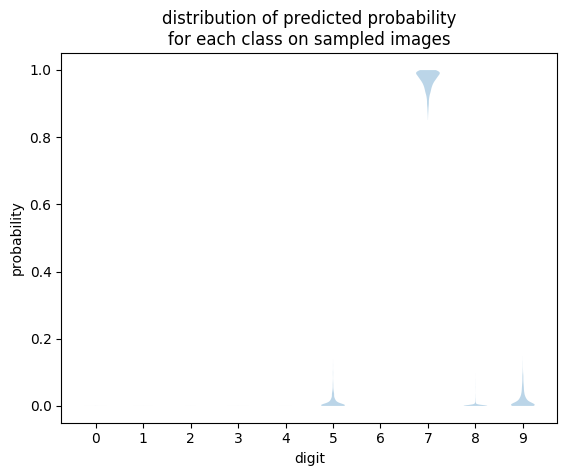}
    \includegraphics[width=0.19\columnwidth]{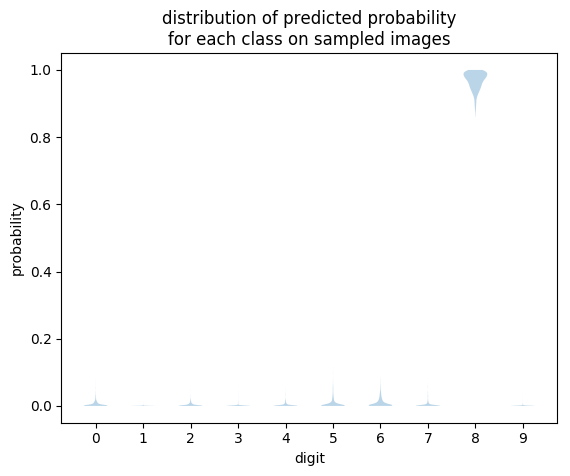}
    \includegraphics[width=0.19\columnwidth]{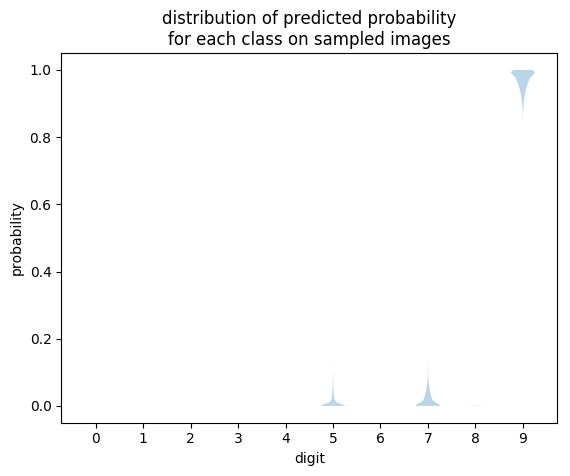}
    \caption{Samples and violin plots for  high confidence misclassified examples. Top row: T-shirt, trousers (sample failure), pullover, dress, coat. Bottom row: sandal, shirt, sneaker, bag, ankle boot. }
    \label{fig:adv-fashion-mnist}
\end{figure}

\clearpage
Fig.~\ref{fig:adv-fashion-mnist} presents high-confidence misclassifications for each classes of Fashion-MNIST, with T-shirt, trousers pullover, dress and coat on the top two rows and sandal, shirt, sneaker, bag and ankle boot on the bottom two rows. The confidence plot for the trousers samples  indicates that the sampling is not successful. 

\clearpage
\section{Novel Class Extrapolation Analysis}

\label{ood-supp}
\begin{figure}[h]
    \centering
    
    \subfigure[$\vec p_\text{1 Sph.} =99.3\%$]{
    \includegraphics[width=0.18\columnwidth]{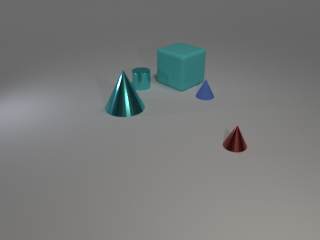}
    }
    \subfigure[$\vec p_\text{1 Sph.} =95.9\%$]{
    \includegraphics[width=0.18\columnwidth]{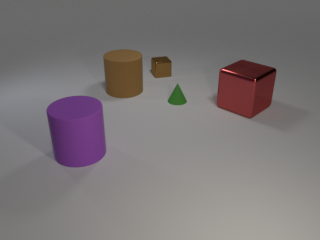}
    }
    \subfigure[$\vec p_\text{1 Sph.} =99.3\%$]{
    \includegraphics[width=0.18\columnwidth]{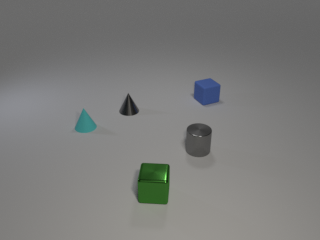}
    }
    \subfigure[$\vec p_\text{1 Sph.} =97.7\%$]{
    \includegraphics[width=0.18\columnwidth]{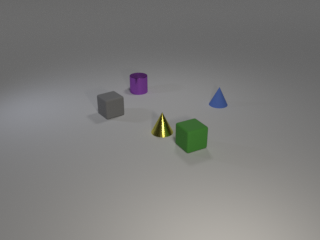}
    }
    \subfigure[$\vec p_\text{1 Sph.} =97.3\%$]{
    \includegraphics[width=0.18\columnwidth]{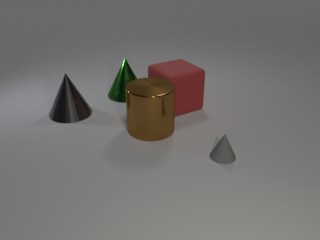}
    }

    \subfigure[$\vec p_\text{1 Cube} =99.2\%$]{
    \includegraphics[width=0.18\columnwidth]{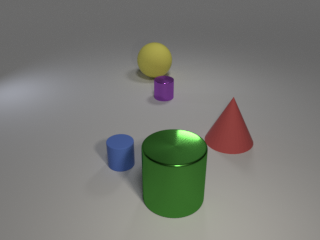}
    }
    \subfigure[$\vec p_\text{1 Cube} =97.5\%$]{
    \includegraphics[width=0.18\columnwidth]{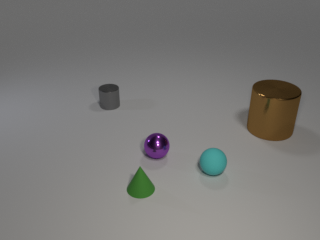}
    }
    \subfigure[$\vec p_\text{1 Cube} =98.7\%$]{
    \includegraphics[width=0.18\columnwidth]{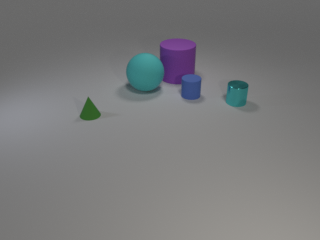}
    }
    \subfigure[$\vec p_\text{1 Cube} =99.0\%$]{
    \includegraphics[width=0.18\columnwidth]{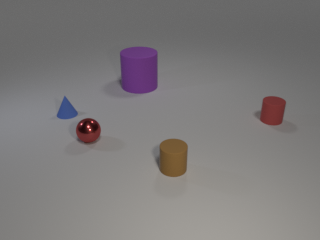}
    }    
    \subfigure[$\vec p_\text{1 Cube} =98.7\%$]{
    \includegraphics[width=0.18\columnwidth]{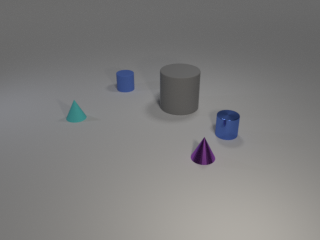}
    }

    \subfigure[$\vec p_\text{1 Cyl.} =96.9\%$]{
    \includegraphics[width=0.18\columnwidth]{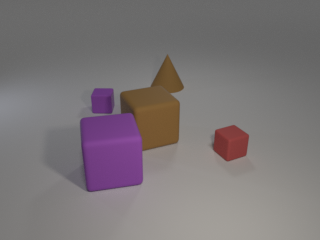}
    }
    \subfigure[$\vec p_\text{1 Cyl.} =99.1\%$]{
    \includegraphics[width=0.18\columnwidth]{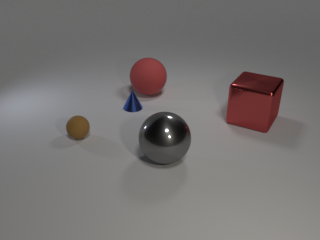}
    }
    \subfigure[$\vec p_\text{1 Cyl.} =96.5\%$]{
    \includegraphics[width=0.18\columnwidth]{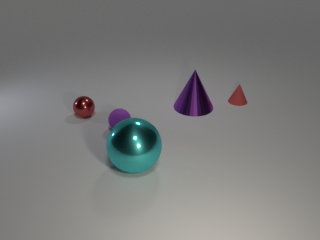}
    }
    \subfigure[$\vec p_\text{1 Cyl.} =97.2\%$]{
    \label{fig:no-extrapolation}
    \includegraphics[width=0.18\columnwidth]{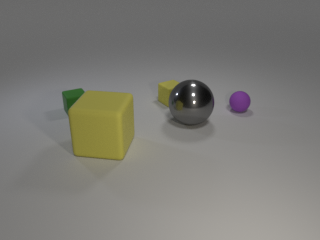}
    }    
    \subfigure[$\vec p_\text{1 Cyl.} =99.0\%$]{
    \includegraphics[width=0.18\columnwidth]{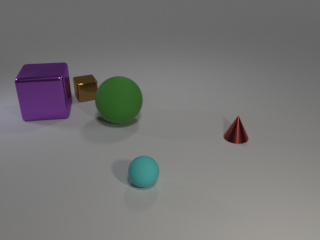}
    }

    \subfigure[$\vec p_\text{5 Cubes} =74.6\%$]{
    \label{fig:extrapolation-low-performance}
    \includegraphics[width=0.18\columnwidth]{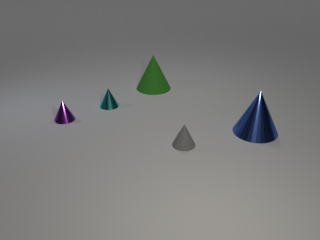}
    }
    \subfigure[$\vec p_\text{5 Cubes} =89.5\%$]{
    \includegraphics[width=0.18\columnwidth]{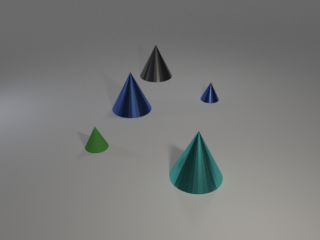}
    }
    \subfigure[$\vec p_\text{5 Cubes} =93.3\%$]{
    \includegraphics[width=0.18\columnwidth]{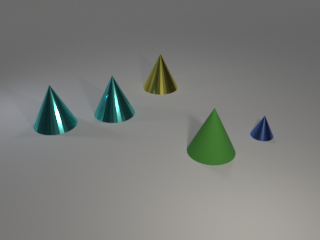}
    }
    \subfigure[$\vec p_\text{5 Cubes} =91.6\%$]{
    \includegraphics[width=0.18\columnwidth]{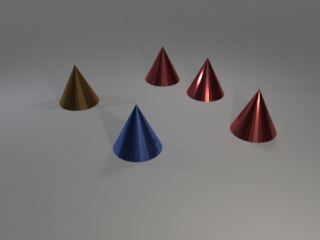}
    }
    \subfigure[$\vec p_\text{5 Cubes} =89.9\%$]{
    \includegraphics[width=0.18\columnwidth]{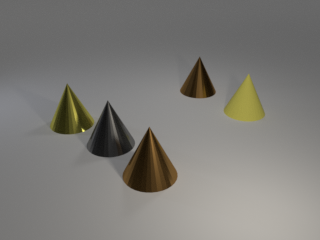}
    }
    
    \caption{Sampled novel class extrapolation examples and their associated prediction confidences. Similar to high confidence misclassified examples, for each target constraint (e.g., ``1 Cube''), we remove examples of the target class (e.g., cubes) from the data distribution, but add to the cone object to it, a novel class not present in the training distribution. \ref{fig:no-extrapolation} is the only example which by chance does not include a novel class object.}
    \label{fig:extrapolation-clevr-supp}
\end{figure}

Figure \ref{fig:extrapolation-clevr-supp} shows novel class extrapolation examples for CLEVR. 

\clearpage

\begin{figure}[!htb]
    \centering
    \includegraphics[width=0.19\columnwidth]{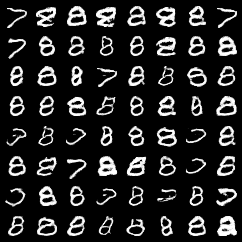}
    \includegraphics[width=0.19\columnwidth]{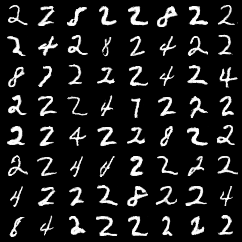}
    \includegraphics[width=0.19\columnwidth]{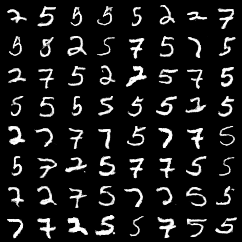}
    \includegraphics[width=0.19\columnwidth]{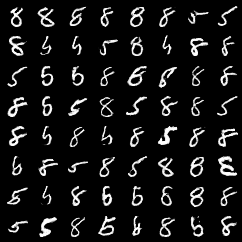}
    \includegraphics[width=0.19\columnwidth]{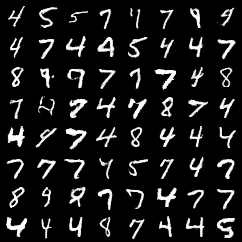}
    
    \includegraphics[width=0.19\columnwidth]{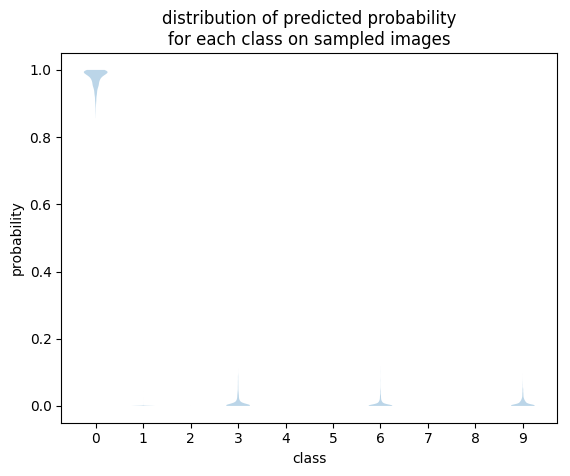}
    \includegraphics[width=0.19\columnwidth]{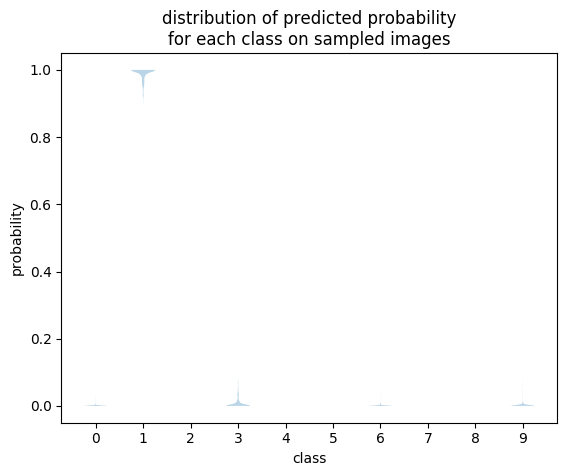}
    \includegraphics[width=0.19\columnwidth]{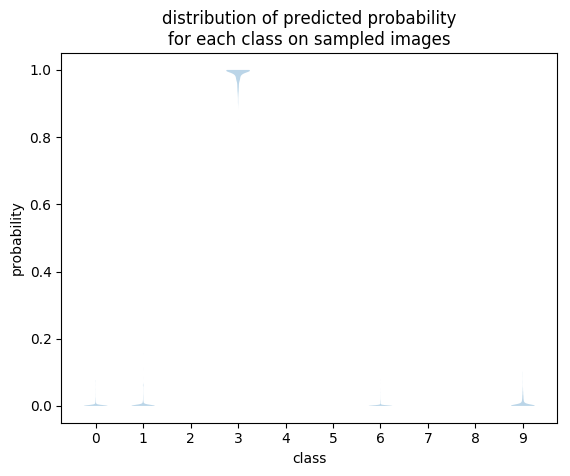}
    \includegraphics[width=0.19\columnwidth]{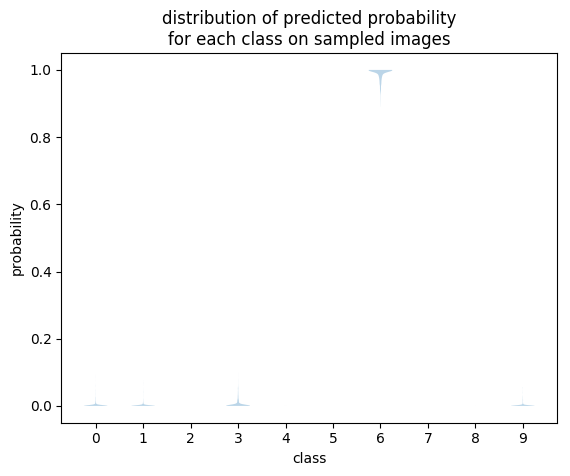}
    \includegraphics[width=0.19\columnwidth]{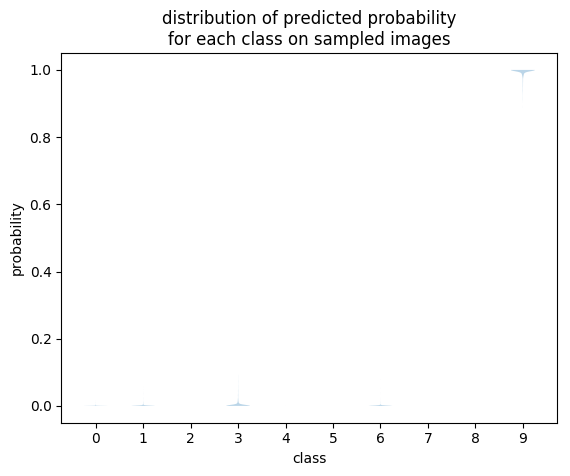}
    
    \caption{Samples and confidence plots for MNIST novel class extrapolation for digits 0, 1, 3, 6 and 9, in that order. }
    \label{fig:ood-mnist-supp}
\end{figure}

Fig.~\ref{fig:ood-mnist-supp} shows examples for novel-class extrapolation on MNIST. The classifier is trained on digit 0, 1, 3, 6 and 9, and tested on images generated by a GAN trained on digit 2, 4, 5, 7 and 8. 

\begin{figure}[!htb]
    \centering
    \includegraphics[width=0.19\columnwidth]{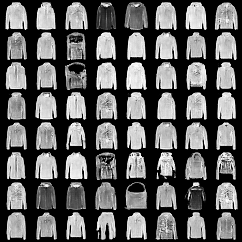}
    \includegraphics[width=0.19\columnwidth]{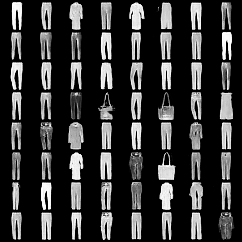}
    \includegraphics[width=0.19\columnwidth]{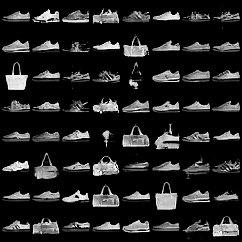}
    \includegraphics[width=0.19\columnwidth]{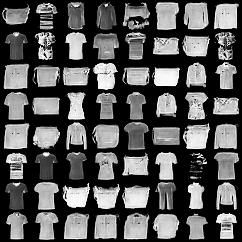}
    \includegraphics[width=0.19\columnwidth]{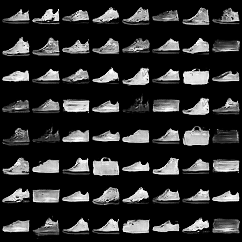}
    
    \includegraphics[width=0.19\columnwidth]{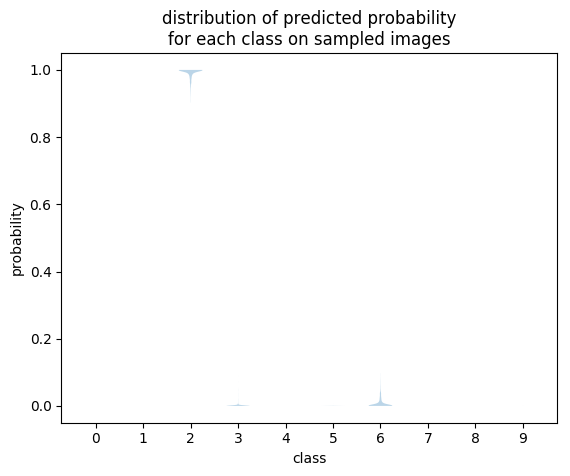}
    \includegraphics[width=0.19\columnwidth]{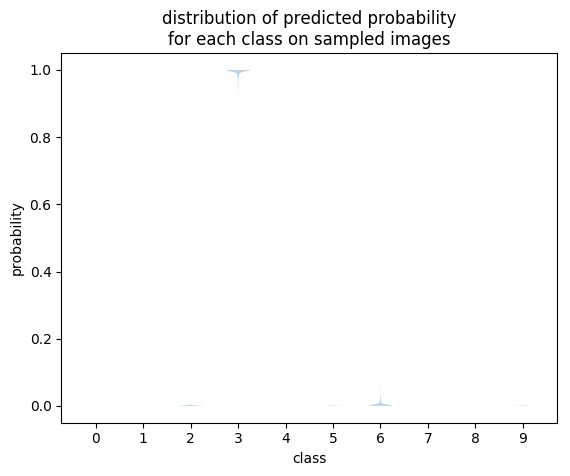}
    \includegraphics[width=0.19\columnwidth]{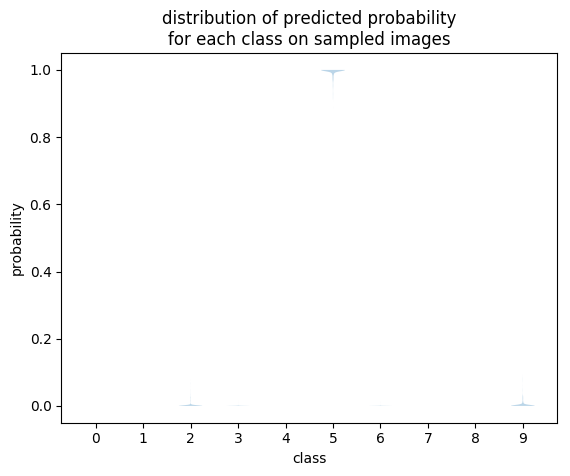}
    \includegraphics[width=0.19\columnwidth]{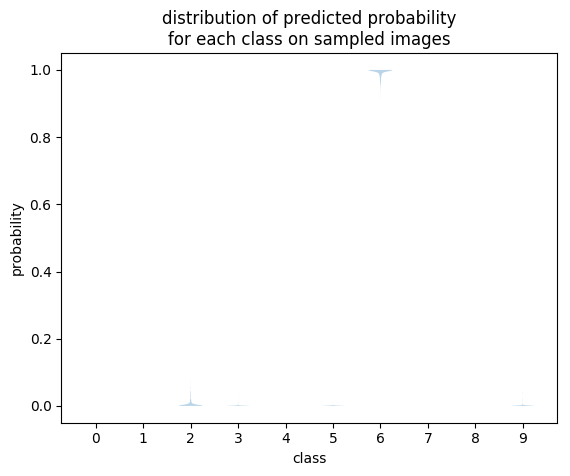}
    \includegraphics[width=0.19\columnwidth]{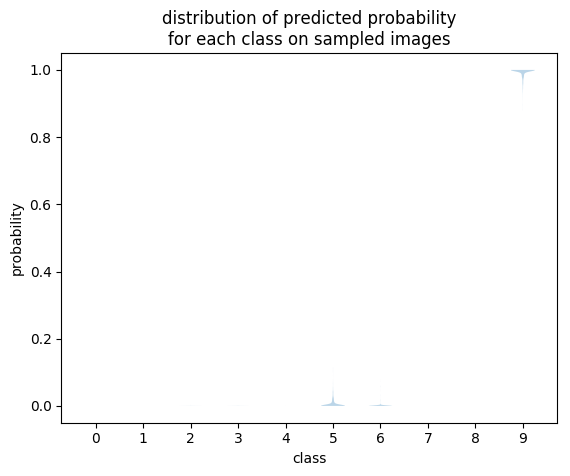}
    
    \caption{Samples and confidence plots for Fashion-MNIST novel class extrapolation for pullover, dress, sandal, shirt and ankle boot, in that order. }
    \label{fig:ood-fashion-mnist-supp}
\end{figure}

Fig.~\ref{fig:ood-fashion-mnist-supp} shows examples for novel-class extrapolation on Fashion-MNIST. The classifier is trained on pullover, dress, sandal, shirt and ankle boot, and tested on images generated by a GAN trained on T-shirt, trousers, coat, sneaker and bag. 

\clearpage
\section{Domain Adaptation Analysis}
\label{da-supp}
\begin{figure}[!htb]
    \centering
    \includegraphics[width=0.14\columnwidth]{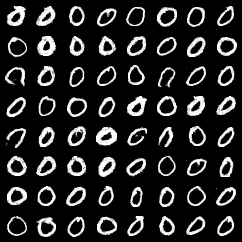}
    \includegraphics[width=0.14\columnwidth]{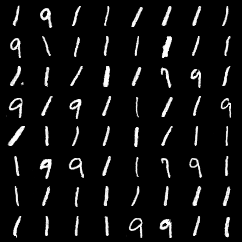}
    \includegraphics[width=0.14\columnwidth]{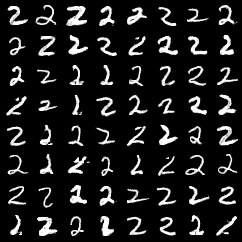}
    \includegraphics[width=0.14\columnwidth]{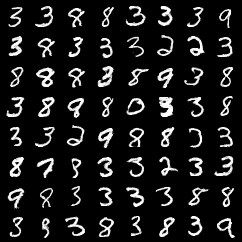}
    \includegraphics[width=0.14\columnwidth]{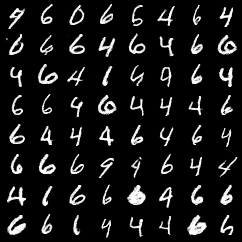}

    \includegraphics[width=0.14\columnwidth]{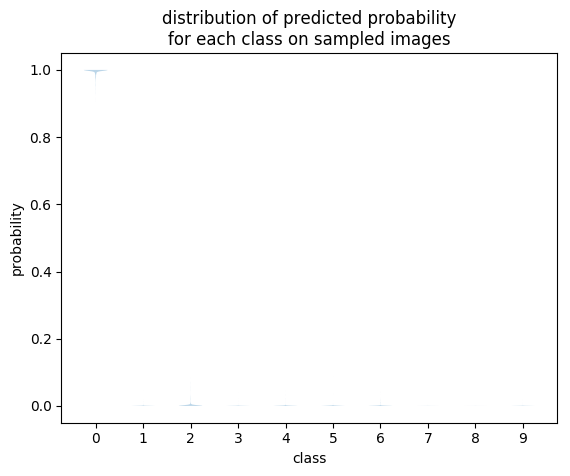}
    \includegraphics[width=0.14\columnwidth]{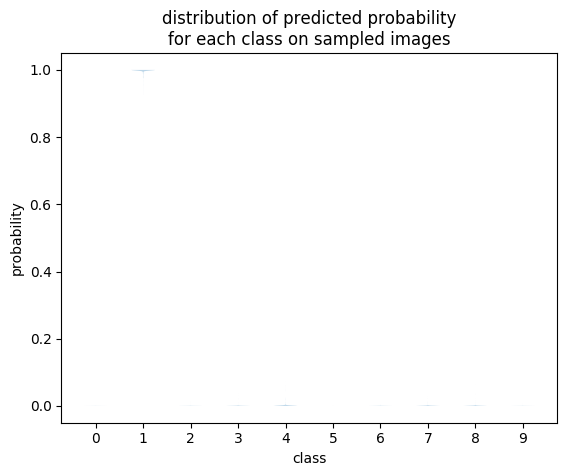}
    \includegraphics[width=0.14\columnwidth]{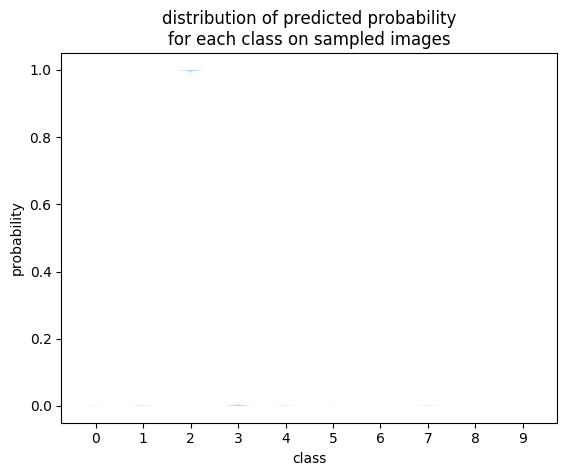}
    \includegraphics[width=0.14\columnwidth]{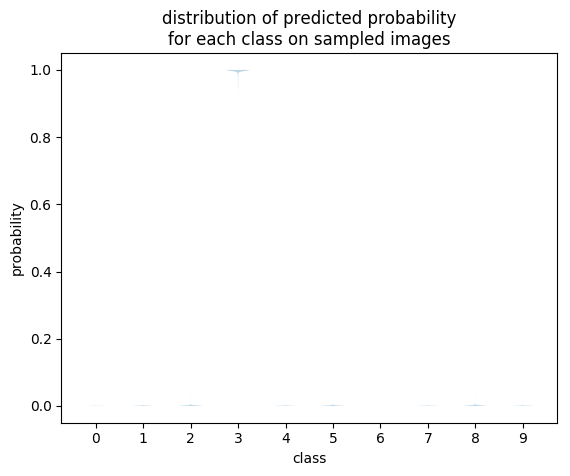}
    \includegraphics[width=0.14\columnwidth]{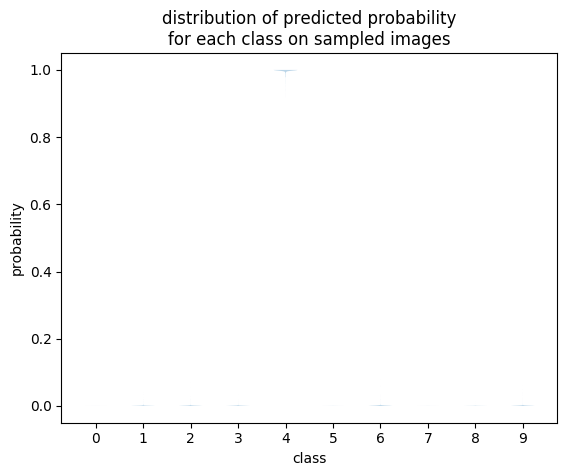}

    \includegraphics[width=0.14\columnwidth]{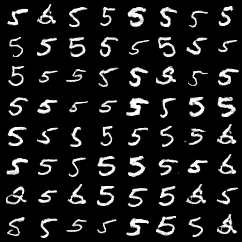}
    \includegraphics[width=0.14\columnwidth]{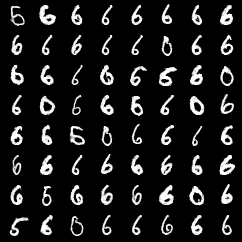}
    \includegraphics[width=0.14\columnwidth]{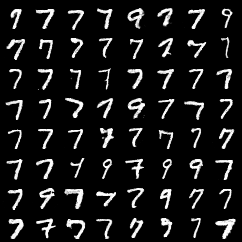}
    \includegraphics[width=0.14\columnwidth]{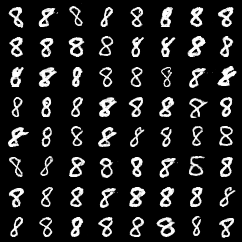}
    \includegraphics[width=0.14\columnwidth]{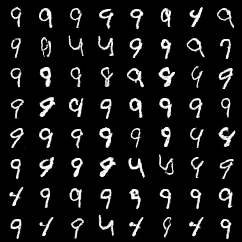}

    \includegraphics[width=0.14\columnwidth]{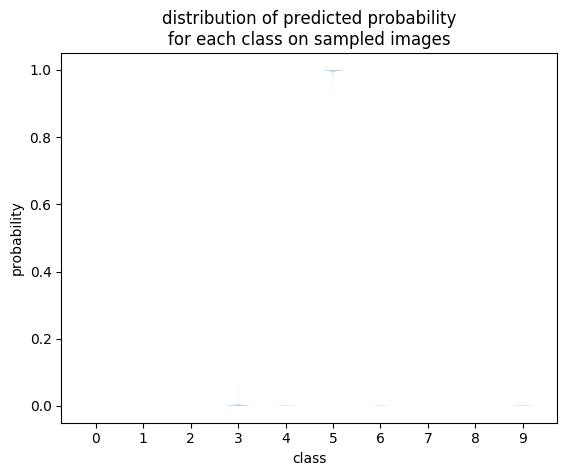}
    \includegraphics[width=0.14\columnwidth]{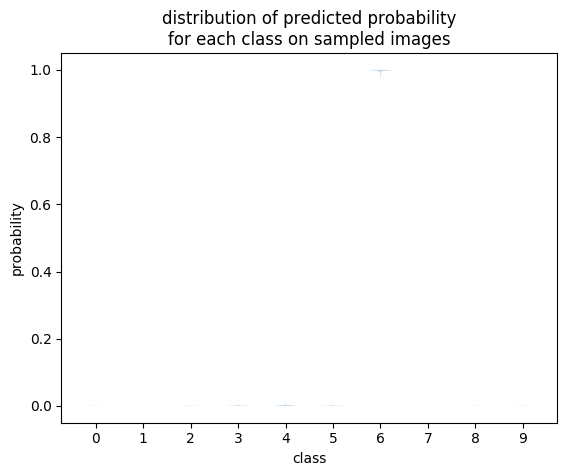}
    \includegraphics[width=0.14\columnwidth]{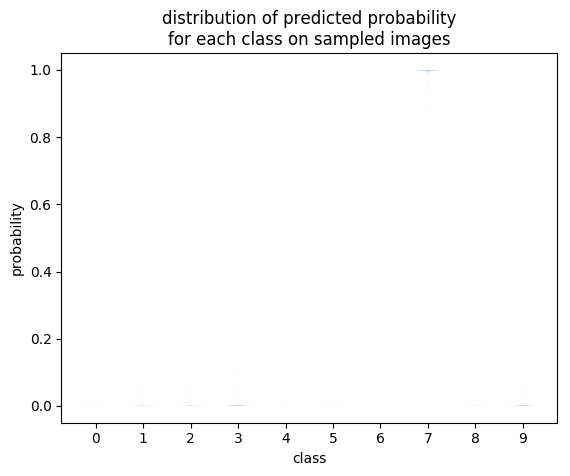}
    \includegraphics[width=0.14\columnwidth]{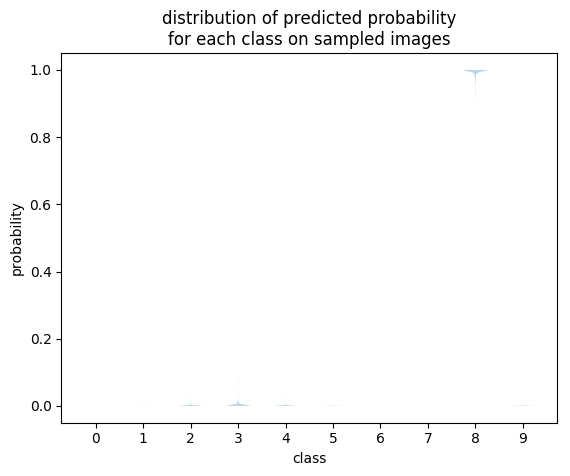}
    \includegraphics[width=0.14\columnwidth]{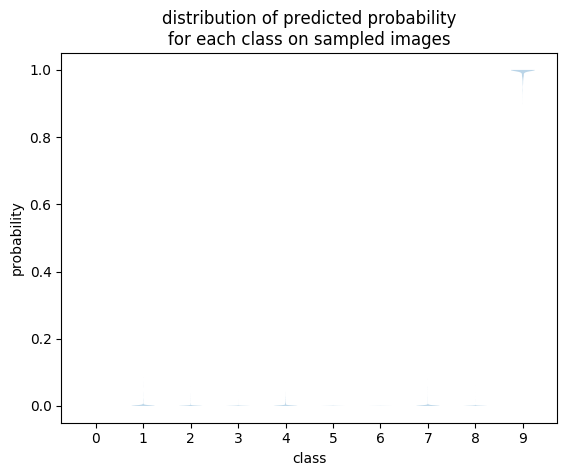}
    \caption{High confident MNIST samples generated for each class as predicted by the baseline model. }
    \label{fig:da-baseline}
\end{figure}

\begin{figure}[!htb]
    \centering
    \includegraphics[width=0.14\columnwidth]{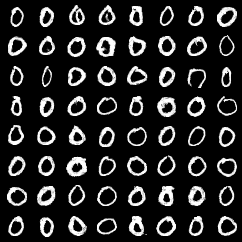}
    \includegraphics[width=0.14\columnwidth]{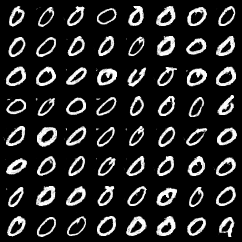}
    \includegraphics[width=0.14\columnwidth]{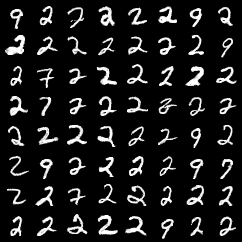}
    \includegraphics[width=0.14\columnwidth]{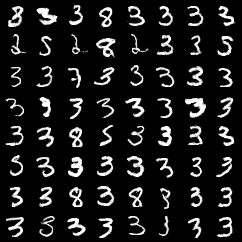}
    \includegraphics[width=0.14\columnwidth]{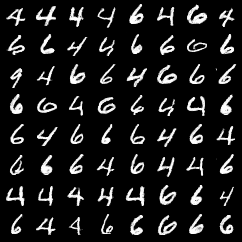}

    \includegraphics[width=0.14\columnwidth]{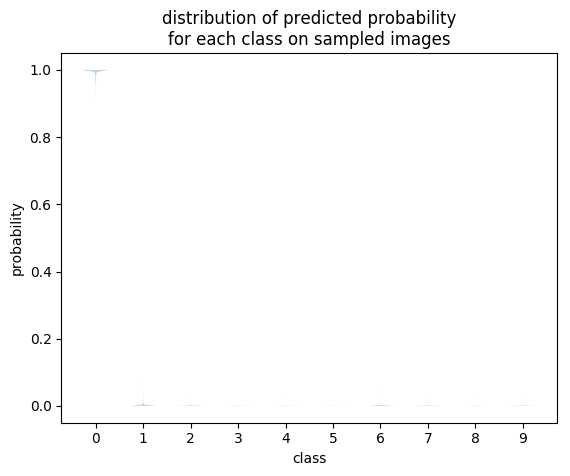}
    \includegraphics[width=0.14\columnwidth]{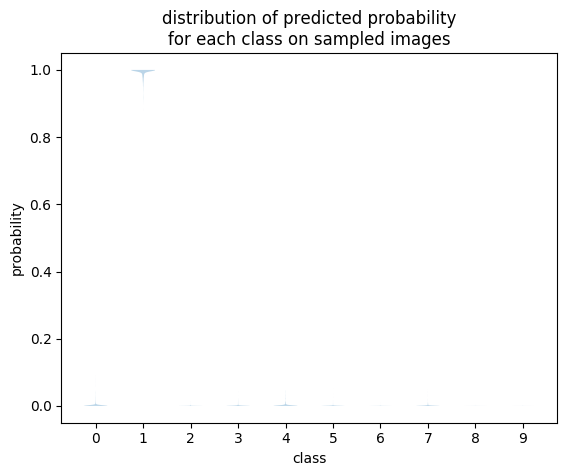}
    \includegraphics[width=0.14\columnwidth]{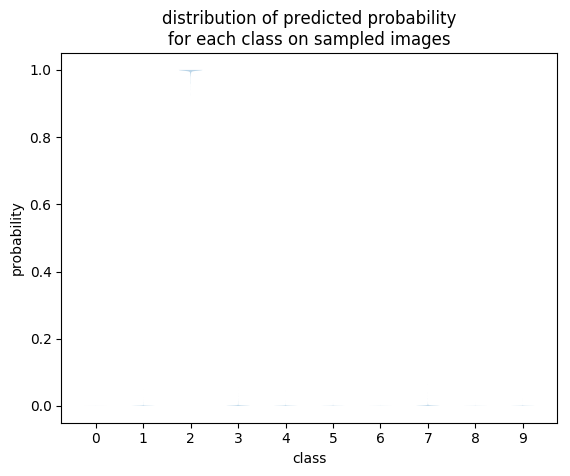}
    \includegraphics[width=0.14\columnwidth]{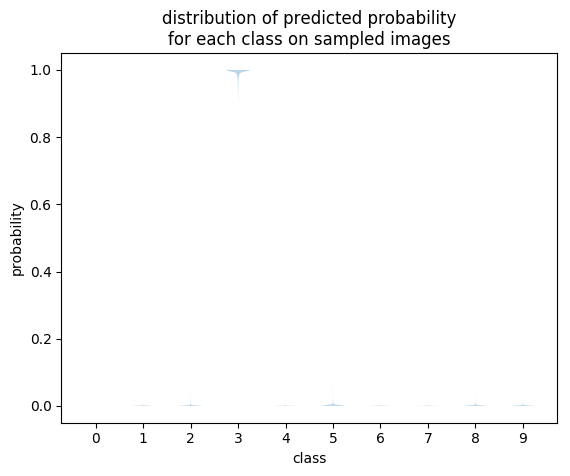}
    \includegraphics[width=0.14\columnwidth]{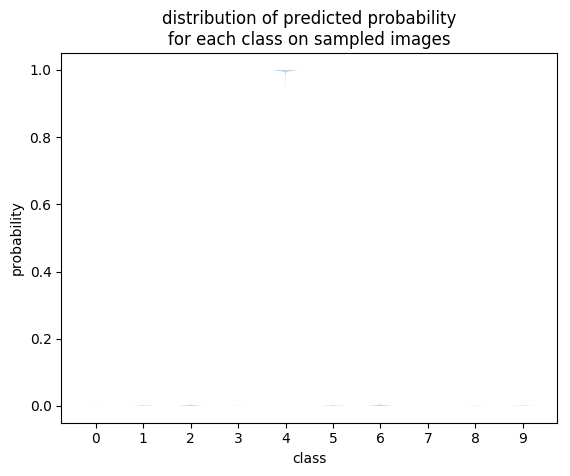}

    \includegraphics[width=0.14\columnwidth]{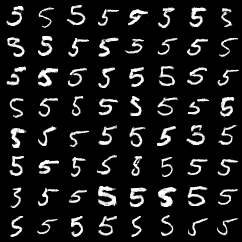}
    \includegraphics[width=0.14\columnwidth]{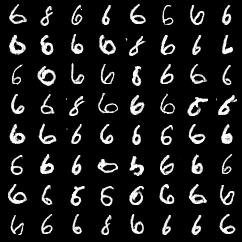}
    \includegraphics[width=0.14\columnwidth]{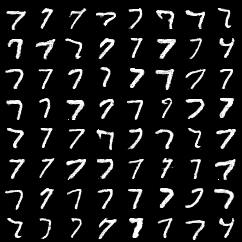}
    \includegraphics[width=0.14\columnwidth]{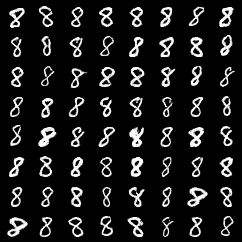}
    \includegraphics[width=0.14\columnwidth]{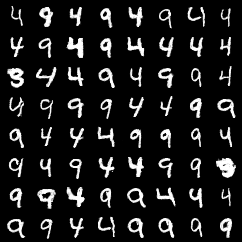}

    \includegraphics[width=0.14\columnwidth]{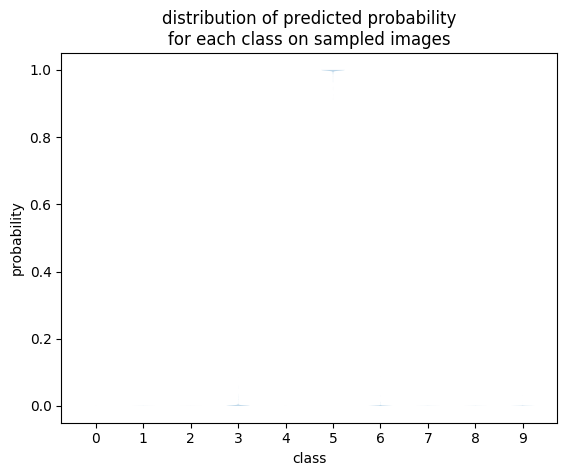}
    \includegraphics[width=0.14\columnwidth]{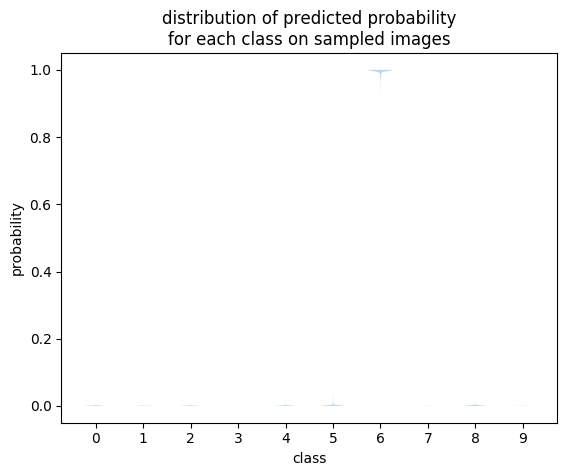}
    \includegraphics[width=0.14\columnwidth]{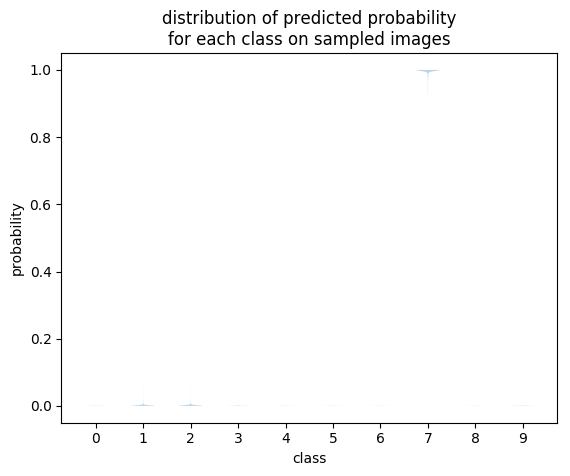}
    \includegraphics[width=0.14\columnwidth]{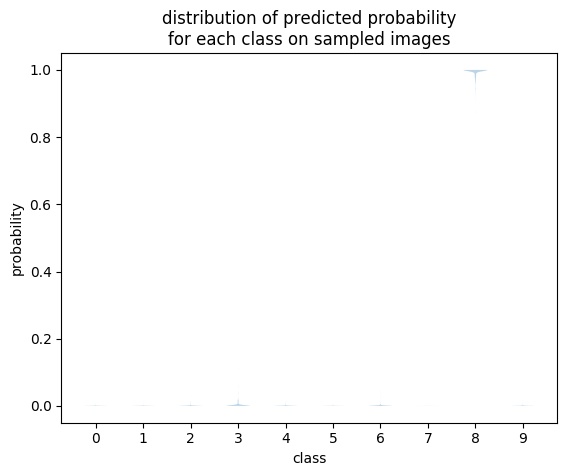}
    \includegraphics[width=0.14\columnwidth]{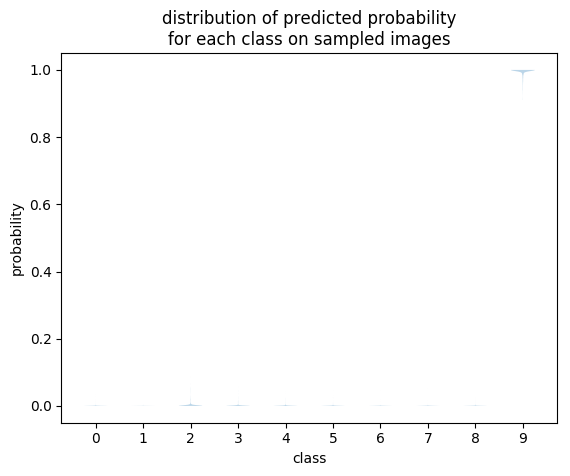}
    \caption{High confident MNIST samples generated for each class as predicted by the ADDA model. }
    \label{fig:da-adda}
\end{figure}

Fig.~\ref{fig:da-baseline} and \ref{fig:da-adda} show additional samples and confidence plots for the baseline and ADDA model, respectively. Top two rows are for digit 0-4, and bottom two rows are for digit 5-9. 

\clearpage
\section{Test Set Evaluation}
\label{test-set-supp}


Tab.~\ref{fig:mnist-test-set-supp} extends Tab.~\ref{tab:baseline-summary} in Sec.~\ref{sec:test-set} and includes misclassified vs.~mislabeled images of all (Fashion-)MNIST classes. 

\begin{table}[!htb]
    \centering
    \caption{An alternative to using \textsc{Bayes-TrEx} for finding highly confident classification failures is to evaluate the high confidence example confusion matrix and associated images from the test set. Here, we show all `misclassified' examples where the classifier failed to predict the given label for the MNIST and Fashion-MNIST datasets. For MNIST, we observe that the majority (60/84) of these images are mislabeled: for example, all of the labeled 2s clearly belong to other classes (8, 7, 7, 3, 1, 7, 7, 7, respectively). While MNIST had 84 total misclassifications, Fashion-MNIST had 802 total misclassifications. We randomly select 10 misclassifications from each class for analysis (with the exception of the ``trousers'' class, as there were 3 total misclassifications for this label). While Fashion-MNIST is more balanced, we again observe a majority of examples to be mislabeled ground truth (52/93) instead of misclassifications.}
    \label{fig:mnist-test-set-supp}
    \newcommand{\imgcharwidth}{0.028\columnwidth}
    
    \begin{tabular}{l m{6cm} m{6.5cm}}
        \toprule
        Class & Misclassified & Mislabeled \\
        \midrule
        0
        &     
        \includegraphics[width=\imgcharwidth]{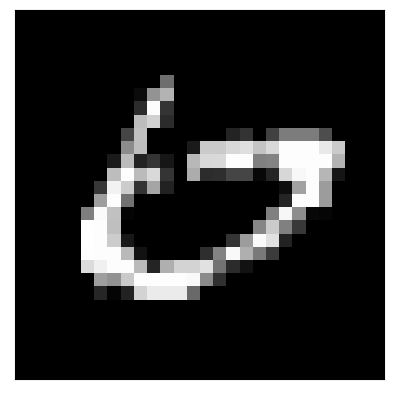}
        \includegraphics[width=\imgcharwidth]{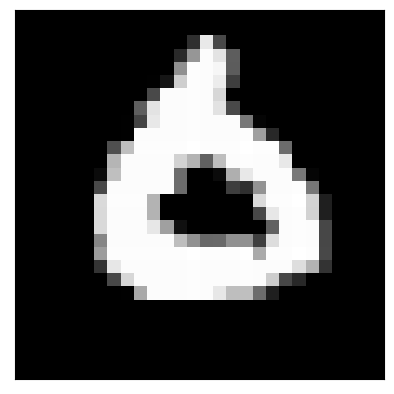}
        \includegraphics[width=\imgcharwidth]{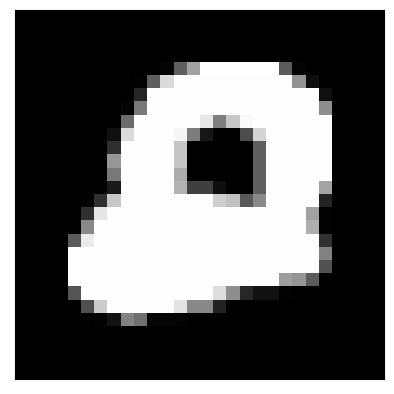} 
        &
        \includegraphics[width=\imgcharwidth]{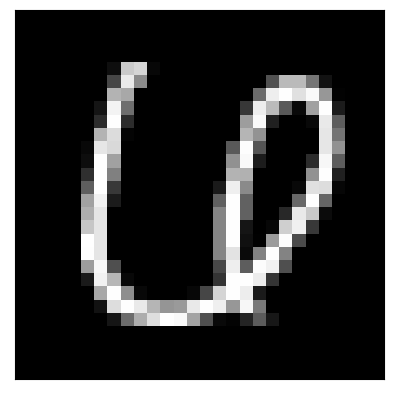}
        \includegraphics[width=\imgcharwidth]{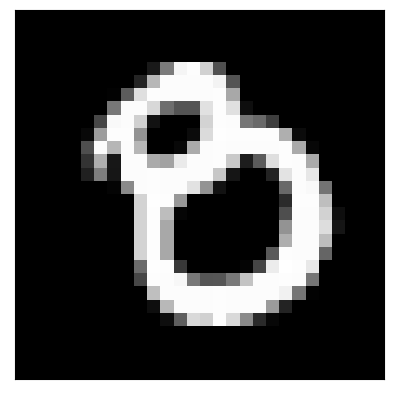}
        \includegraphics[width=\imgcharwidth]{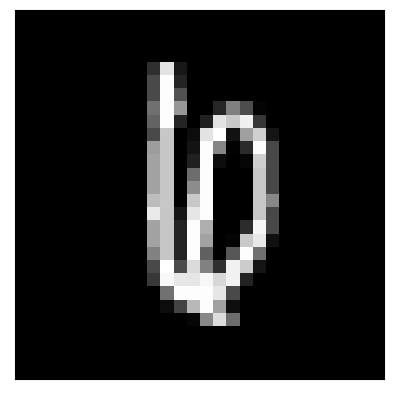}
        \includegraphics[width=\imgcharwidth]{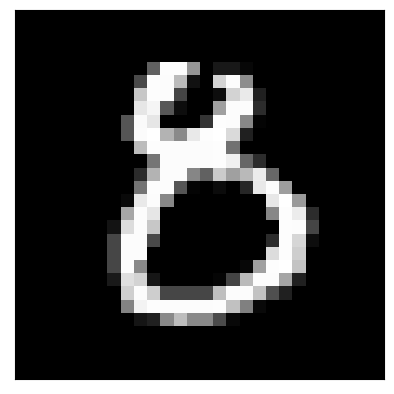}
        \\
        1
        &
        \includegraphics[width=\imgcharwidth]{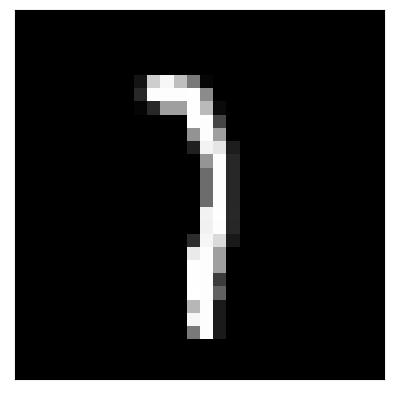}
        \includegraphics[width=\imgcharwidth]{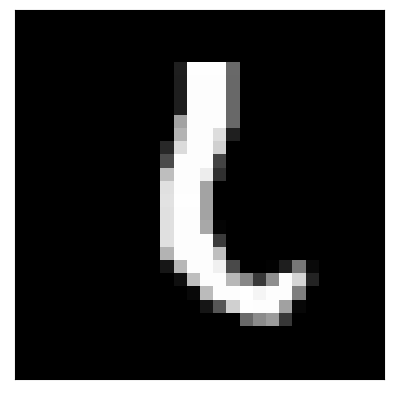}
        \includegraphics[width=\imgcharwidth]{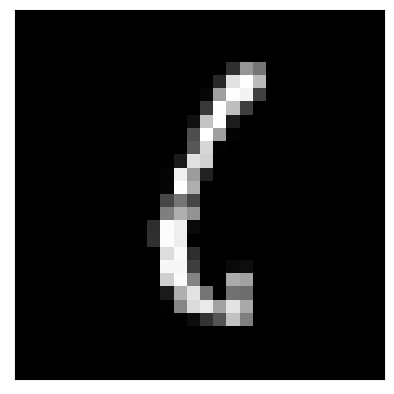}
        &
        \includegraphics[width=\imgcharwidth]{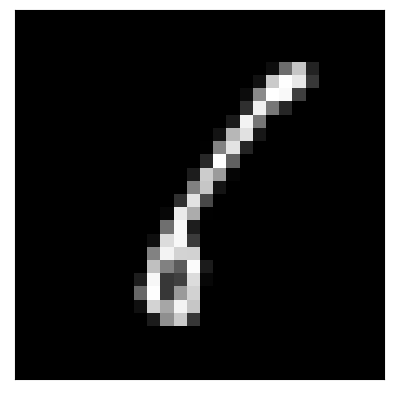}
        \includegraphics[width=\imgcharwidth]{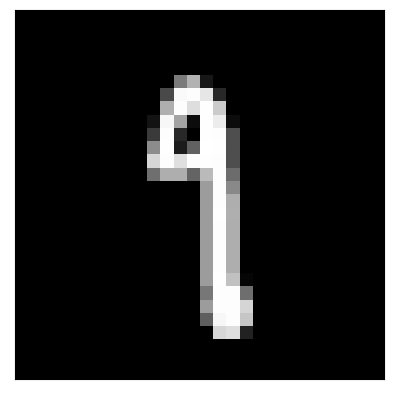}
        \includegraphics[width=\imgcharwidth]{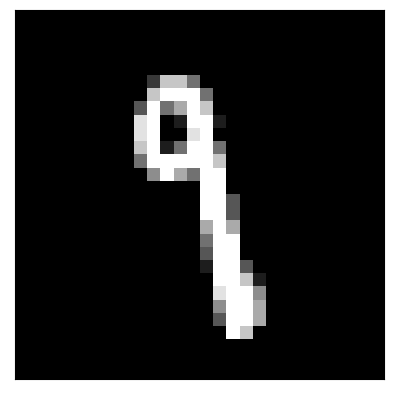}
        \includegraphics[width=\imgcharwidth]{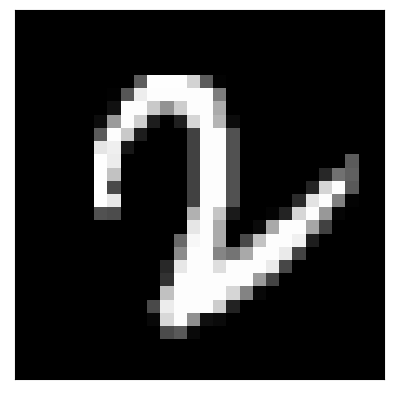}
        \\
        2
        &
        $\varnothing$
        &
        \includegraphics[width=\imgcharwidth]{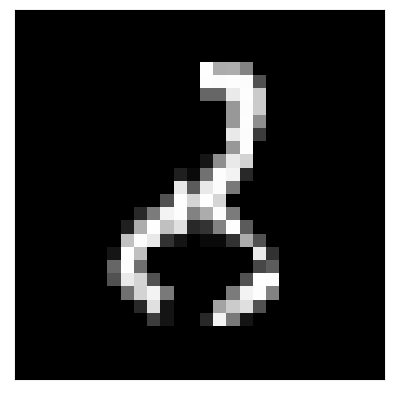}
        \includegraphics[width=\imgcharwidth]{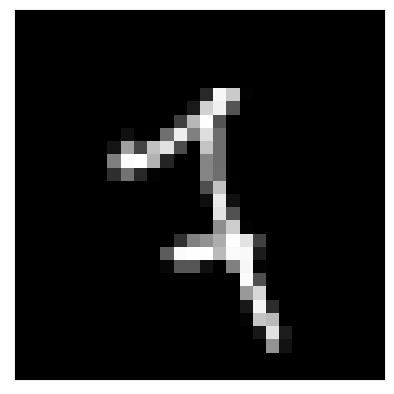}
        \includegraphics[width=\imgcharwidth]{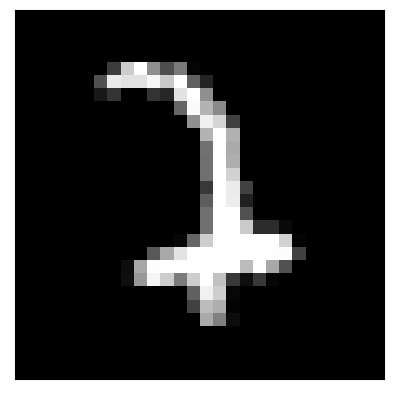}
        \includegraphics[width=\imgcharwidth]{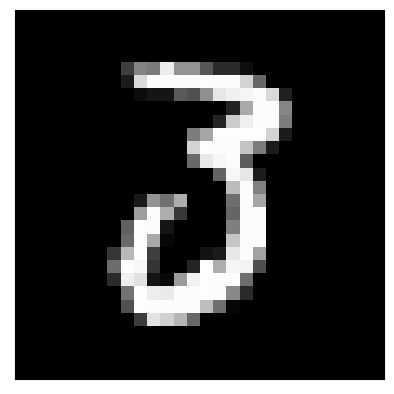}
        \includegraphics[width=\imgcharwidth]{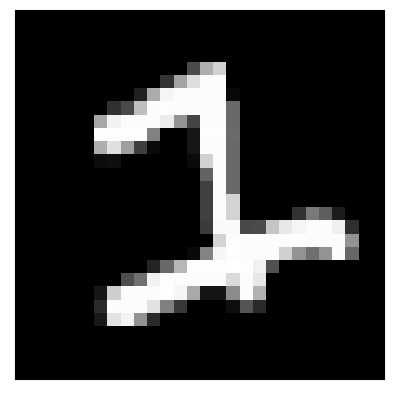}
        \includegraphics[width=\imgcharwidth]{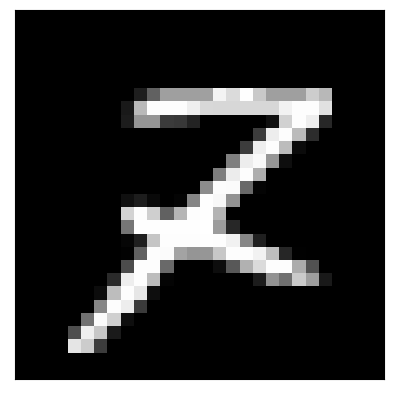}
        \includegraphics[width=\imgcharwidth]{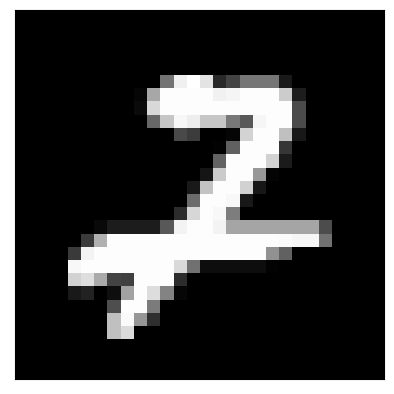}
        \includegraphics[width=\imgcharwidth]{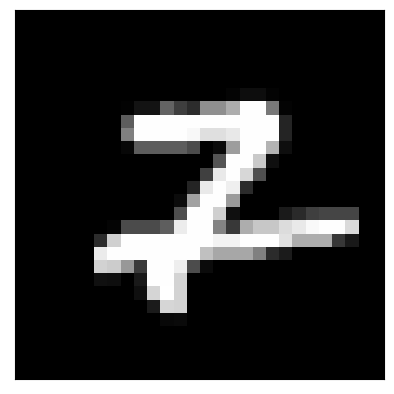}
        \\
        3
        &
        \includegraphics[width=\imgcharwidth]{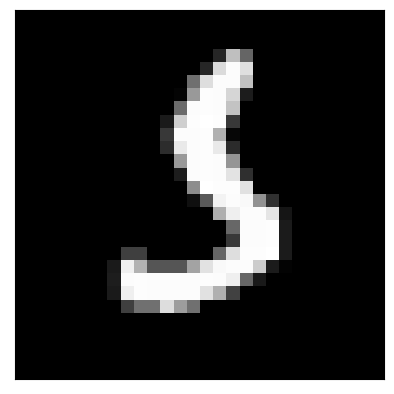}
        \includegraphics[width=\imgcharwidth]{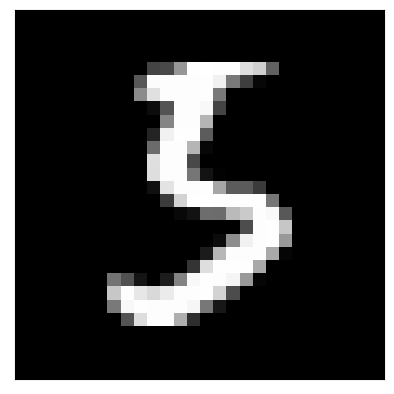}
        \includegraphics[width=\imgcharwidth]{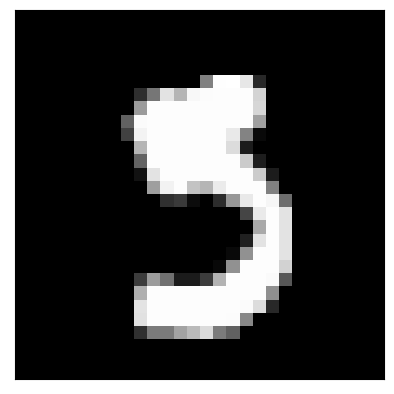}
        \includegraphics[width=\imgcharwidth]{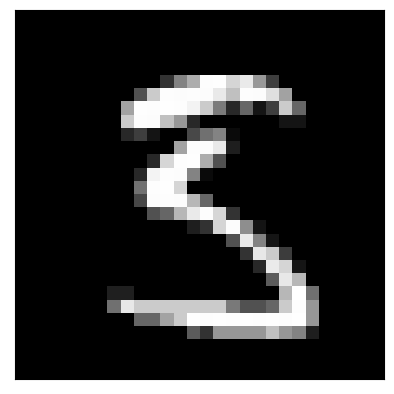}
        \includegraphics[width=\imgcharwidth]{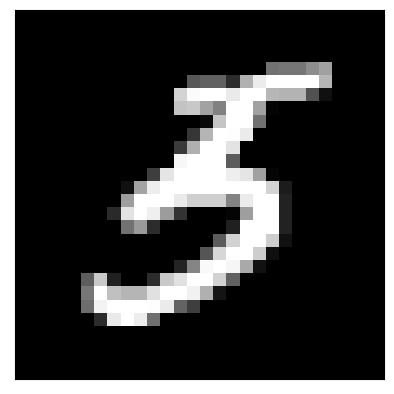}
        &
        \includegraphics[width=\imgcharwidth]{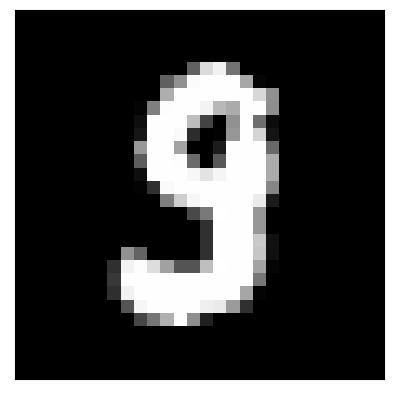}
        \includegraphics[width=\imgcharwidth]{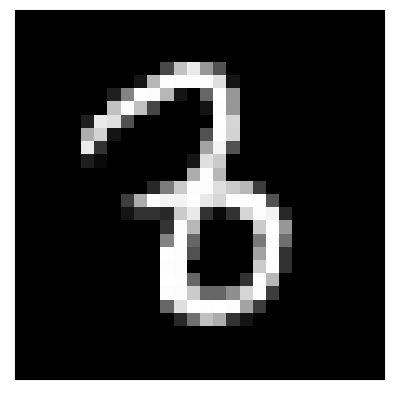}
        \includegraphics[width=\imgcharwidth]{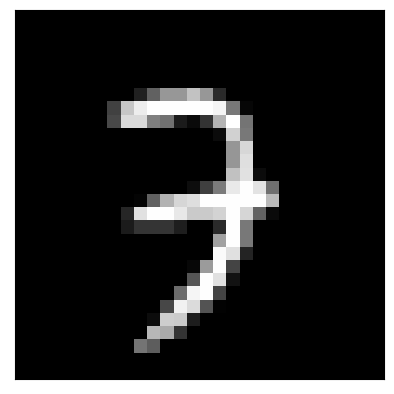}
        \includegraphics[width=\imgcharwidth]{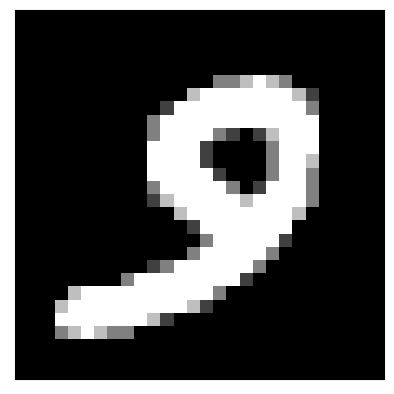}
        \\
        4
        &
        \includegraphics[width=\imgcharwidth]{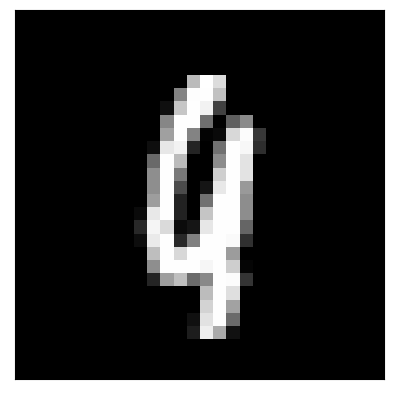}
        \includegraphics[width=\imgcharwidth]{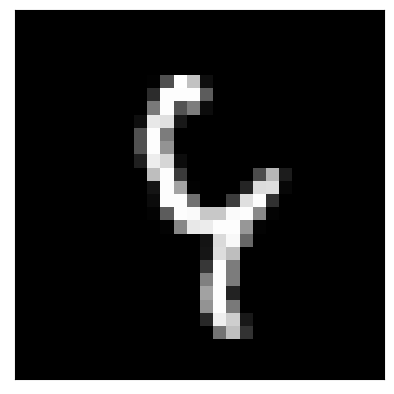}
        \includegraphics[width=\imgcharwidth]{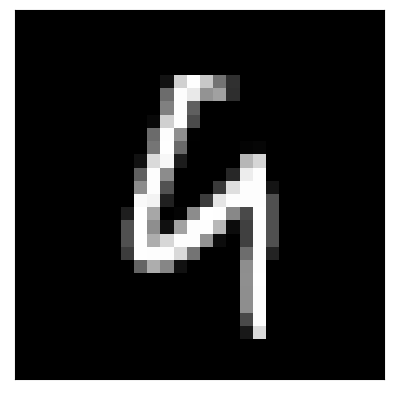}
        &
        \includegraphics[width=\imgcharwidth]{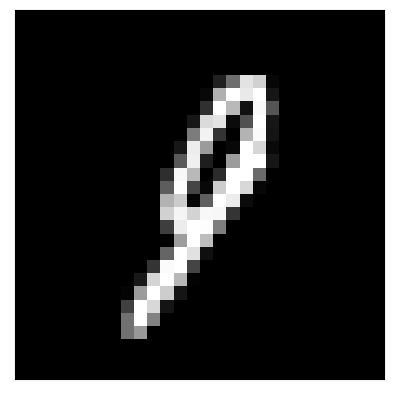}
        \includegraphics[width=\imgcharwidth]{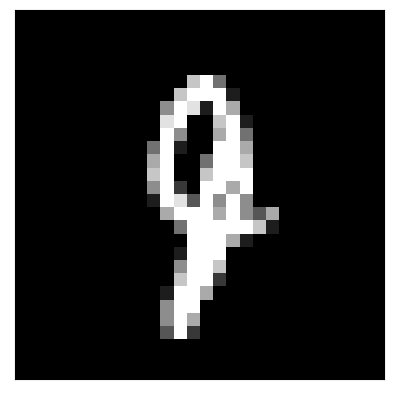}
        \includegraphics[width=\imgcharwidth]{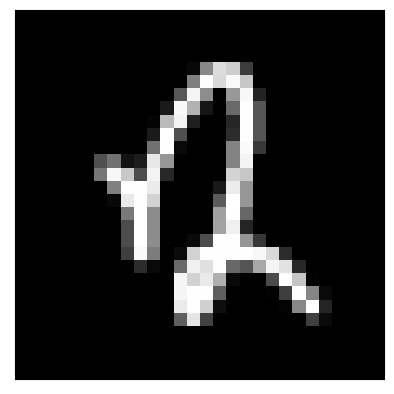}
        \includegraphics[width=\imgcharwidth]{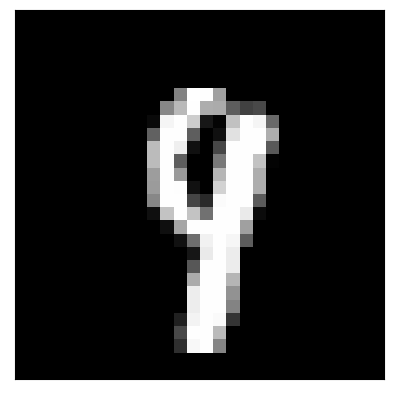}
        \includegraphics[width=\imgcharwidth]{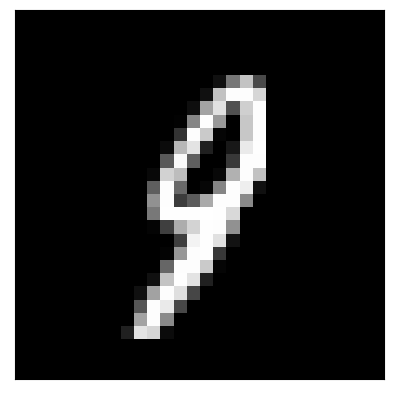}
        \includegraphics[width=\imgcharwidth]{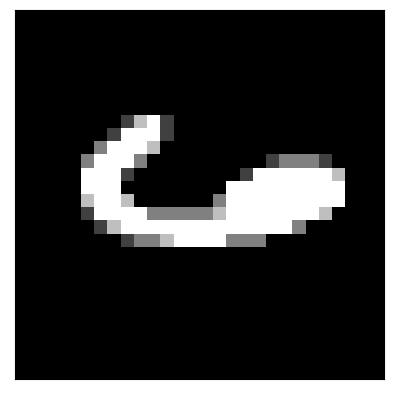}
        \includegraphics[width=\imgcharwidth]{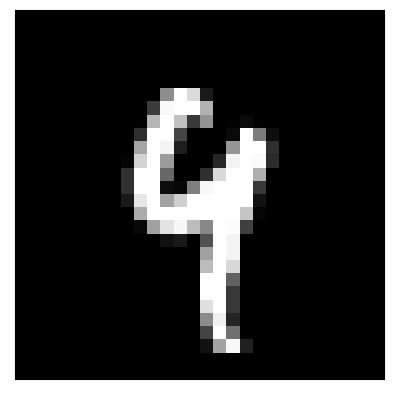}
        \includegraphics[width=\imgcharwidth]{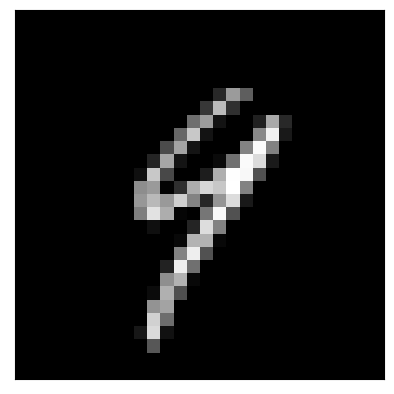}
        \includegraphics[width=\imgcharwidth]{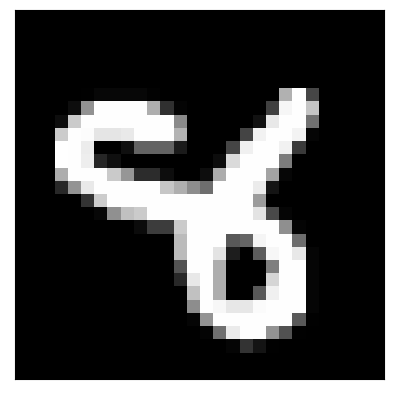}
        \includegraphics[width=\imgcharwidth]{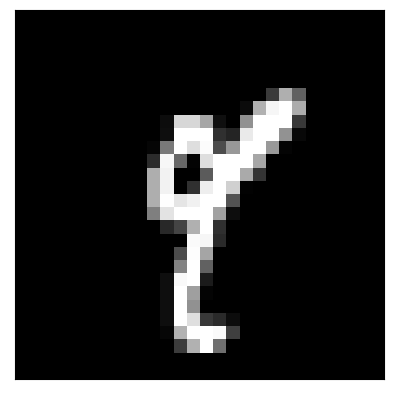}
        \includegraphics[width=\imgcharwidth]{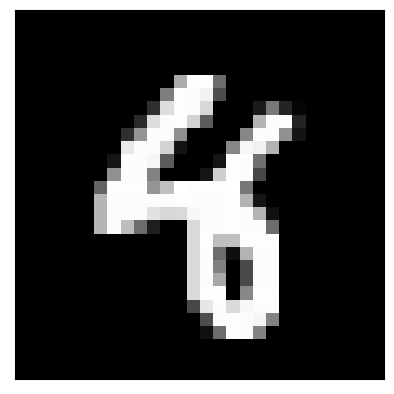}
        \\
        5
        &
        \includegraphics[width=\imgcharwidth]{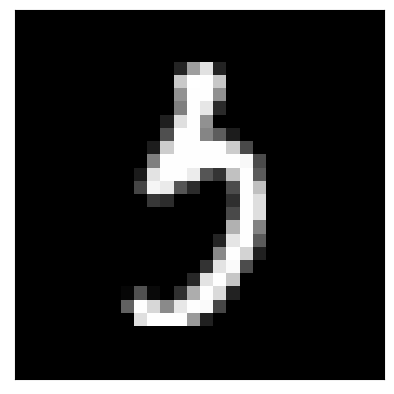}
        &
        \includegraphics[width=\imgcharwidth]{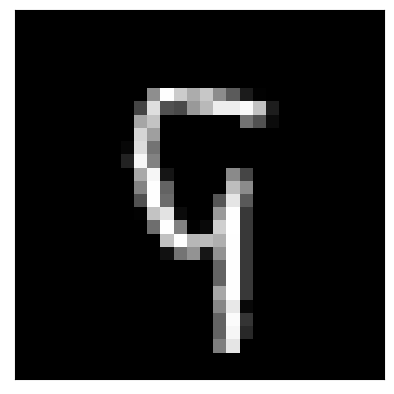}
        \includegraphics[width=\imgcharwidth]{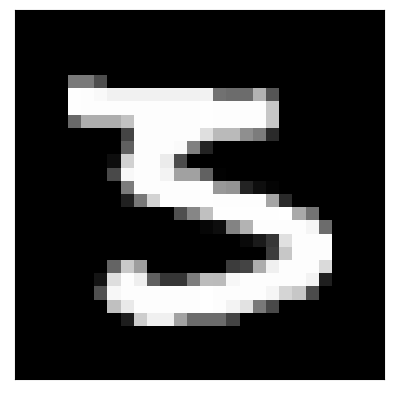}
        \includegraphics[width=\imgcharwidth]{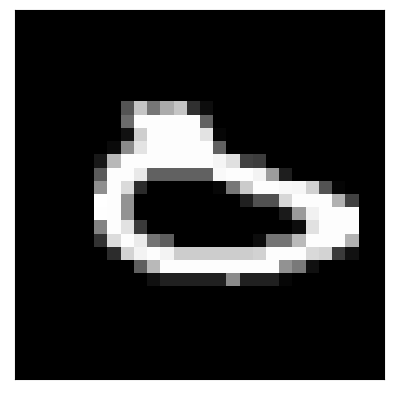}
        \includegraphics[width=\imgcharwidth]{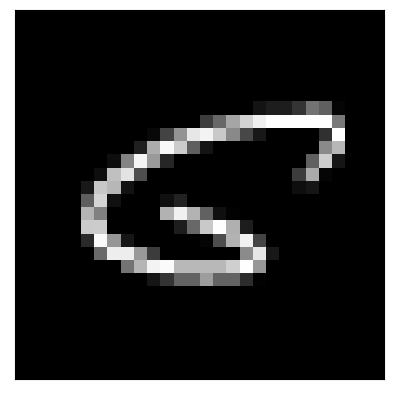}
        \includegraphics[width=\imgcharwidth]{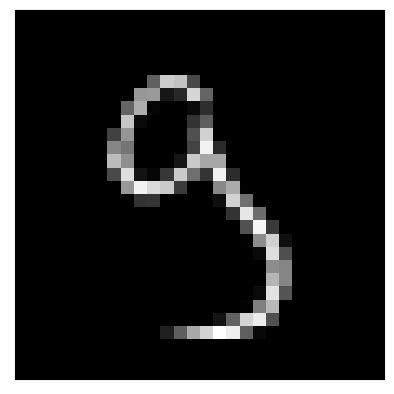}
        \includegraphics[width=\imgcharwidth]{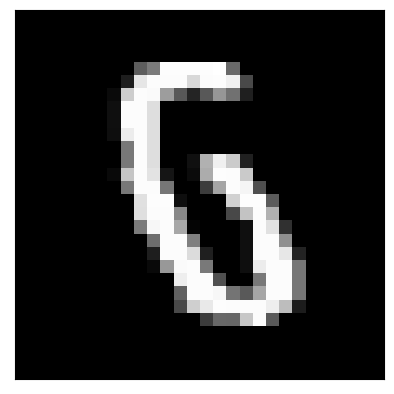}
        \includegraphics[width=\imgcharwidth]{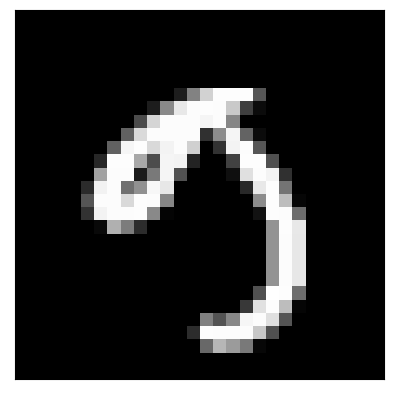}
        \includegraphics[width=\imgcharwidth]{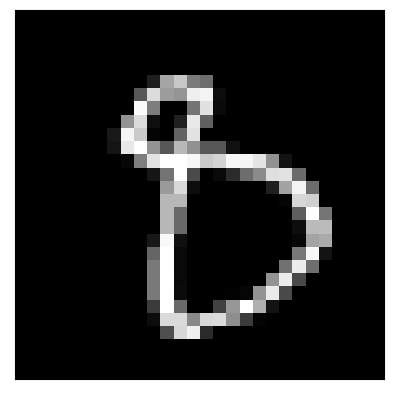}
        \includegraphics[width=\imgcharwidth]{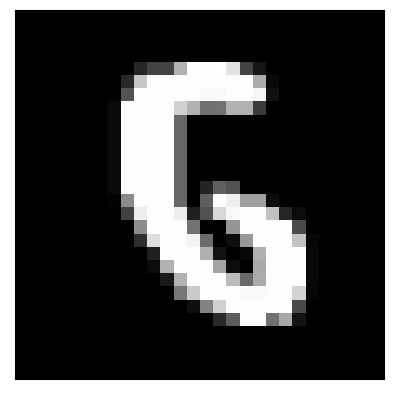}
        \includegraphics[width=\imgcharwidth]{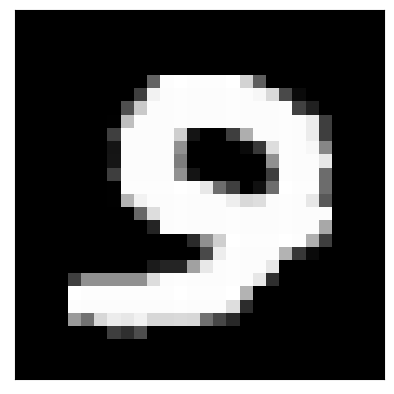}
        \includegraphics[width=\imgcharwidth]{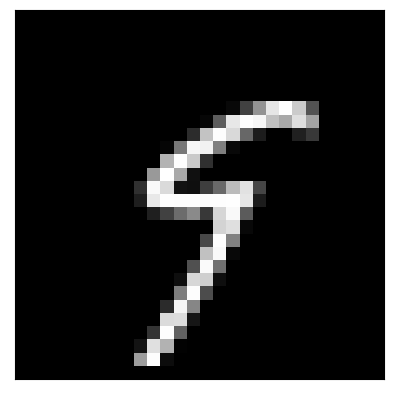}
        \\
        6
        &
        \includegraphics[width=\imgcharwidth]{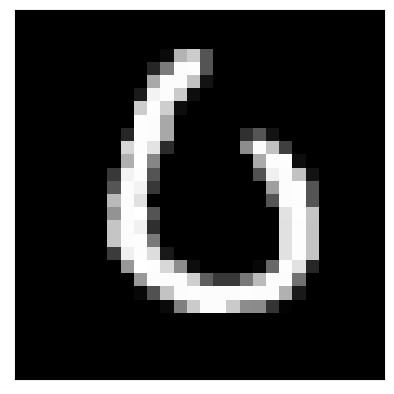}
        \includegraphics[width=\imgcharwidth]{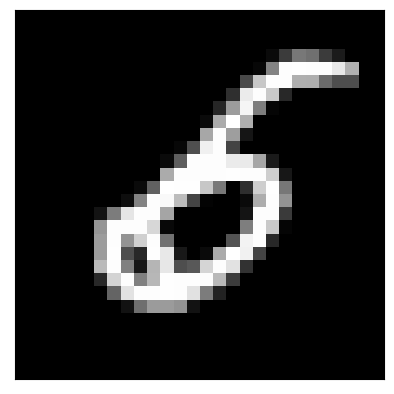}
        \includegraphics[width=\imgcharwidth]{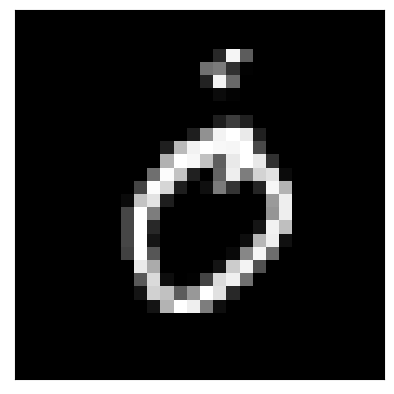}
        &
        \includegraphics[width=\imgcharwidth]{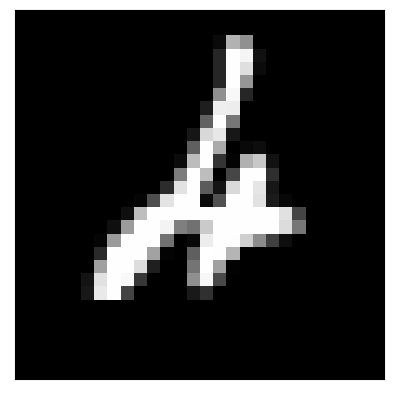}
        \includegraphics[width=\imgcharwidth]{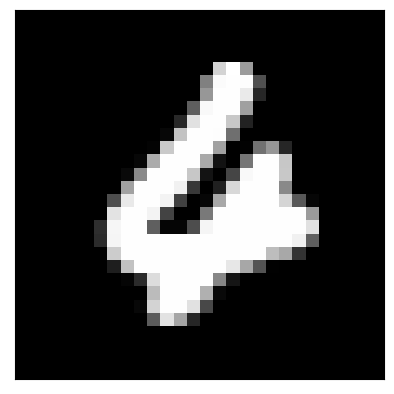}
        \includegraphics[width=\imgcharwidth]{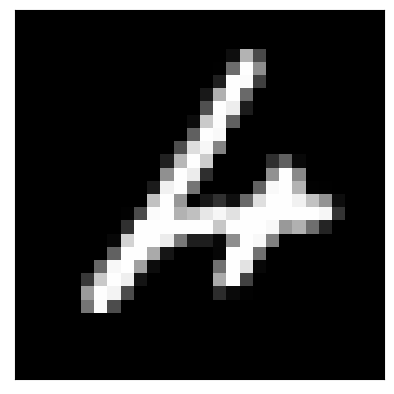}
        \includegraphics[width=\imgcharwidth]{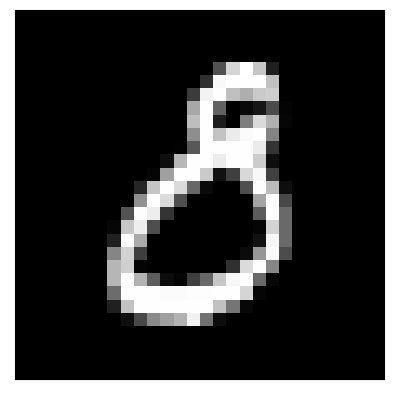}
        \includegraphics[width=\imgcharwidth]{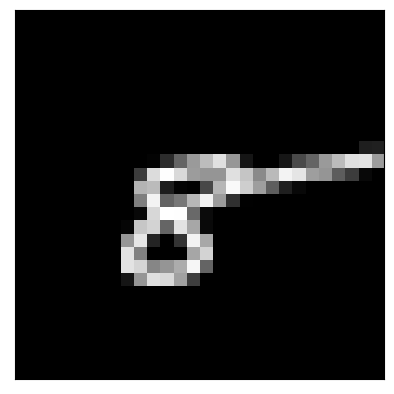}
        \\
        7
        &
        \includegraphics[width=\imgcharwidth]{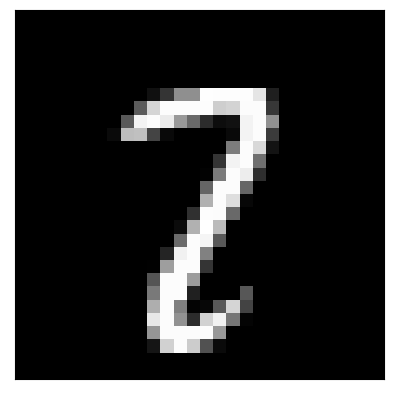}
        \includegraphics[width=\imgcharwidth]{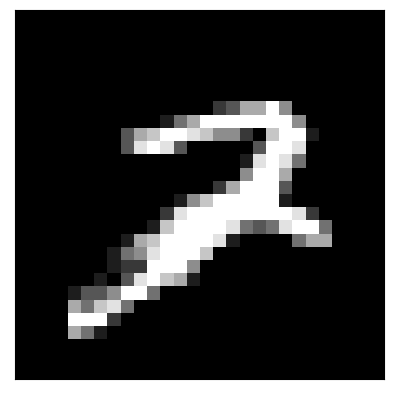}
        \includegraphics[width=\imgcharwidth]{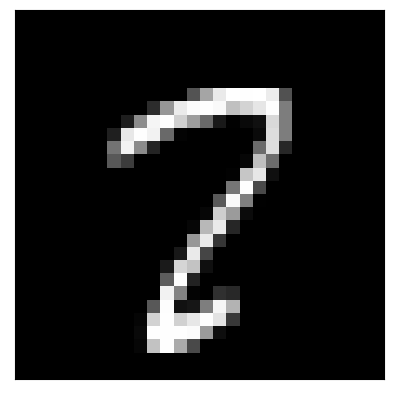}
        \includegraphics[width=\imgcharwidth]{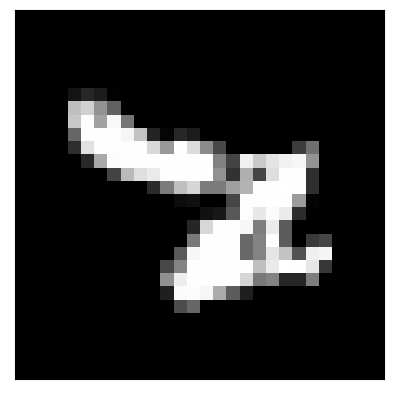}
        &
        \includegraphics[width=\imgcharwidth]{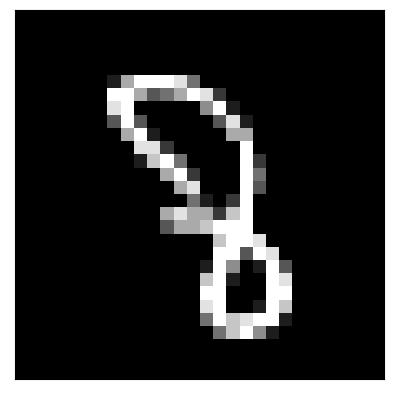}
        \includegraphics[width=\imgcharwidth]{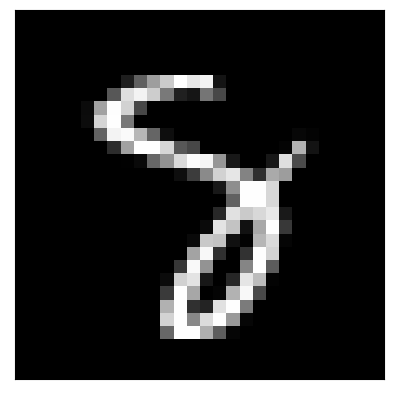}
        \includegraphics[width=\imgcharwidth]{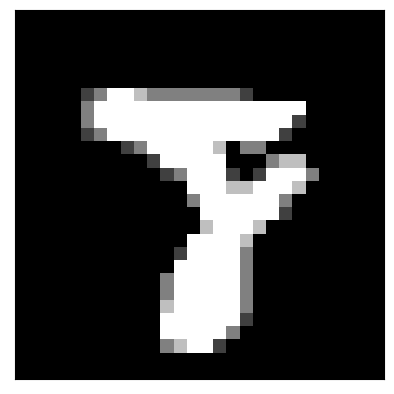}
        \includegraphics[width=\imgcharwidth]{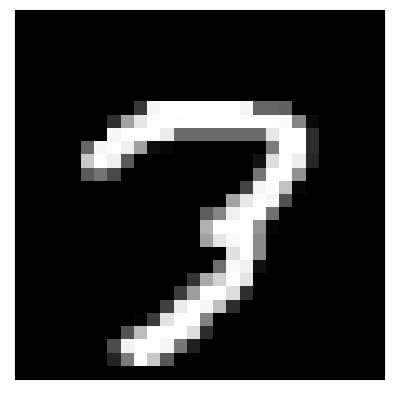}
        \includegraphics[width=\imgcharwidth]{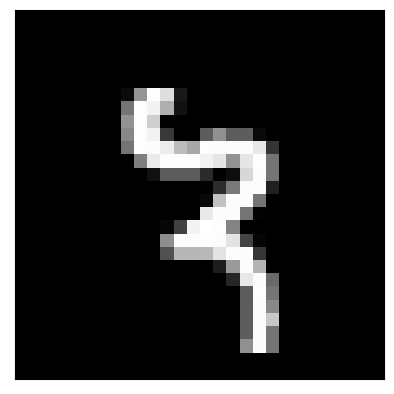}
        \includegraphics[width=\imgcharwidth]{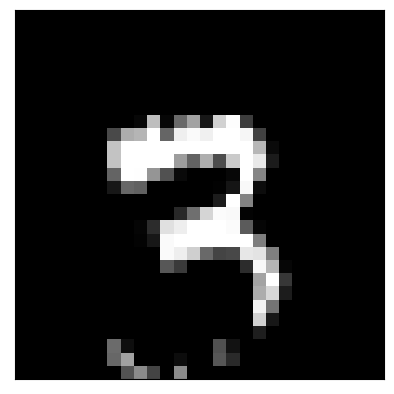}
        \includegraphics[width=\imgcharwidth]{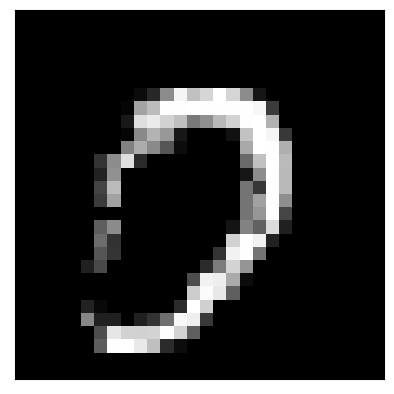}
        \includegraphics[width=\imgcharwidth]{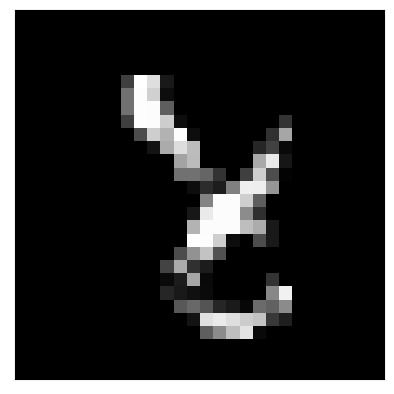}
        \\
        8
        &
        $\varnothing$
        &
        \includegraphics[width=\imgcharwidth]{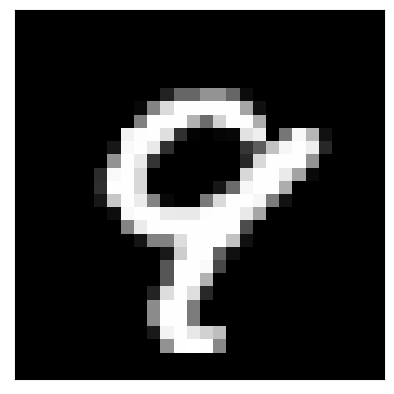}
        \includegraphics[width=\imgcharwidth]{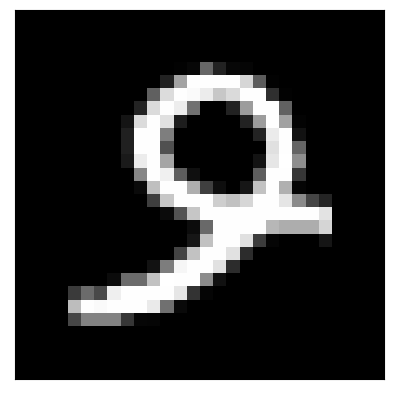}
        \includegraphics[width=\imgcharwidth]{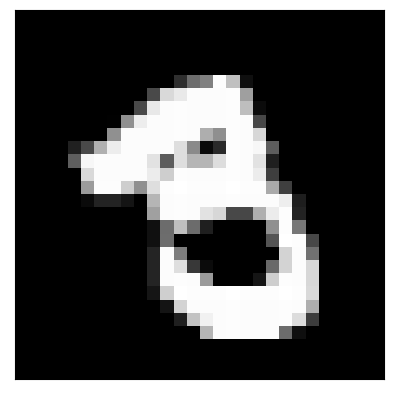}
        \\
        9
        &
        \includegraphics[width=\imgcharwidth]{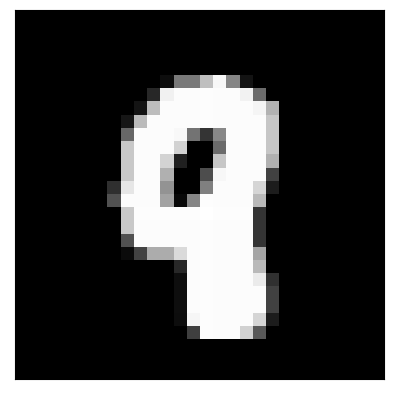}
        \includegraphics[width=\imgcharwidth]{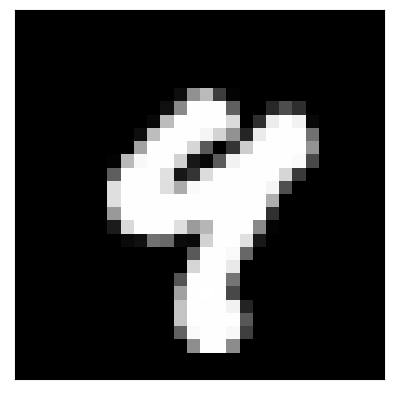}
        &
        \includegraphics[width=\imgcharwidth]{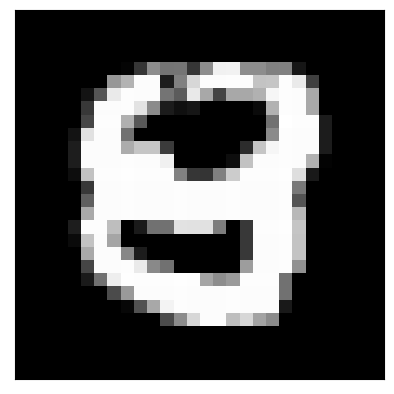}
        \includegraphics[width=\imgcharwidth]{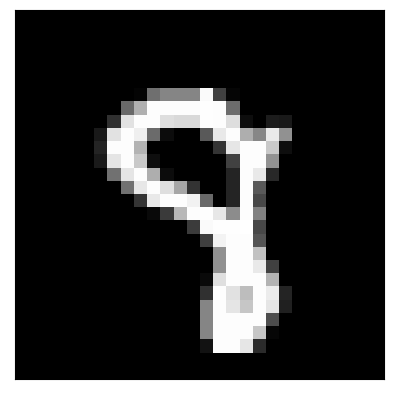}
        \\\midrule
        Tshirt
        &     
        \includegraphics[width=\imgcharwidth]{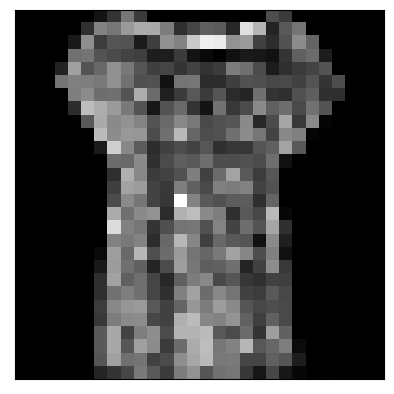}
        \includegraphics[width=\imgcharwidth]{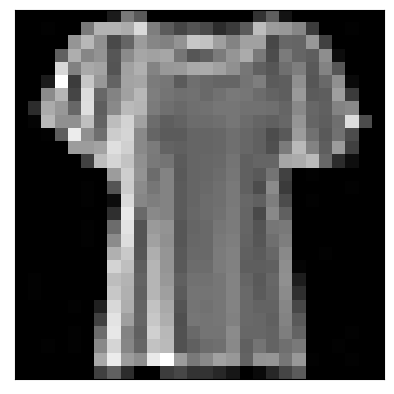}
        &
        \includegraphics[width=\imgcharwidth]{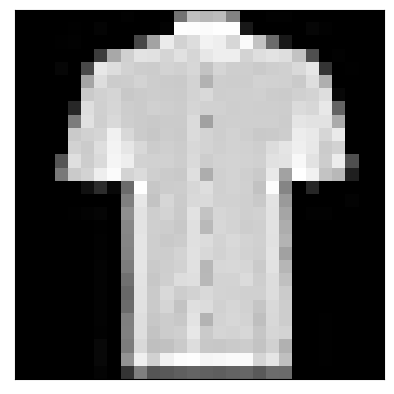}
        \includegraphics[width=\imgcharwidth]{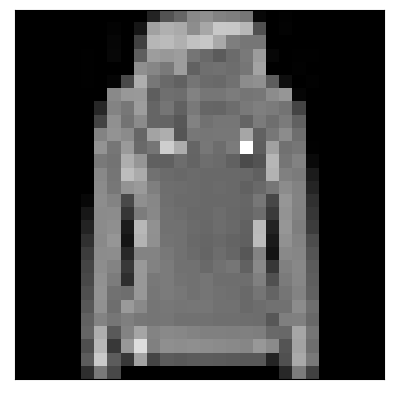}
        \includegraphics[width=\imgcharwidth]{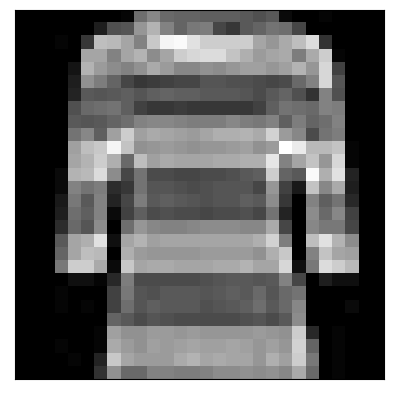}
        \includegraphics[width=\imgcharwidth]{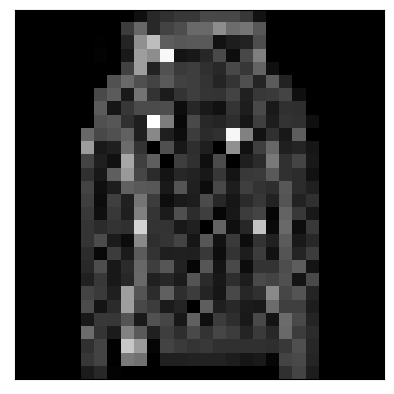}
        \includegraphics[width=\imgcharwidth]{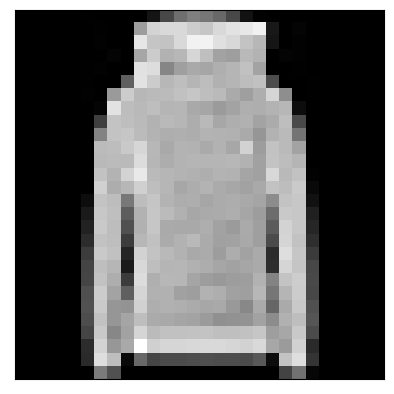}
        \includegraphics[width=\imgcharwidth]{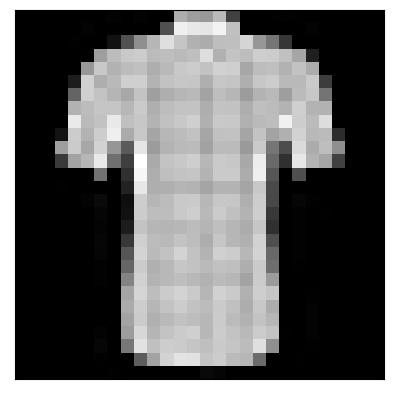}
        \includegraphics[width=\imgcharwidth]{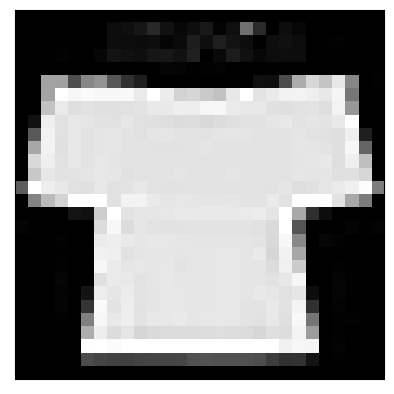}
        \includegraphics[width=\imgcharwidth]{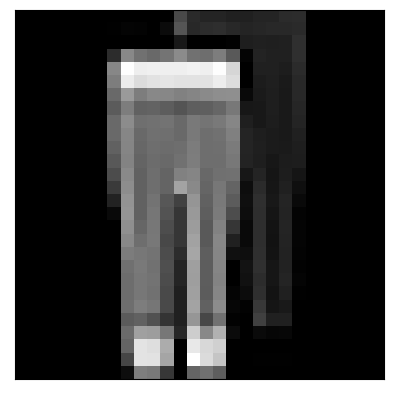}
        \\
        Trouser
        &     
        $\varnothing$
        &
        \includegraphics[width=\imgcharwidth]{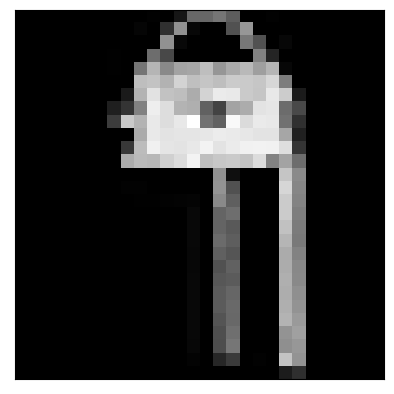}
        \includegraphics[width=\imgcharwidth]{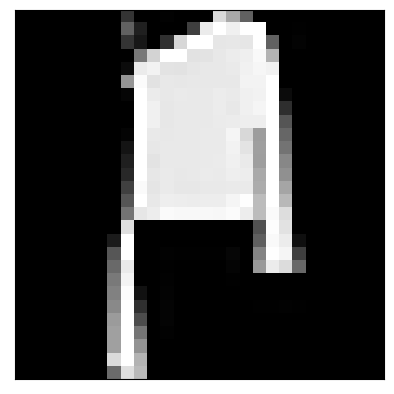}
        \includegraphics[width=\imgcharwidth]{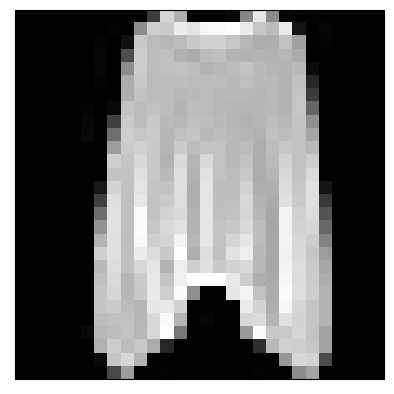}
        \\
        Pullover
        &     
        \includegraphics[width=\imgcharwidth]{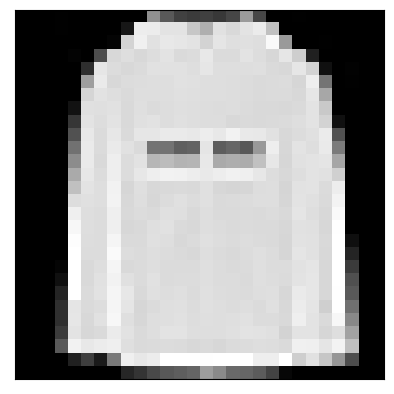}
        \includegraphics[width=\imgcharwidth]{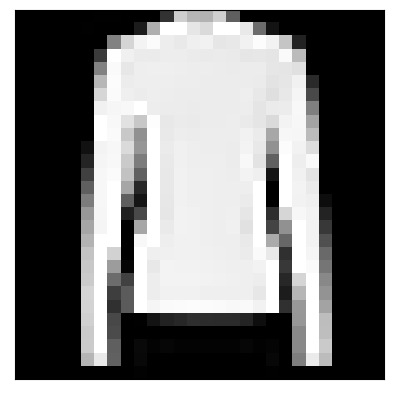}
        \includegraphics[width=\imgcharwidth]{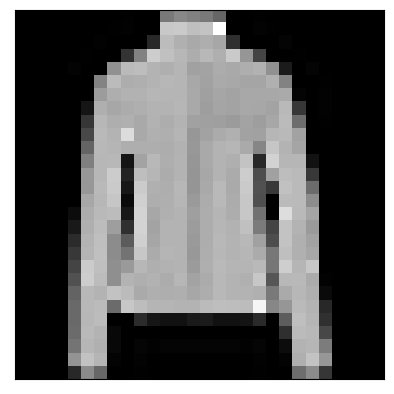}
        \includegraphics[width=\imgcharwidth]{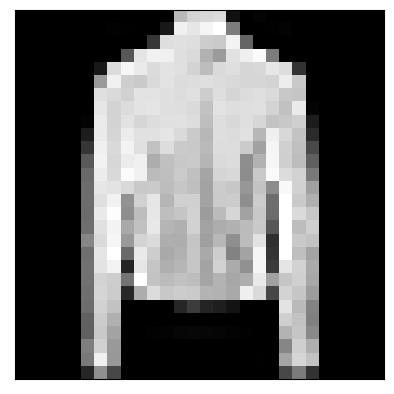}
        \includegraphics[width=\imgcharwidth]{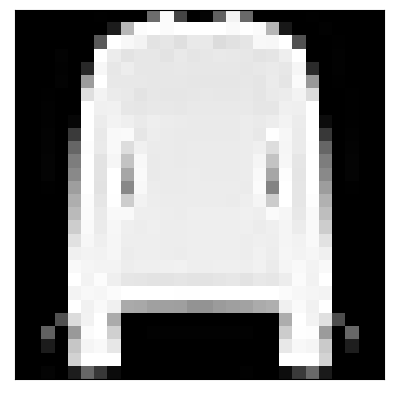}
        \includegraphics[width=\imgcharwidth]{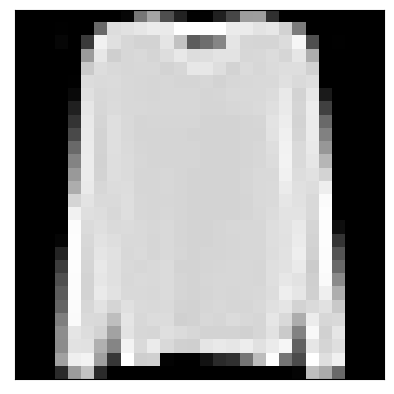}
        \includegraphics[width=\imgcharwidth]{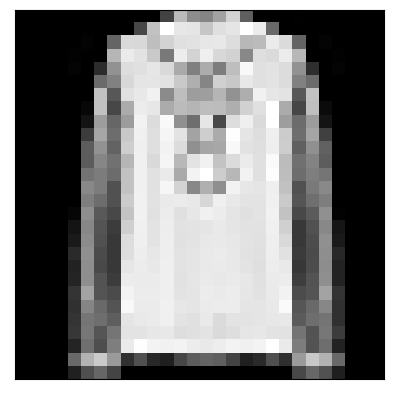}
        \includegraphics[width=\imgcharwidth]{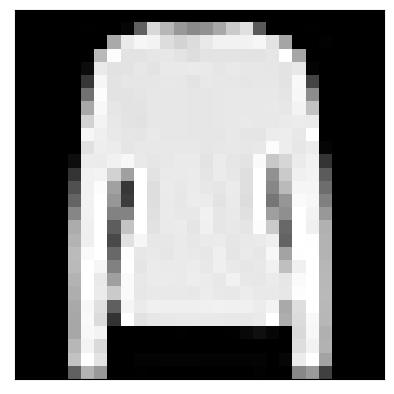}
        \includegraphics[width=\imgcharwidth]{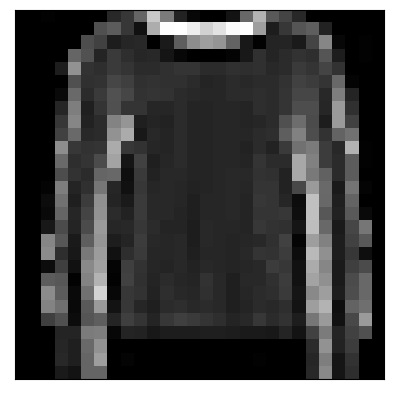}
        &
        \includegraphics[width=\imgcharwidth]{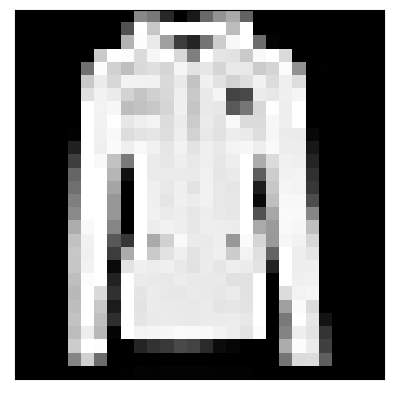}
        \\
        Dress
        &     
        \includegraphics[width=\imgcharwidth]{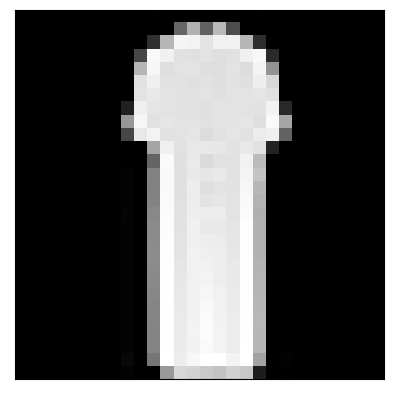}
        \includegraphics[width=\imgcharwidth]{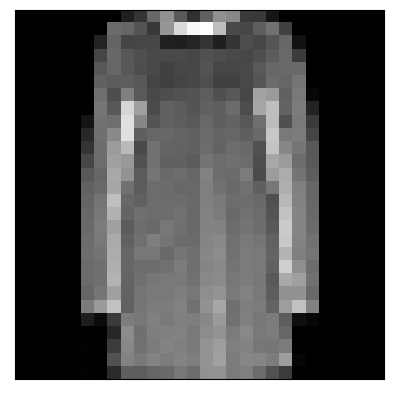}
        \includegraphics[width=\imgcharwidth]{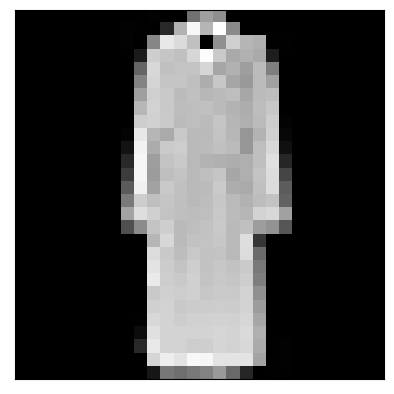}
        \includegraphics[width=\imgcharwidth]{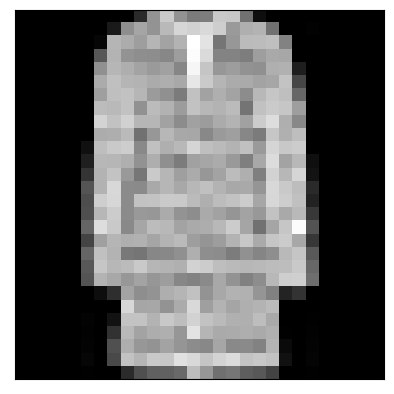}
        &
        \includegraphics[width=\imgcharwidth]{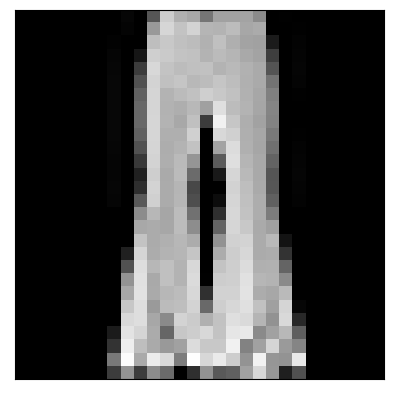}
        \includegraphics[width=\imgcharwidth]{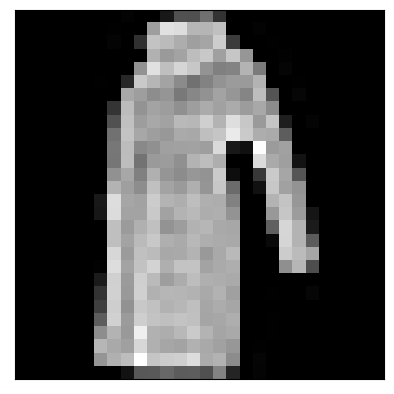}
        \includegraphics[width=\imgcharwidth]{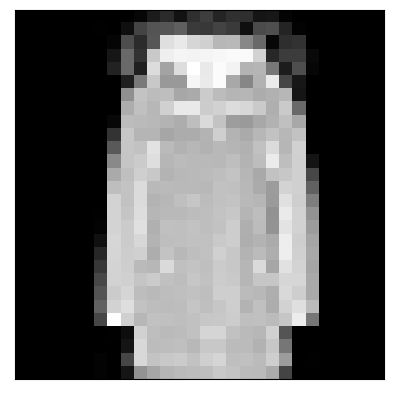}
        \includegraphics[width=\imgcharwidth]{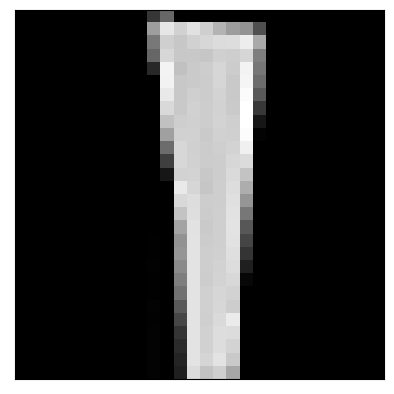}
        \includegraphics[width=\imgcharwidth]{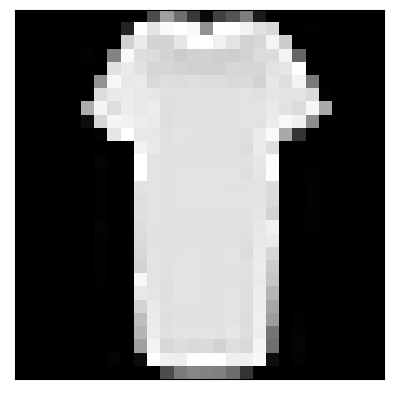}
        \includegraphics[width=\imgcharwidth]{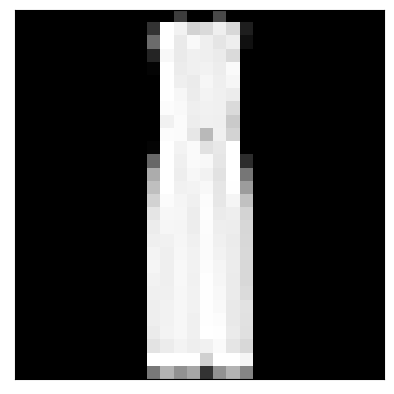}
        \\
        Coat
        &     
        \includegraphics[width=\imgcharwidth]{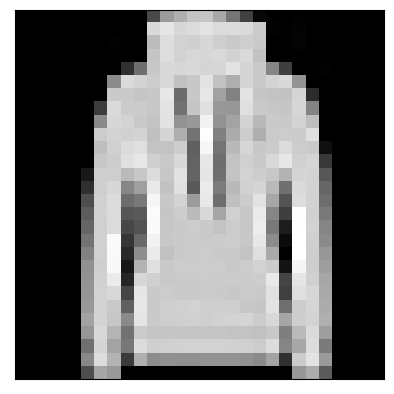}
        \includegraphics[width=\imgcharwidth]{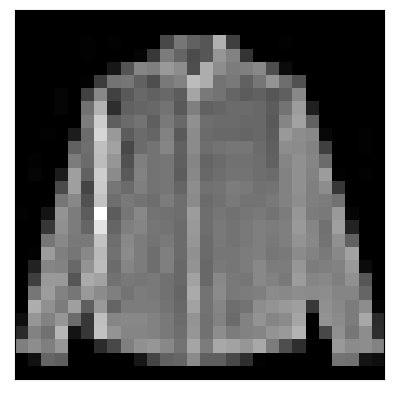}
        \includegraphics[width=\imgcharwidth]{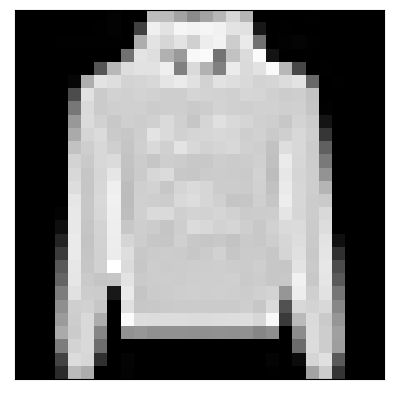}
        \includegraphics[width=\imgcharwidth]{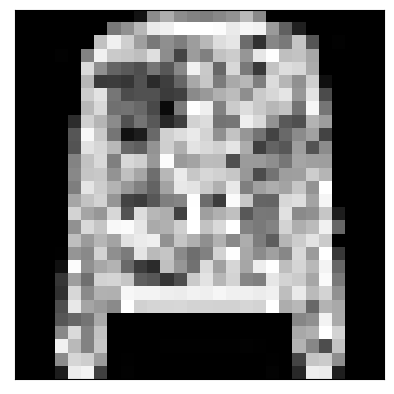}
        \includegraphics[width=\imgcharwidth]{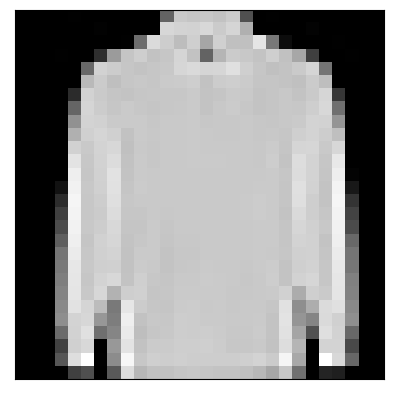}
        \includegraphics[width=\imgcharwidth]{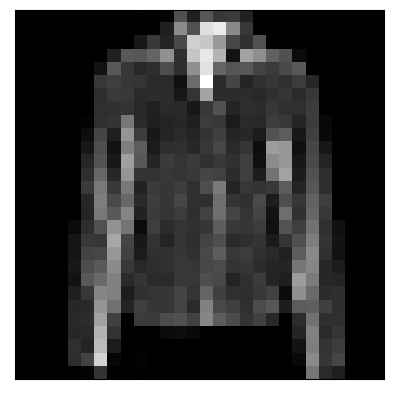}
        \includegraphics[width=\imgcharwidth]{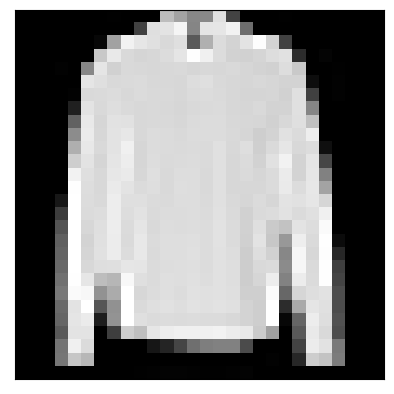}
        \includegraphics[width=\imgcharwidth]{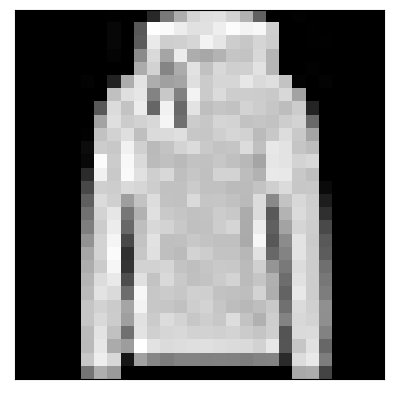}
        \includegraphics[width=\imgcharwidth]{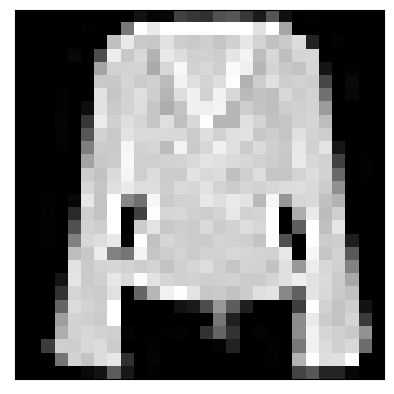}
        &
        \includegraphics[width=\imgcharwidth]{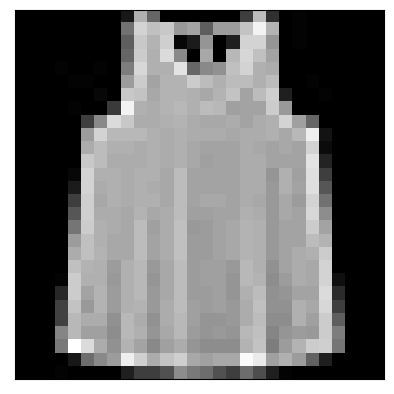}
        \\
        Sandal
        &     
        \includegraphics[width=\imgcharwidth]{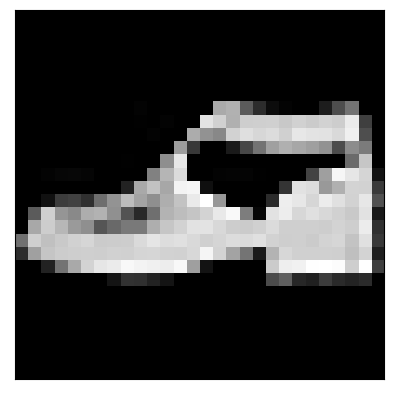}
        &
        \includegraphics[width=\imgcharwidth]{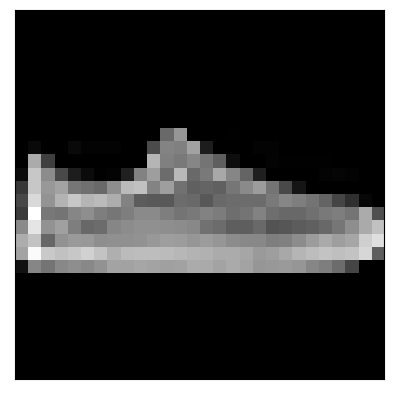}
        \includegraphics[width=\imgcharwidth]{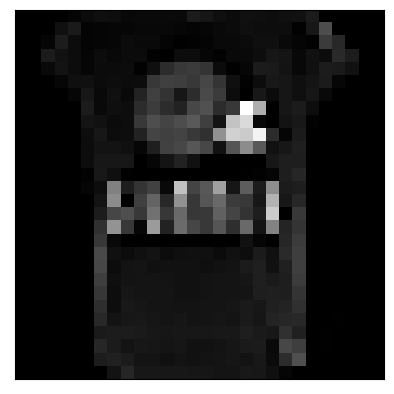}
        \includegraphics[width=\imgcharwidth]{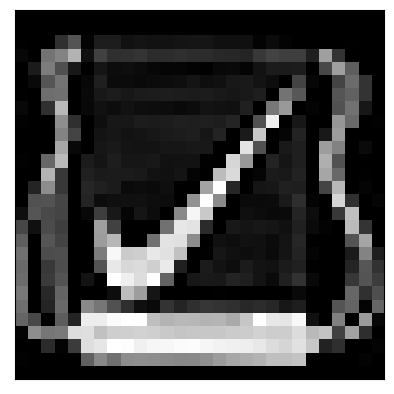}
        \includegraphics[width=\imgcharwidth]{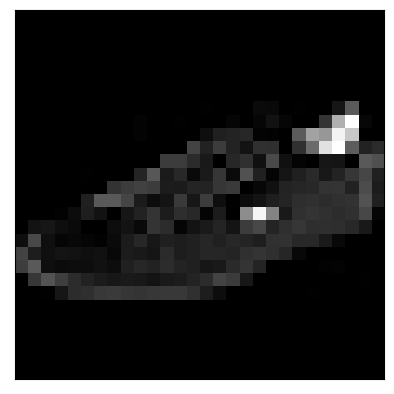}
        \includegraphics[width=\imgcharwidth]{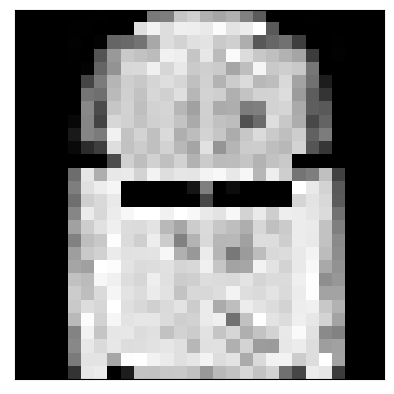}
        \includegraphics[width=\imgcharwidth]{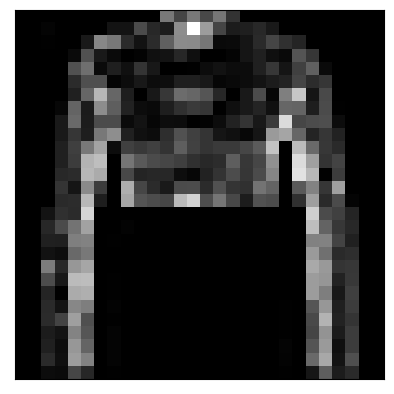}
        \includegraphics[width=\imgcharwidth]{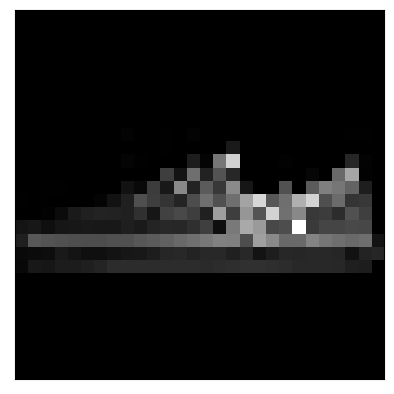}
        \includegraphics[width=\imgcharwidth]{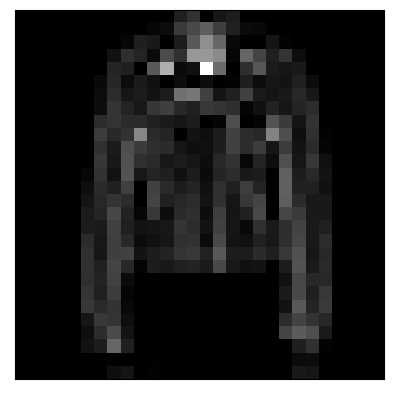}
        \includegraphics[width=\imgcharwidth]{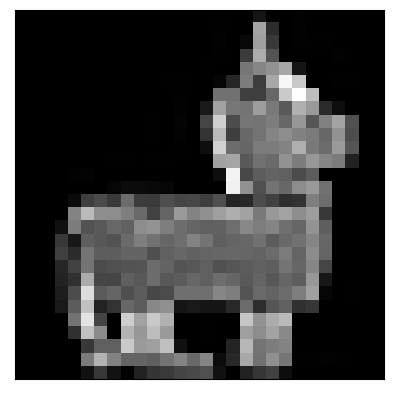}
        \\
        Shirt
        &     
        \includegraphics[width=\imgcharwidth]{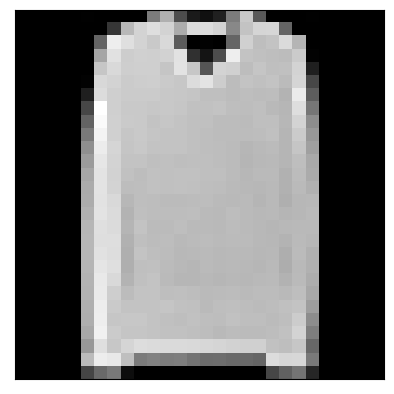}
        \includegraphics[width=\imgcharwidth]{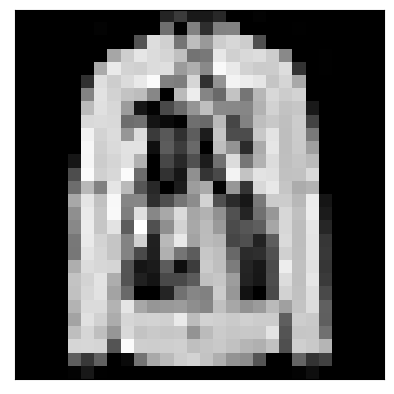}
        \includegraphics[width=\imgcharwidth]{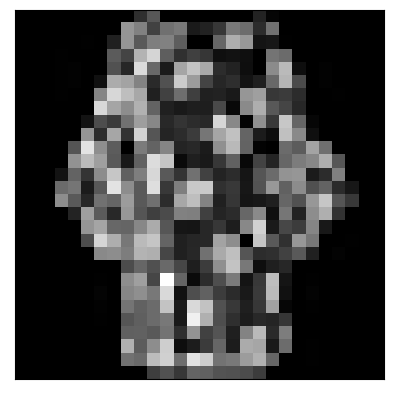}
        &
        \includegraphics[width=\imgcharwidth]{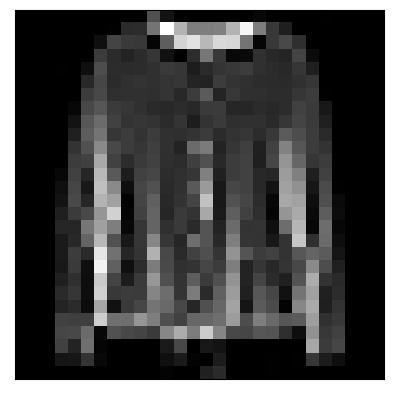}
        \includegraphics[width=\imgcharwidth]{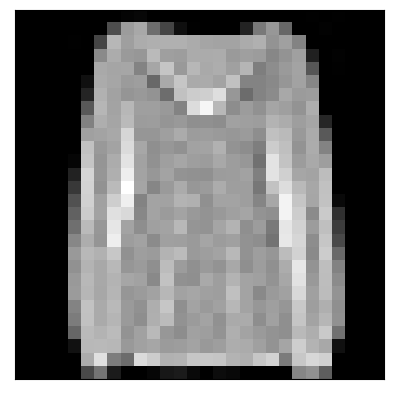}
        \includegraphics[width=\imgcharwidth]{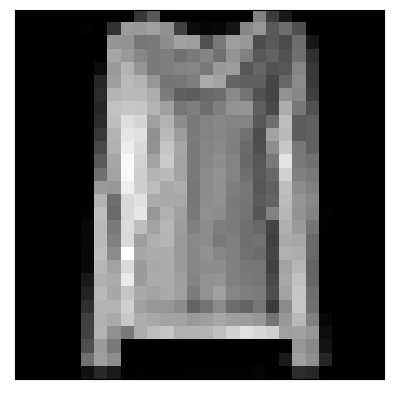}
        \includegraphics[width=\imgcharwidth]{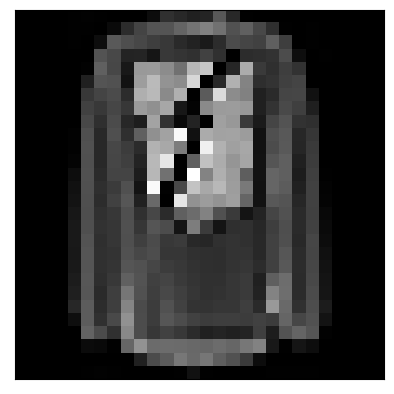}
        \includegraphics[width=\imgcharwidth]{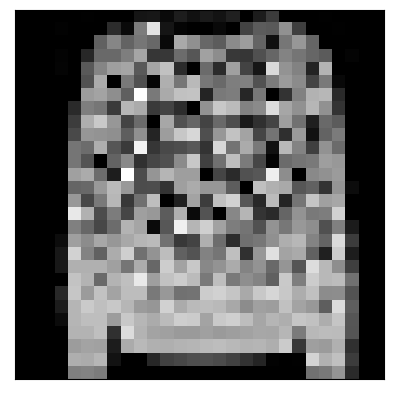}
        \includegraphics[width=\imgcharwidth]{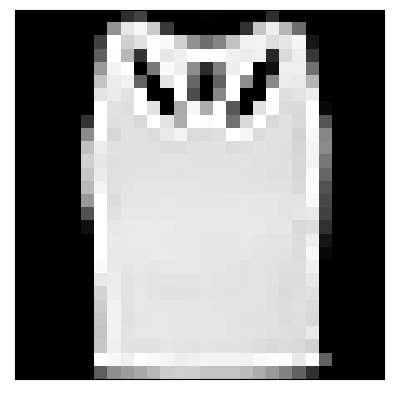}
        \includegraphics[width=\imgcharwidth]{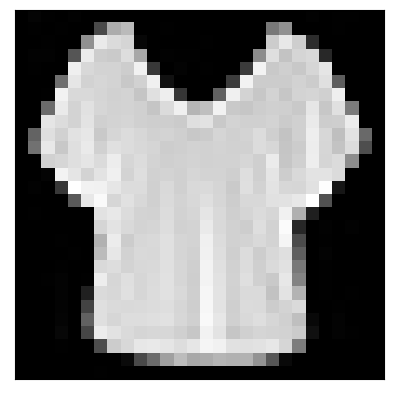}
        \\
        Sneaker
        &     
        \includegraphics[width=\imgcharwidth]{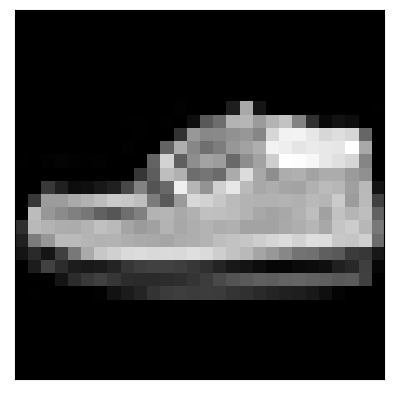}
        \includegraphics[width=\imgcharwidth]{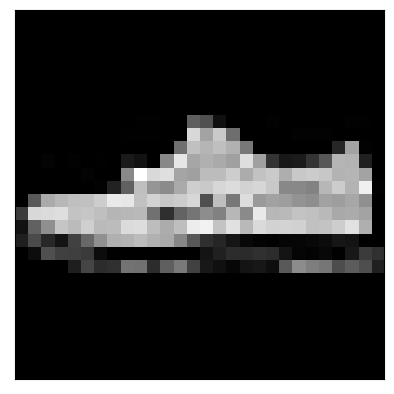}
        &
        \includegraphics[width=\imgcharwidth]{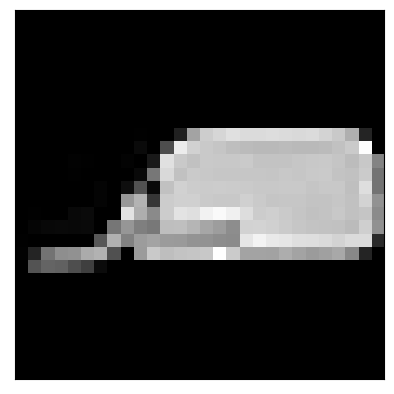}
        \includegraphics[width=\imgcharwidth]{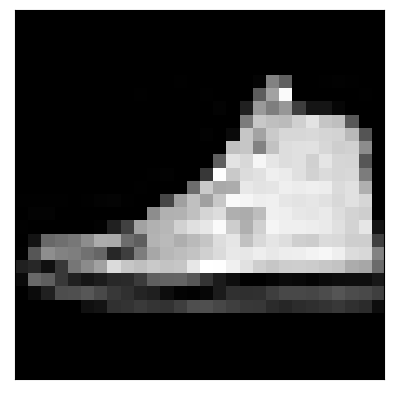}
        \includegraphics[width=\imgcharwidth]{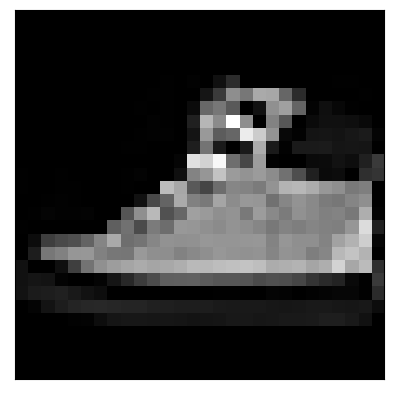}
        \includegraphics[width=\imgcharwidth]{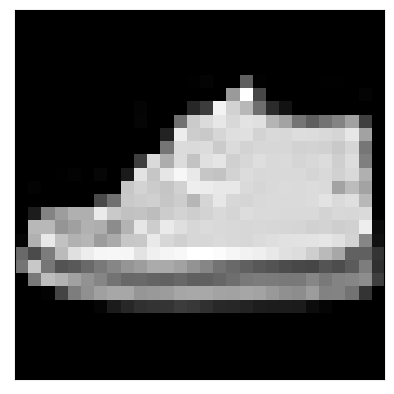}
        \includegraphics[width=\imgcharwidth]{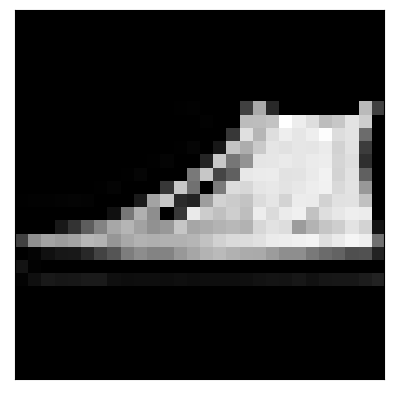}
        \includegraphics[width=\imgcharwidth]{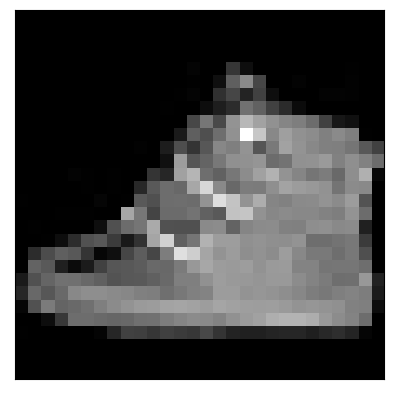}
        \includegraphics[width=\imgcharwidth]{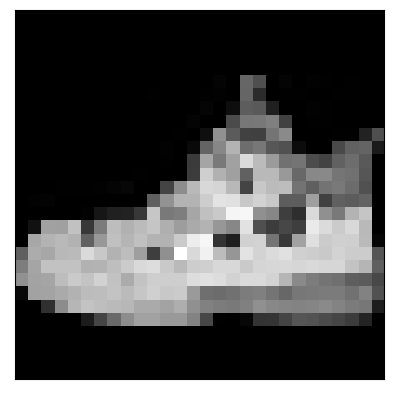}
        \includegraphics[width=\imgcharwidth]{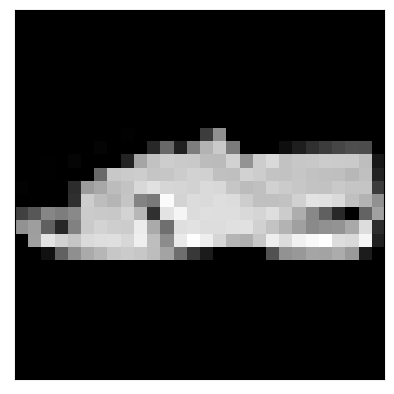}
        \\
        Bag
        &     
        \includegraphics[width=\imgcharwidth]{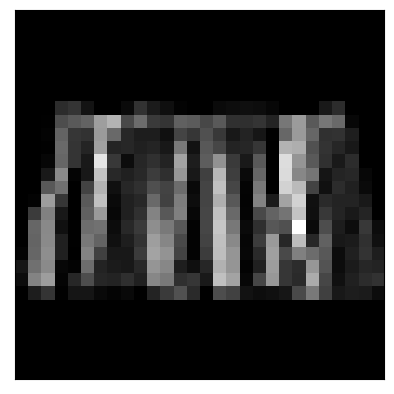}
        &
        \includegraphics[width=\imgcharwidth]{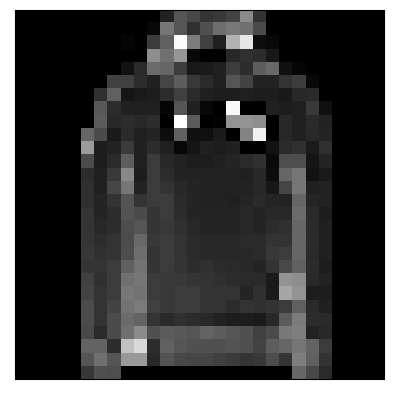}
        \includegraphics[width=\imgcharwidth]{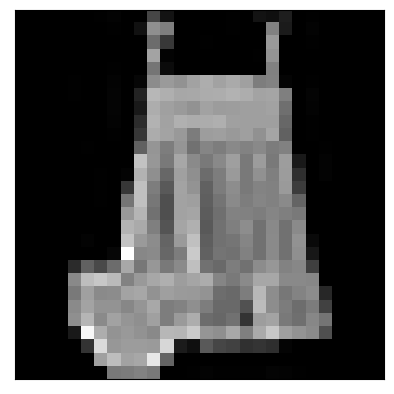}
        \includegraphics[width=\imgcharwidth]{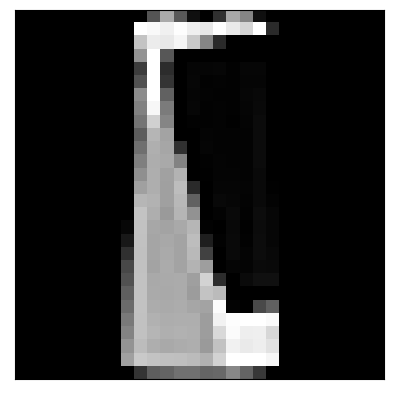}
        \includegraphics[width=\imgcharwidth]{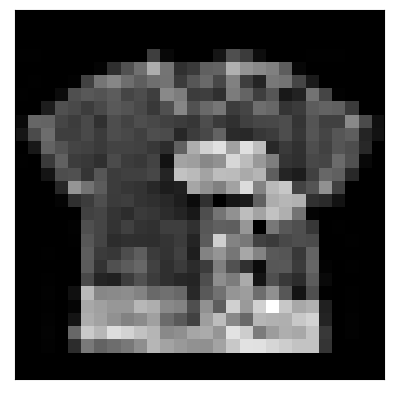}
        \includegraphics[width=\imgcharwidth]{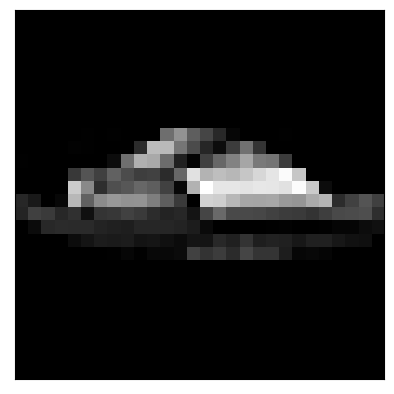}
        \includegraphics[width=\imgcharwidth]{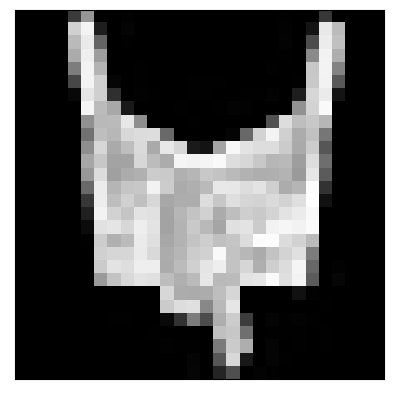}
        \includegraphics[width=\imgcharwidth]{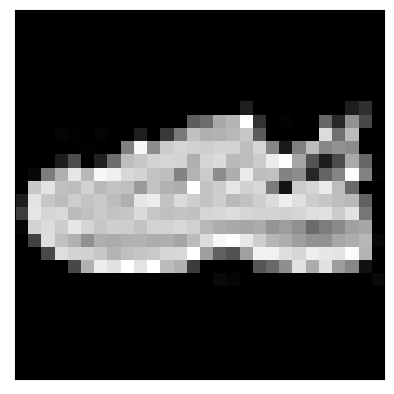}
        \includegraphics[width=\imgcharwidth]{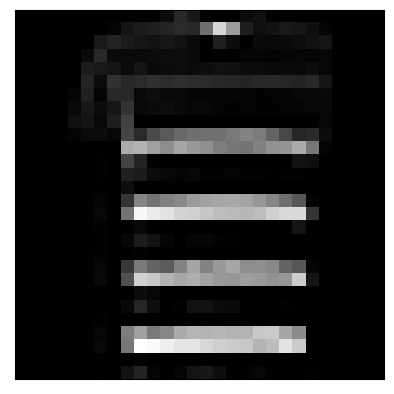}
        \includegraphics[width=\imgcharwidth]{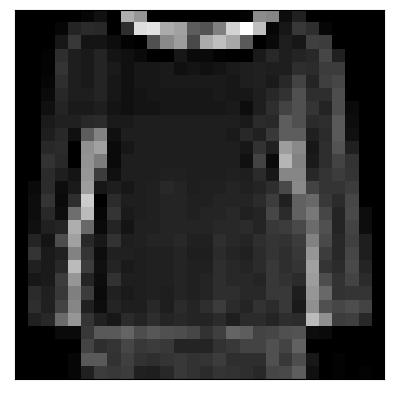}
        \\
        Boot
        &     
        \includegraphics[width=\imgcharwidth]{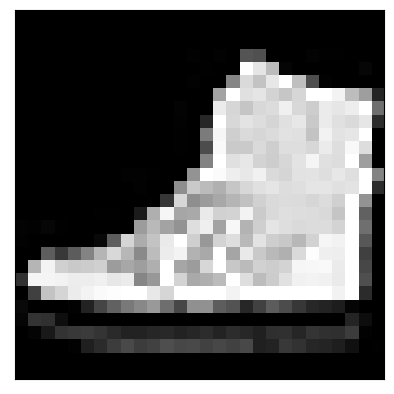}
        \includegraphics[width=\imgcharwidth]{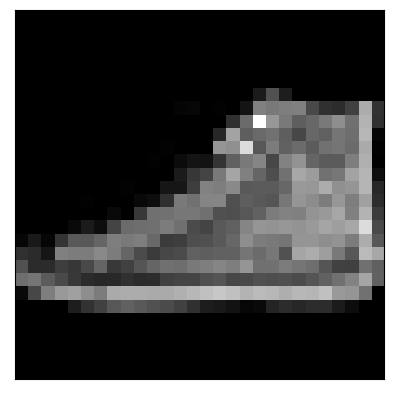}
        \includegraphics[width=\imgcharwidth]{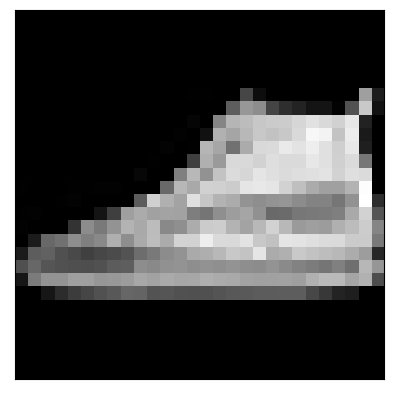}
        \includegraphics[width=\imgcharwidth]{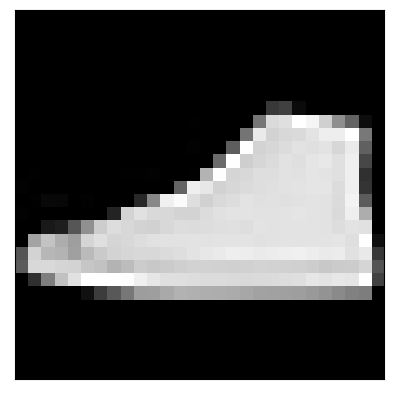}
        \includegraphics[width=\imgcharwidth]{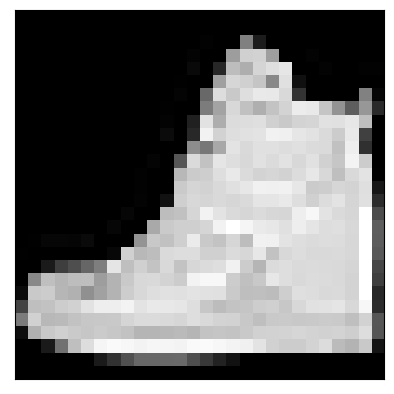}
        \includegraphics[width=\imgcharwidth]{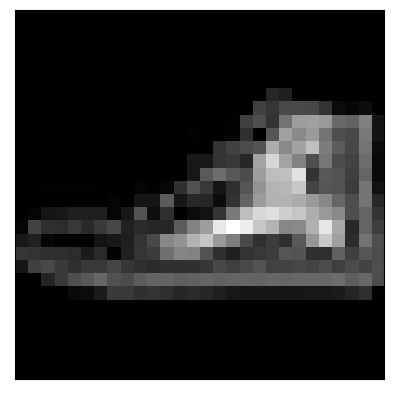}
        \includegraphics[width=\imgcharwidth]{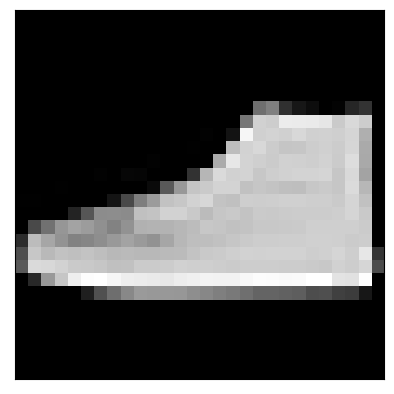}
        \includegraphics[width=\imgcharwidth]{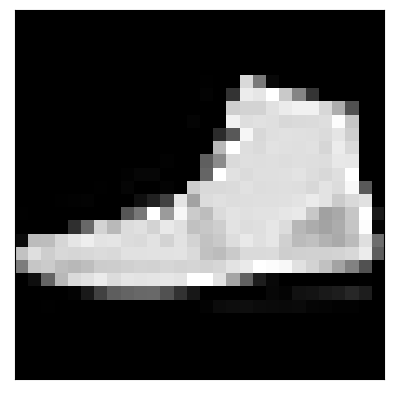}
        \includegraphics[width=\imgcharwidth]{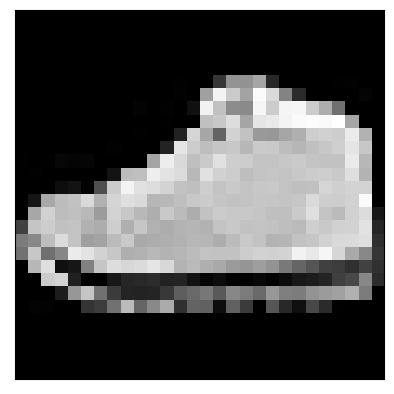}
        \includegraphics[width=\imgcharwidth]{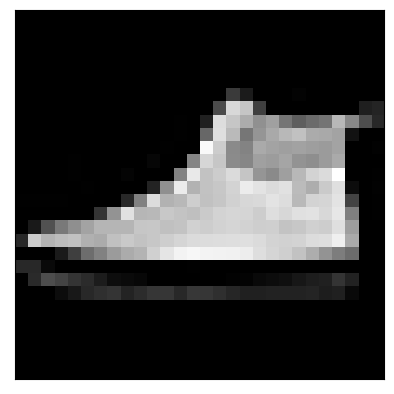}
        &
        $\varnothing$
        \\
        \bottomrule
    \end{tabular}
\end{table}

\clearpage

\begin{figure}[!htb]
    \centering
    \includegraphics[width=0.95\columnwidth]{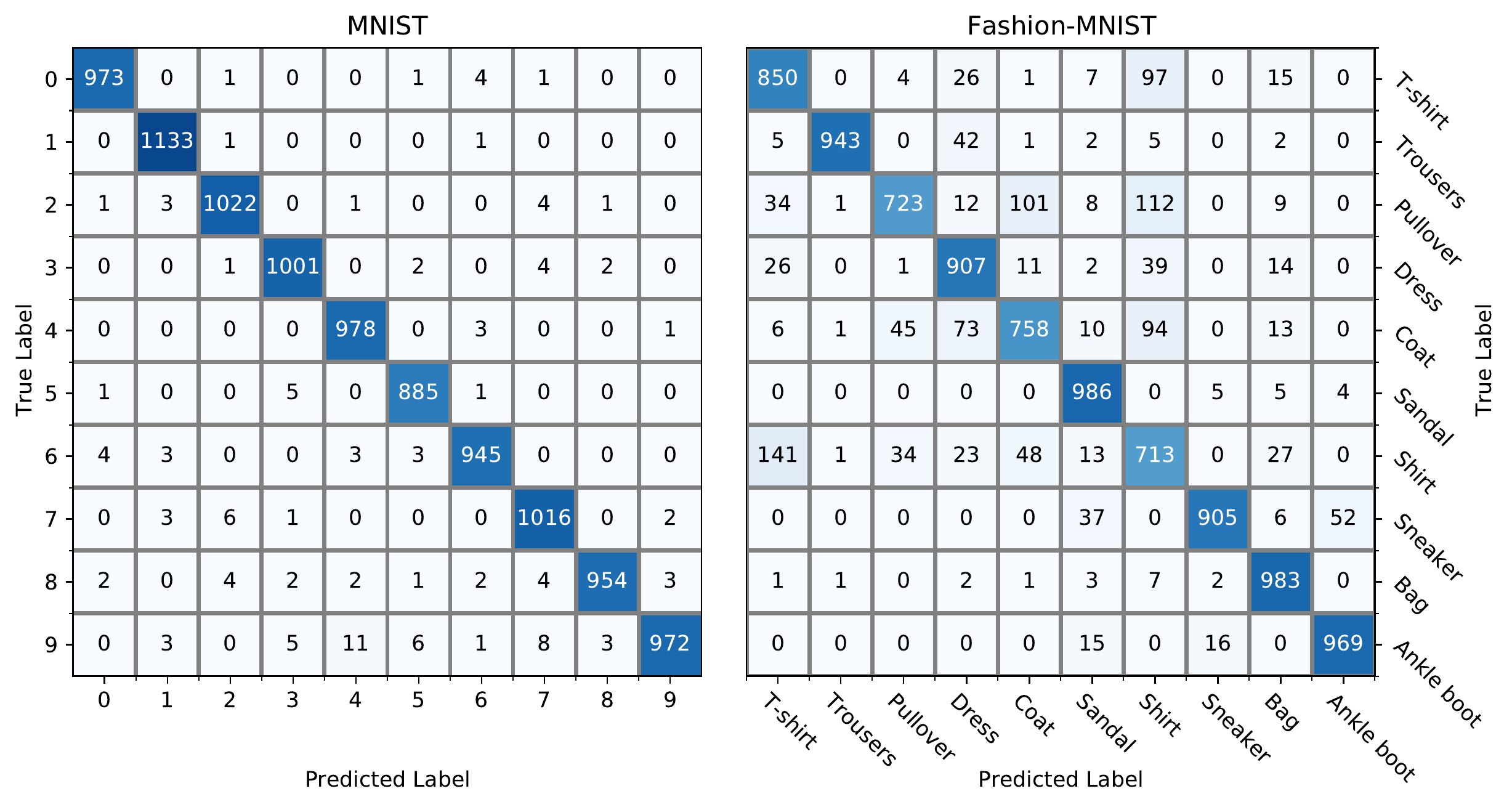}
    \caption{Confusion matrices for MNIST (left) and Fashion-MNIST (right) classifiers. Note that these matrices include all test set examples, not just those which evoke high confidence responses from the classifier.}
    \label{fig:confusion-matrices}
\end{figure}

Figure \ref{fig:confusion-matrices} shows the confusion matrix of the MNIST (left) and Fashion-MNIST (right) classifiers. 

\clearpage
\section{\textsc{Bayes-TrEx} with Saliency Maps}
\label{saliency-maps-supp}

\begin{figure}[!htb]
    \centering
    \includegraphics[width=0.9\columnwidth,trim={1cm 1cm 0 0},clip]{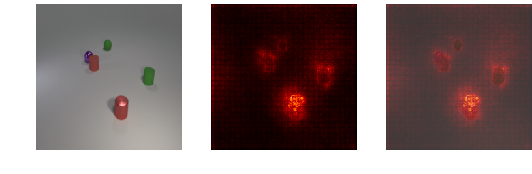}
    \caption{Left: the original image, preprocessed for classification by resizing and normalizing. The classifier is $93.5\%$ confident this scene contains 1 cube, when in fact it is composed of 3 cylinders and 2 spheres. Middle: the SmoothGrad saliency map for this input. Right: the saliency map overlaid upon the original image. This saliency map most strongly highlights the red metal cylinder, indicating that this cylinder is likely the cause of the misclassification.}
    \label{fig:saliency-map-example}
\end{figure}

\begin{figure}[!htb]
    \centering

    \subfigure[$\vec p_\text{1 Cube} = 29.0\%$]{
    \includegraphics[width=0.18\columnwidth,trim={1cm 1cm 0 0},clip]{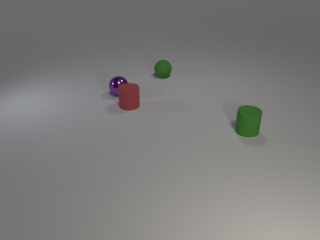}
    }
    \subfigure[$\vec p_\text{1 Cube} = 68.5\%$]{
    \includegraphics[width=0.18\columnwidth,trim={1cm 1cm 0 0},clip]{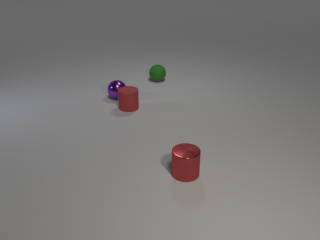}
    }
    \subfigure[$\vec p_\text{1 Cube} = 81.2\%$]{
    \includegraphics[width=0.18\columnwidth,trim={1cm 1cm 0 0},clip]{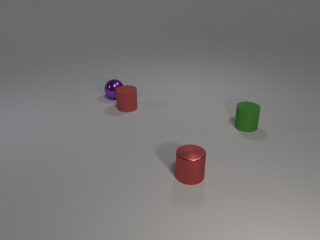}
    }
    \subfigure[$\vec p_\text{1 Cube} = 99.0\%$]{
    \includegraphics[width=0.18\columnwidth,trim={1cm 1cm 0 0},clip]{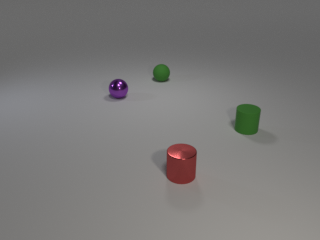}
    }
    \subfigure[$\vec p_\text{1 Cube} = 99.4\%$]{
    \includegraphics[width=0.18\columnwidth,trim={1cm 1cm 0 0},clip]{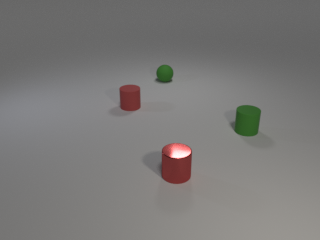}
    }
\caption{Prediction confidence for 1-cube after every single object is removed in turn. As suggested by the saliency map, the removal of the red metal cylinder most prominently reduces the classification confidence, from $93.5\%$ to $29.0\%$. }
\label{fig:saliency-map-red-deleted}
\end{figure}

We demonstrate a simple use case of combining with \textsc{Bayes-TrEx} samples with downstream interpretability methods. Fig.~\ref{fig:saliency-map-example} (left) shows an image for which the classifier mistakes it to contain one cube with 93.5\% accuracy. Fig.~\ref{fig:saliency-map-example} (middle) presents its SmoothGrad~\cite{smilkov2017smoothgrad} saliency map and Fig.~\ref{fig:saliency-map-example} (right) overlays it on top of the image. We can see that the most salient part contributing to the 1-cube decision is the front red cylinder. Indeed, as we confirm in Fig.~\ref{fig:saliency-map-red-deleted}, among all single object removals, removing this object has the biggest effect to the classifier confidence, decreasing it to 29.0\%. 

\clearpage

\begin{figure}[!htb]
    \centering
    \subfigure[Original image: $\vec p_\text{1 Cube} = 85.5\%$. Purple cylinder removed: $\vec p_\text{1 Cube} = 1.9\%$]{
    \includegraphics[width=0.7\linewidth,trim={1cm 1cm 0 0},clip]{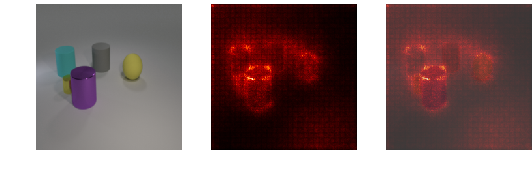}
    \label{in-dist-1-cube-saliency}
    }
    
    \vspace{-0.1in}
    
    \subfigure[Original image: $\vec p_\text{1 Sphere} = 97.9\%$. Yellow cylinder removed: $\vec p_\text{1 Sphere} = 5.2\%$]{
    \includegraphics[width=0.7\linewidth,trim={1cm 1cm 0 0},clip]{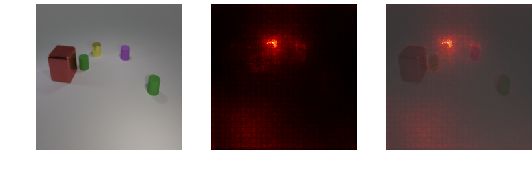}
    \label{in-dist-1-sphere-saliency}
    }
    
    \vspace{-0.1in}
    
    \subfigure[Original image: $\vec p_\text{1 Cylinder} = 85.4\%$. Red sphere removed: $\vec p_\text{1 Cylinder} = 0.9\%$]{
    \includegraphics[width=0.7\linewidth,trim={1cm 1cm 0 0},clip]{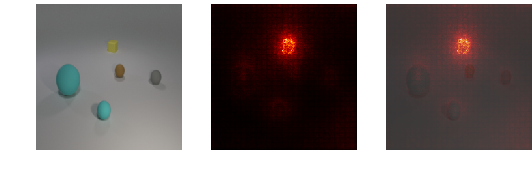}
    \label{in-dist-1-cylinder-saliency}
    }
    
    \vspace{-0.1in}
    
    \subfigure[Original image: $\vec p_\text{1 Cube} = 99.7\%$. Cone removed: $\vec p_\text{1 Cube} = 0.4\%$]{
    \includegraphics[width=0.7\linewidth,trim={1cm 1cm 0 0},clip]{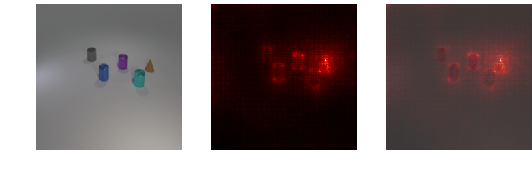}
    \label{ood-1-cube-saliency}
    }
    
    \vspace{-0.1in}
    
    \subfigure[Original image: $\vec p_\text{1 Sphere} = 98.0\%$. Gray cone removed: $\vec p_\text{1 Sphere} = 0.3\%$]{
    \includegraphics[width=0.7\linewidth,trim={1cm 1cm 0 0},clip]{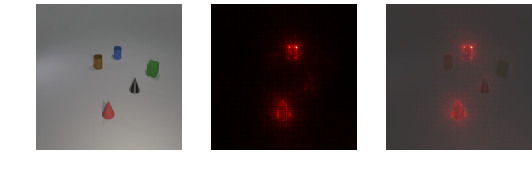}
    \label{saliency-map-failure}
    }
    
    \vspace{-0.1in}
    
    \caption{Images sampled with \textsc{Bayes-TrEx} and their saliency maps. \ref{in-dist-1-cube-saliency}-\ref{in-dist-1-cylinder-saliency} are high confidence misclassified examples; \ref{ood-1-cube-saliency}-\ref{saliency-map-failure} are novel class extrapolation examples. In \ref{saliency-map-failure}, the saliency map primarily highlights two objects: the red cone and the blue cylinder. Removing either of these objects does not result in a change of prediction. Instead, the misclassification of 1 Sphere is due to the marginally-highlighted gray cone.}
    \label{fig:additional-saliency-map-supp}

\end{figure}

Fig.~\ref{fig:additional-saliency-map-supp} presents additional case studies with the same setup. Note that Fig.~\ref{saliency-map-failure} shows a failure of SmoothGrad. 

\end{document}